\documentclass[10pt,journal,compsoc]{IEEEtran}

\ifCLASSOPTIONcompsoc
  \usepackage[nocompress]{cite}
\else
  \usepackage{cite}
\fi

\ifCLASSINFOpdf
  \usepackage[pdftex]{graphicx}
\else
\fi

\usepackage{comment}
\usepackage{url}
\usepackage{tabularx}
\usepackage{booktabs}       
\usepackage{amssymb}
\usepackage{adjustbox}
\usepackage{multirow}
\usepackage{hyperref} 
\usepackage{pifont}
\newcommand{\cmark}{\ding{51}}
\newcommand{\xmark}{\ding{55}}
\usepackage{makecell}

\newcommand{\etal}{\textit{et al}. }
\newcommand{\ie}{\textit{i}.\textit{e}. }
\newcommand{\eg}{\textit{e}.\textit{g}. }

\usepackage{xcolor}
\usepackage{soul}
\usepackage[normalem]{ulem}

{\itshape\color{blue}}%
{\nolinebreak}

\newenvironment{finalquote}%
{\ignorespaces}%
{\nolinebreak}

    {\begin{enumerate}\setlength{\parskip}{6pt}}%
    {\end{enumerate}}

\definecolor{revcolor}{rgb}{0.0, 0.0, 1.0}

\newcommand{\myparagraphwospace}[1]{\noindent{\bf #1.}} 
\newcommand{\myparagraphwithspace}[1]{\vspace{6pt}\noindent{\bf #1.}} 
\def\mypar#1{\vspace{0.15cm}\noindent{\bf #1.}}

\IfFileExists{darkmode.tex}{
    \usepackage{pagecolor}
    \definecolor{myfg}{gray}{0.94} 
    \definecolor{mybg}{gray}{0}
    \input{darkmode.tex} 
    \pagecolor{mybg}
    \color{myfg}
}{}

\def\etal{et al.\ }

\usepackage{amsmath,amssymb,amsbsy,xspace}
\makeatletter
\DeclareRobustCommand\onedot{\futurelet\@let@token\@onedot}
\def\@onedot{\ifx\@let@token.\else.\null\fi\xspace}

\def\eg{\emph{e.g}\onedot} 
\def\ie{\emph{i.e}\onedot} 
 
\def\etc{\emph{etc}\onedot}  
 
\def\etal{\emph{et al}\onedot}

\def\be{\begin{equation}}
\def\ee{\end{equation}}
\def\bea{\begin{eqnarray}}
\def\eea{\end{eqnarray}}

\DeclareMathOperator*{\argmax}{arg\,max}

\makeatother

\begin{document}
\title{Towards Zero-shot Sign Language Recognition}

\author{Yunus~Can~Bilge,
        Ramazan~Gokberk~Cinbis,
        and~Nazli~Ikizler-Cinbis
\IEEEcompsocitemizethanks{\vspace{-.2cm}\IEEEcompsocthanksitem E-mail~addresses: yunuscanbilge@cs.hacettepe.edu.tr, 
gcinbis@ceng.metu.edu.tr,
nazli@cs.hacettepe.edu.tr \protect\\
\vspace{-.2cm} \IEEEcompsocthanksitem Yunus Can Bilge is with Graduate School of Science and Engineering and Nazli Ikizler-Cinbis is with Computer Engineering Department both from Hacettepe University, Ankara/Turkey. Ramazan Gokberk Cinbis is with the Department of Computer Engineering at METU, Ankara/Turkey. \protect \\
 \vspace{-.2cm} \IEEEcompsocthanksitem{This work was performed as a part of the Ph.D. studies of the first author, and supported in part by the TUBITAK Grant 116E445. © 2022 IEEE. Personal use of this material is permitted. Permission from IEEE must be obtained for all other uses, in any current or future media, including reprinting/republishing this material for advertising or promotional purposes, creating new collective works, for resale or redistribution to servers or lists, or reuse of any copyrighted component of this work in other works.}}
}

\markboth{TO APPEAR IN IEEE TRANSACTIONS ON PATTERN ANALYSIS AND MACHINE INTELLIGENCE}%
{Shell \MakeLowercase{\textit{et al.}}: Bare Demo of IEEEtran.cls for Computer Society Journals}

\IEEEtitleabstractindextext{%
\begin{abstract}

This paper tackles the problem of zero-shot sign language recognition (ZSSLR), where the goal is to leverage models learned over the seen sign classes to recognize the instances of unseen sign classes. In this context, readily available textual sign descriptions and attributes collected from sign language dictionaries are utilized as semantic class representations for knowledge transfer. For this novel problem setup, we introduce three benchmark datasets with their accompanying textual and attribute descriptions to analyze the problem in detail. Our proposed approach builds spatiotemporal models of body and hand regions. By leveraging the descriptive text and attribute embeddings along with these visual representations within a zero-shot learning framework, we show that textual and attribute based class definitions can provide effective knowledge for the recognition of previously unseen sign classes. We additionally introduce techniques to analyze the influence of binary attributes in correct and incorrect zero-shot predictions. We anticipate that the introduced approaches and the accompanying datasets will provide a basis for further exploration of zero-shot learning in sign language recognition.

\end{abstract}
}

\maketitle
\IEEEdisplaynontitleabstractindextext

\IEEEpeerreviewmaketitle

\IEEEraisesectionheading{\section{Introduction}\label{sec:intro}}

\IEEEPARstart{S}{ign} language recognition (SLR) is among the open problems with great practical importance in computer vision. One of the factors that render SLR a unique problem is the fact that although most signs have clear-cut definitions, most of them have only subtle visual differences across them \cite{wu1999vision,stokoe2005sign}. Therefore, towards turning SLR into a ubiquitous technology, fundamental progress in modeling and recognizing fine-grained spatiotemporal patterns expressed by the movement of hands with various shapes, orientations, and locations, as well as body posture, and non-manual features, such as facial expressions, are needed. Besides, the photometric and geometric factors, such as viewpoint changes \cite{neidle2012challenges}, variations in sign languages across regions \cite{valli2000linguistics}, and the fact that sign languages embrace significant variations over time \cite{lucas2011variation} \begin{finalquote} render the task more challenging. \end{finalquote} 

The existing SLR approaches typically require a large number of annotated examples for each  sign class of interest \cite{cihan2018neural,camgoz2017subunets,koller2015continuous,koller2016deep,stoll2018sign}. This dependency means that one presumably needs to collect annotated samples for all signs in all sign languages of interests, including variations expressed by multiple persons per sign under various recording conditions. Noting that over 140 sign languages and many more dialects are estimated to exist around the world \cite{lewis2015simons}, with typically 2500 to 5500 signs per language \cite{valli2006gallaudet, brien1992dictionary, johnston1998signs}, the need for supervised examples create a {\em data bottleneck} problem in scaling up SLR.  Towards bypassing this problem, we explore the possibility of recognizing sign classes with no annotated visual examples, purely based on existing definitions, which we call \textit{zero-shot sign language recognition} (ZSSLR). Unlike the traditional supervised SLR, where training data is needed for all classes of interest, in ZSSLR, the aim is to recognize novel sign classes that do not have any visual training samples. 
We additionally introduce the problem of \textit{generalized zero-shot sign language recognition} (GZSSLR), which allows joint evaluation of the ability of a model to generalize out of the training set for classes with training samples and recognize novel classes.

In order to transfer knowledge across classes, we use textual descriptions of signs taken from a sign language dictionary, together with attribute descriptions gathered from a sign hand shapes dictionary. Using a sign language dictionary for obtaining the class representations has two significant advantages for being
(i) readily available, and (ii) prepared by the sign language experts in a detailed way.
In this manner, we explore the expressive power of both the textual sign language descriptions and the attributes acquired from dictionaries to form the auxiliary information that is required for recognizing novel sign language classes. Note that, since this auxiliary data is acquired from dictionaries, there is no need for any ad-hoc manual annotations.

\begin{figure*}
\centering
\includegraphics[width=0.9\textwidth]{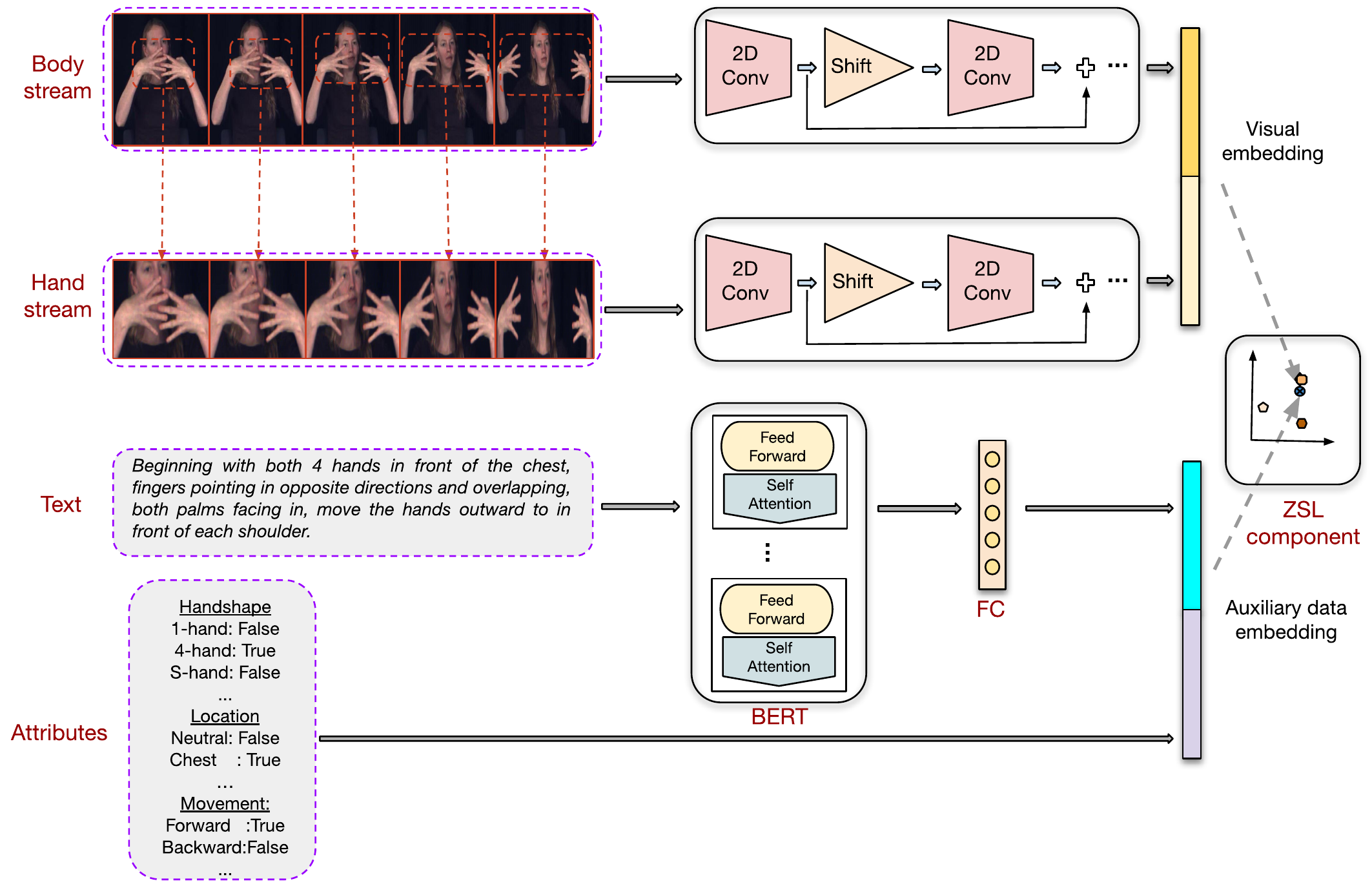}
   \caption{Illustration of the proposed zero-shot sign language recognition (ZSSLR) approach. Two separate streams are used for both visual and auxiliary class representations. Visual representation is obtained by encoding body and estimated hand streams through spatiotemporal deep architectures. Class embeddings are obtained by encoding textual dictionary definitions and attribute combinations. We learn a compatibility function that links the visual representation to the auxiliary class embeddings for zero-shot recognition. \label{fig:main}}
\end{figure*}

To study the ZSSLR problem, we introduce three benchmark datasets. We obtain the first one, \begin{finalquote} ASL-Text \end{finalquote}, by adapting the ASLLVD dataset~\cite{neidle2012challenges}. We define the next two ones, MS-ZSSLR-C and MS-ZSSLR-W, by adjusting and manually filtering the MS-ASL~\cite{vaezijoze2019ms-asl} dataset. We augment all three datasets with textual and attribute-based class descriptions based on sign language dictionaries.

To tackle the ZSSLR problem, we propose an embedding-based framework that consists of two main components. The first component aims to capture the temporal and spatial patterns, for which we offer alternative architectures based on 3D-CNN, LSTM, and the recently introduced shift-based CNN \cite{lin2019tsm} modules. This component operates over the body and hand regions extracted from videos in parallel. The second component, \ie, zero-shot learning (ZSL) component, learns a scoring function for measuring the compatibility between a given visual representation and text and/or attribute-based sign description. We provide an overview in Figure~\ref{fig:main}. We rigorously evaluate our approach on our three benchmark datasets for both ZSSLR and GZSSLR problems and analyze the results.

A preliminary version of this work has previously appeared in \cite{bilge19zsslr}, \begin{finalquote} where ZSSLR problem was first formulated and the benchmark dataset ASL-Text was defined. The applied model in \cite{bilge19zsslr} was based on 3D-CNNs and LSTMs, the benchmark results were presented on ASL-Text. \end{finalquote} This work extends it in multiple ways: first, we improve the semantic embeddings of sign classes by collecting new attribute-based representations and show that our attribute-based class definitions yield significant empirical improvements, mainly when used in combination with textual descriptions. Second, we add new temporal shift modules to improve the spatiotemporal video representation. Third, we extend our study by introducing the GZSSLR problem in addition to ZSSLR. Fourth, we introduce two new (G)ZSSLR benchmark datasets. Fifth, we extend our empirical analyses with additional insights into the ZSSLR problems over the three benchmarks, \ie ASL-Text, MS-ZSSLR-W, MS-ZSSLR-C. Finally, we propose two new techniques for analyzing the impact of binary attribute definitions of unseen classes to help us gain insights about attribute-based correct and incorrect zero-shot predictions, respectively.

The remainder of the paper is organized as follows. We start by reviewing the related literature in Section~\ref{sec:relwork}. We explain the proposed ZSSLR benchmarks in Section~\ref{sec:dataset}. We present our approach to the problem and the attribute analysis technique in Section~\ref{sec:method}. We present a detailed evaluation and analysis of our models in Section~\ref{sec:exp}. We conclude the paper in Section~\ref{sec:conclusions}.

\section{Related Work}
\label{sec:relwork}

\subsection{Sign Language Recognition (SLR)} SLR is a major computer vision task that has been studied for more than three decades \cite{tamura1988recognition}. The mainstream SLR approaches can be grouped into two categories: (i) isolated SLR \cite{wang2016isolated}, where a single sign is recognized at a time, and (ii) continuous SLR \cite{cihan2018neural}, where one or more sentences are recognized over the streaming input. Our work belongs to isolated SLR category as we opt to focus on the study sign recognition with incomplete supervision, isolated from the orthogonal challenges of recognition on streaming data.

Early SLR methods mostly use hand-crafted features in combination with a classifier like support vector machine \cite{tamura1988recognition, waldron1995isolated, kadous1996machine, zahedi2005combination, cooper2007sign}. Hidden Markov Models (HMM), Conditional Random Fields and neural network based approaches have also been explored to model the temporal patterns \cite{grobel1997isolated, huang1998sign}. Recently, several deep learning based SLR approaches have been proposed \cite{camgoz2020sign, li2020word, saunders2020progressive}.

Despite the relative popularity of the topic, the problem of annotated data scarcity has been seldomly addressed in SLR research. Farhadi and Forsyth \cite{farhadi2006aligning} are the first to study the alignment of sign language video subtitles and signs in action to overcome annotation difficulty. Similarly, \cite{farhadi2007transfer} proposes to transfer large amounts of labeled avatar data with few labeled human signer videos to spot words in videos. Buehler \etal~\cite{buehler2009learning} aim to reduce the annotation effort by using the subtitles of British Sign Language TV broadcasts. The idea is to apply Multiple Instance Learning (MIL) to recognize signs out of TV broadcast subtitles. There are also other studies that utilizes the subtitles of TV broadcasts \cite{kelly2011weakly}, \cite{pfister2013large}. Pfister \etal~\cite{pfister2013large} differ from the two aforementioned MIL studies as these studies track the co-occurrences of lip and hand movements to reduce the search space for visual and textual content mapping. Nayak \etal~\cite{nayak2009automated} propose to locate signs in continuous sign language sentences using iterated conditional modes. 
Pfister \etal \cite{pfister2014domain} define each sign class with one strongly supervised example and train an SVM based detector out of these one-shot examples. The resulting detector is then used to acquire more training samples from another weakly-labeled data. Koller \etal \cite{koller2016deep} propose a combined CNN and HMM approach to train a model with large but noisy data. Koller \etal \cite{koller2017re, koller2019weakly} later suggest using CNN-HMM with LSTM for weakly-labeled noisy data. Albanie \etal \cite{albanie20_bsl1k} underline the difficulty in collecting sign language data and propose to extract annotations using keyword spotting in interpreted TV broadcasts. The key idea behind this approach is that signers mostly mouth the words they are signing. \begin{finalquote} Recently, Momeni \etal \cite{momeni2020watch} propose a unified Multiple Instance Learning sign spotting framework in continuous sign language videos and evaluate the model in low-shot settings. \end{finalquote} None of the aforementioned models approach the problem of annotated data scarcity from a zero-shot learning perspective, which can potentially play a central role in modeling larger vocabularies and a large number of languages.

\begin{table}[t]
\centering
\caption{Statistics of the origin datasets; ASLLVD and MS-ASL.}
\vspace{-3mm}
\label{fig:originaldatasetstatistics}
\small{
\scalebox{0.7}{
\begin{tabular}{lrrrrr}
\toprule
Dataset & Classes  & Signer & Videos & Videos per Class & Controlled \\
\midrule
RWTH-Boston-50\cite{zahedi2005combination} & 50 & 3 & 201 & 4 & \cmark\\
Purdue ASL \cite{wilbur2006purdue} & 104 & 14 & 1834 & 18 & \cmark \\
SIGNUM \cite{von2008significance} & 450 & 25 & 12150 & 27 & \cmark \\
LSA64 \cite{ronchetti2016lsa64} & 64 & 10 & 3200 & 50 & \cmark \\
GSL Corpus \cite{efthimiou2007gslc} & 981 & 2 & 9810 & 10 & \cmark \\ 
DEVISIGN \cite{chai2014devisign} & 2000 & 8 & 24000 & 12 & \cmark \\
ASLLVD\cite{neidle2012challenges} & \textbf{3300} & 1-6 & 9800 &  3 & \cmark \\
MS-ASL\cite{vaezijoze2019ms-asl}  & 1000 & 11-45 & \textbf{25513} & 25 & \xmark  \\
\bottomrule
\end{tabular}
}}
\vspace{-2mm}
\end{table}

\subsection{Sign language datasets}
There are many SL datasets available for both isolated \cite{zahedi2005combination, wilbur2006purdue, efthimiou2007gslc, ronchetti2016lsa64, neidle2012challenges, chai2014devisign, vaezijoze2019ms-asl, von2008significance} and continuous sign language recognition
\cite{von2008significance, roussos2013dynamic, forster2014extensions, huang2018video, albanie20_bsl1k}. Table~\ref{fig:originaldatasetstatistics} summarizes the details of the currently available public datasets that include isolated sign data to the best of our knowledge. ASLLVD \cite{neidle2012challenges} includes the most class count while MS-ASL \cite{vaezijoze2019ms-asl} includes the most in-class sample count. As opposed to most of the available isolated sign language datasets that their recordings are performed in a controlled laboratory environment, MS-ASL \cite{vaezijoze2019ms-asl} covers unconstrained sign recordings concerning large variation in signer, background and positioning. We have chosen ASLLVD \cite{neidle2012challenges} and MS-ASL \cite{vaezijoze2019ms-asl} a basis for constructing our zero-shot oriented datasets because of their large class count with challenging real-life recording conditions. 

\vspace{-2mm}
\subsection{Zero-Shot Learning (ZSL)}
ZSL problems have received significant attention over the past several years, following the pioneering works of Lampert \etal~\cite{lampert2009learning} and Farhadi \etal~\cite{farhadi2009describing}. The main goal in ZSL is learning to generalize a recognition model for identifying unseen classes. Most of the ZSL approaches rely on transferring semantic information from seen to unseen classes utilizing auxiliary information. Among the possible sources of auxiliary information, attributes and textual representations stand out as powerful and
practical options, for which we present a brief overview in the following.

\mypar{Attribute-based ZSL} 
Attribute-based ZSL approaches in computer vision rely on class representations based on the information about the relevance of each visual attribute for each class. Attribute information can take many forms, including binary~\cite{ferrari2008learning}, continuous~\cite{jayaraman2014zero} or relative~\cite{parikh2011relative}. Attributes have been used for image classification~\cite{lampert2009learning, akata2013label}, image description~\cite{ferrari2008learning, farhadi2009describing}, object detection~\cite{lampert2014attribute, li2014attributes}, and caption generation~\cite{kulkarni2013babytalk, ordonez2011im2text}. We use attributes for high level representation of the sign classes. 

Earlier zero-shot recognition approaches use attribute prediction \cite{farhadi2009describing, lampert2009learning}, where each attribute is learned
independently. More recent works typically focus on directly modeling relations between image
features and class attributes, see, \eg \cite{akata2013label}.
Jayaraman \etal \cite{jayaraman2014zero} propose to model
attribute relationships. In recent studies \cite{escorcia2015relationship, peng2016joint}, deep
learning based approaches to learn visual attributes is proposed.

\mypar{Text-based ZSL} In text-based ZSL, the semantic space and the representation of a class are constructed out of textual descriptions.
Semantic word/sentence vectors and concept ontologies have been studied in this context \cite{rohrbach2011evaluating,  elhoseiny2013write, mensink2014costa, lei2015predicting}. Label embedding models are explored to make a connection between seen and unseen classes via semantic representations \cite{akata2013label, fu2014transductive, romera2015embarrassingly, qin2017zero}. Textual descriptions are acquired from various resources; such as Wikipedia \cite{qiao2016less}. In our case, we use a sign language dictionary as the textual information source.
In order to utilize these textual resources, various unsupervised embedding models exist, such as word2vec \cite{mikolov2013distributed}, GloVe \cite{pennington2014glove}, and BERT \cite{devlin2018bert}.

\mypar{Zero-shot action recognition} In relation to SLR, ZSL of human actions has been studied. Liu \etal~\cite{liu2011recognizing} are the first to propose attribute based model for recognizing novel actions. Jain \etal~\cite{jain2015objects2action} propose a semantic embedding based approach using commonly available textual descriptions, images, and object names. Xu \etal~\cite{Xu2015SemanticES} propose a regression based method to embed actions and labels to a common embedding space. 
 Xu \etal~\cite{xu2017transductive} also use word-vectors as a semantic action embedding space in a transductive setting. Wang \etal \cite{wang2017alternative} exploit human actions via related textual descriptions and still images, where the idea is to improve word vector semantic representations with additive information. Habibian \etal \cite{habibian2017video2vec} also propose to learn semantic representations of videos with freely available video and relevant descriptions. Qin \etal~\cite{qin2017zero} use error-correcting output codes to overcome the disadvantages of attributes and/or semantic word embeddings of actions.  
Hahn \etal~\cite{hahn2019action2vec} explore the idea of learning a cross-modal embedding space for textual features of class labels and spatiotemporal features of action videos.

Compared to action recognition, in SLR, even a subtle change in motion and/or handshape can change the entire meaning, which makes SLR a different problem that requires specialized methods for zero-shot recognition. Relatively speaking, gesture recognition is a problem with more similarities to SLR. While there are a couple of recent methods on zero-shot gesture recognition, these methods are limited to either gestures for human-robot interaction~\cite{thomason2016recognizing}, or, few classes, \eg, 10 hand gesture classes~\cite{madapana2018hard}. In this work, we introduce and work on much larger-scale and comprehensive datasets for sign language recognition and incorporate both textual and attribute-based sign representations to our study.

\mypar{Understanding the influence of class attributes} Recently, there has been a great interest in
developing techniques for producing explanations of predictions made by deep architectures.  Most
main-stream works introduce methods that can be used to estimate the importance of input features,
such as LIME~\cite{ribeiro_why_2016}, DeepSHAP~\cite{lundberg_unified_2017,chen_explaining_2019},
and Grad-CAM~\cite{grad_cam_2017}. 
Only very few recent works introduce explanation approaches for zero-shot learning models: Selvaraju
\etal~\cite{choose_neuron_2018} define a method towards associating neurons with attributes and
uses the associations learnt by the model to determine attributes with largest influence on the
resulting class scores. Liu and Tuytelaars~\cite{liu_deep_2020} propose a ZSL model that can
produce visual and textual explanations for recognition results. In our work, we are interested in
understanding the role of binary class attributes on zero-shot predictions. We note that this
differs from analyzing input feature dimensions as attributes can be considered as
auxiliary model parameters that are used to infer the class predictors, as opposed to being
instance-specific input data.

\vspace{-0.3cm}
 \section{Proposed Datasets}
\label{sec:dataset}

Since there is no available dataset for ZSSLR, we repurpose two existing supervised sign language datasets to create three ZSSLR benchmarks: \textit{ASL-Text}, \textit{MS-ZSSLR-W} and \textit{MS-ZSSLR-C}. These datasets jointly provide a rich experimental setup thanks to significant differences across them, \eg, in terms of per-class sample counts and embracing controlled or uncontrolled recording setups. We provide the details of the datasets in the following paragraphs, followed by the explanations about the way we obtain the text-based and attribute-based class definitions.

\subsection{ASL-Text dataset}

To the best of our knowledge, the ASLLVD dataset~\cite{neidle2012challenges} is the largest isolated sign language recognition dataset available in vocabulary size. We select the top 250 sign classes, ranked by the number of samples and number of signers per class. We then augment this dataset with the textual definitions of the signs from Webster American Sign Language Dictionary~\cite{costello1999random} and attribute definitions from American Sign Language Handshape Dictionary~\cite{tennant1998american}. We refer to this new benchmark dataset as \textit{ASL-Text}. Example frames and their textual descriptions from the ASL-Text dataset are presented in the upper half of Figure~\ref{fig:dataset}.

In this dataset, the average sequence length for a video sample is $33$ frames. While splitting the dataset into three disjoint train, validation and test sets, the classes with the most signer variations, and in-class samples are assigned to the training set. The remaining classes with relatively fewer visual examples are allocated into the validation and test sets to obtain a more realistic experimental setup. 
The average length of a textual description is $30$ words, where the total vocabulary includes 154 distinct words.   
With only $6.3$ samples per class, the dataset provides considerably few training samples, in comparison to commonly used zero-shot image classification datasets, such as aPY~\cite{farhadi2009describing} and AWA-2~\cite{xian2017zero}, which contain approximately $630$ and $750$ samples per training class, respectively.

\subsection{MS-ZSSLR-W/C datasets}

We construct our other ZSSLR benchmarks by adapting the recently introduced {\em MS-ASL} large-scale dataset~\cite{vaezijoze2019ms-asl}. 
The MS-ASL dataset provides a compelling collection of videos with challenging and realistic sign language sequences.  

The dataset is originally collected from public videos with manual captions, descriptions, titles, and subtitles. These videos are processed using automated techniques such as Optical Character Recognition (OCR) to capture the labels of longer video clips, title extraction to be used as the label in the short ones. As a result of these semi-automatic procedures, the dataset contains some artifacts, such as repetitive signs in some single-sign sequences. In addition, most sign classes do not include the standard sign examples but also their sign dialects as well. 

We construct two separate benchmarks based on MS-ASL: (i) {\bf MS-ZSSLR-W}(ild), which can be seen as an {\em in-the-wild} test bed and (ii) {\bf MS-ZSSLR-C}(lean), which provides a more controlled and cleaner benchmark. For both, we first select the largest 200 of the available sign classes in MS-ASL, ranked by the number of samples per class. We then augment the dataset with textual and attribute-based descriptions as we do for constructing the ASL-Text benchmark. We define the resulting dataset, with minimal modifications \footnote{As an exception, four of the top-200 (\textit{out/outside}, \textit{bus}, \textit{light} and \textit{born}) classes are replaced with top-250 (\textit{money}, \textit{wife}, \textit{week}, \textit{old}) counterparts due to the mismatch between visual sign execution and relevant textual and attribute dictionary descriptions.}, as the MS-ZSSLR-W benchmark.

The fact that MS-ASL contains language dialect variations, in addition to the within-language variations (\eg photometric, geometric, and person-to-person differences), is considered as an advantage for supervised sign recognition purposes. However, these lingual variations can equally be regarded as {\em label noise} that reduces the quality of the experimental setup for ZSL purposes: each dialect should typically be accompanied by its own sign class definitions. For instance, there are multiple ways of signing \textit{pizza} in ASL, varying across geographical regions \cite{lucas2003s}. One popular variation of \textit{pizza} sign mimics shoving a piece of pizza in your mouth, while another highly used variation use \textit{Double-Z} and \textit{A} handshape in conjunction sequentially, which corresponds to completely different signs.\footnote{Variations of the \textit{pizza} sign can be observed in \url{http://www.lifeprint.com/asl101/pages-signs/p/pizza.htm}.} Since we have neither the dialect annotations in MS-ASL videos and nor the per-dialect sign definitions, the ZSL experimental results based on reference sign class definitions mixed with non-compliant sign examples can be misleading. 

Based on these observations, we construct a second MS-ASL based benchmark with minimal data and label noise. For this purpose, we have manually verified each video sample and excluded the dialects, and retained only the sign samples that adhere to the reference. We have also split the repetitive sequences into multiple sequences to be used as separate samples. This constitutes our third benchmark, MS-ZSSLR-C. The number of frames per video in these MS-ZSSLR-W/C datasets range from 4 to 298.

\begin{figure*}
\centering
\begin{tabular}{>{\raggedright\arraybackslash}p{0.235\linewidth}@{$\;\;$}>{\raggedright\arraybackslash}p{0.235\linewidth}@{$\;\;$}>{\raggedright\arraybackslash}p{0.235\linewidth}@{$\;\;$}>{\raggedright\arraybackslash}p{0.235\linewidth}}
\toprule
{\textit{ASL-Text Dataset}} \\
\scriptsize{\textbf{BICYCLE}} & 
\scriptsize{\textbf{EAT}} & 
\scriptsize{\textbf{ALONE}} & 
\scriptsize{\textbf{LIBRARY}} \\
\includegraphics[width=0.25\linewidth, height=1.5cm]{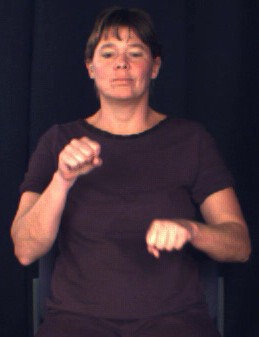}
\includegraphics[width=0.25\linewidth, height=1.5cm]{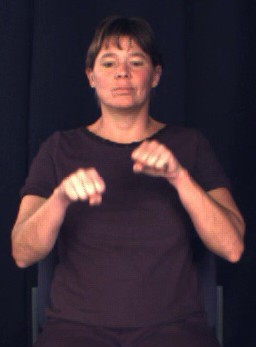}
\includegraphics[width=0.25\linewidth, height=1.5cm]{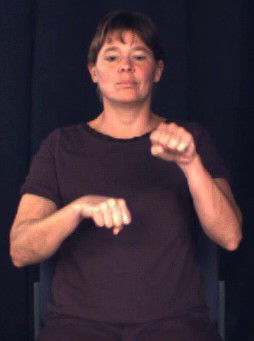} 
\begin{minipage}{0.23\textwidth}
\vspace{1mm}
\tiny{Move both \textcolor{red}{S} hands in alternating forward circles, palms facing down, in front of each side of the body.\\
\textbf{Handshape:}  \includegraphics[width=0.15\linewidth, height=0.5cm]{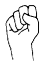} \quad S hand\\
\textbf{Orientation:} \includegraphics[width=0.15\linewidth, height=0.5cm]{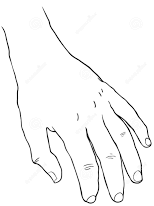} \quad Palms Down \\
\textbf{Location:} \qquad \includegraphics[width=0.15\linewidth, height=0.5cm]{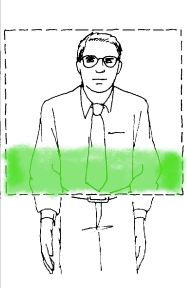} \enspace Neutral \\
\textbf{Movement:} \quad \includegraphics[width=0.15\linewidth, height=0.5cm]{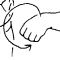} \enspace Circular \\
\textbf{Others:} Repeats, Two-hand sign \\ \\ }
\end{minipage} 
 & 
 \includegraphics[width=0.25\linewidth, height=1.5cm]{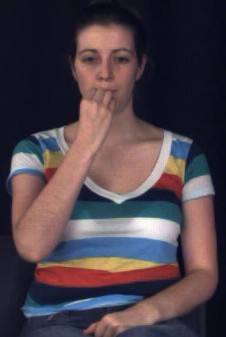}
\includegraphics[width=0.25\linewidth, height=1.5cm]{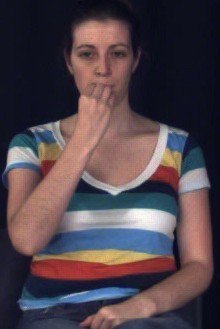}
\includegraphics[width=0.25\linewidth, height=1.5cm]{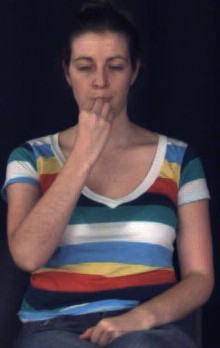}
\begin{minipage}{0.23\textwidth}
\vspace{1mm}
\tiny{Bring the fingertips of the right flattened \textcolor{red}{O} hand, palm facing in, to the lips with a repeated movement. \\
\textbf{Handshape:}  \includegraphics[width=0.15\linewidth, height=0.5cm]{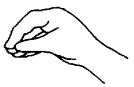} \quad Flattened O hand\\
\textbf{Orientation:} \includegraphics[width=0.15\linewidth, height=0.5cm]{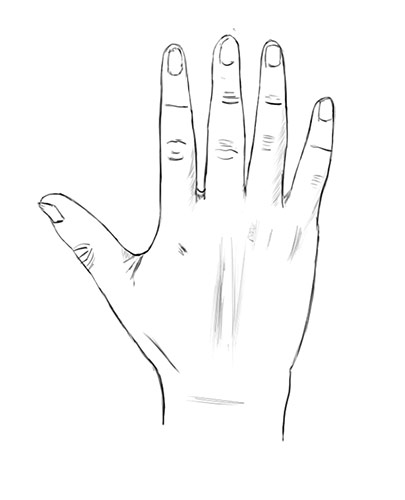} \quad Palm In \\
\textbf{Location:} \qquad \includegraphics[width=0.15\linewidth, height=0.5cm]{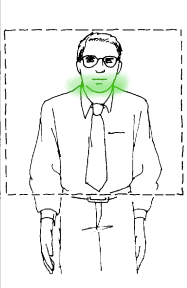} \enspace Cheek/Chin/Mouth/Nose \\
\textbf{Movement:} \quad \includegraphics[width=0.15\linewidth, height=0.5cm]{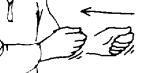} \enspace Towards Body \\
\textbf{Others:} Repeats, One-hand sign \\ \\ }
\end{minipage} 
 & 
 \includegraphics[width=0.25\linewidth, height=1.5cm]{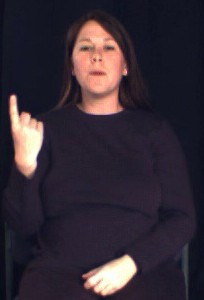}
\includegraphics[width=0.25\linewidth, height=1.5cm]{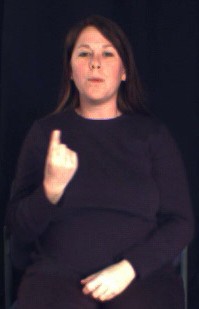}
\includegraphics[width=0.25\linewidth, height=1.5cm]{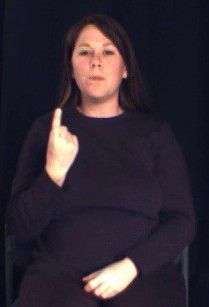}
\begin{minipage}{0.23\textwidth}
\vspace{1mm}
\tiny{With the right index finger extended up, move the right hand, palm facing back, in a small repeated circle in front of the right shoulder.\\
\textbf{Handshape:}  \includegraphics[width=0.15\linewidth, height=0.5cm]{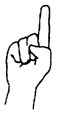} \quad 1 hand\\
\textbf{Orientation:} \includegraphics[width=0.15\linewidth, height=0.5cm]{images/handshapes/palm_face-back.jpg} \quad Palm In \\
\textbf{Location:} \qquad \includegraphics[width=0.15\linewidth, height=0.5cm]{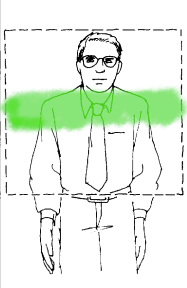} \enspace Shoulder \\
\textbf{Movement:} \quad \includegraphics[width=0.15\linewidth, height=0.5cm]{images/handshapes/circular.png} \enspace Circular \\
\textbf{Others:} Repeats, One-hand sign \\}
\end{minipage} 
& 
\includegraphics[width=0.25\linewidth, height=1.5cm]{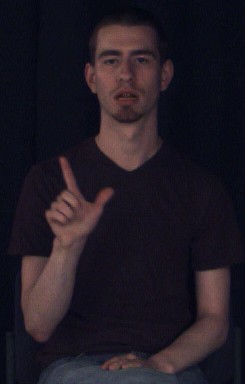}
\includegraphics[width=0.25\linewidth, height=1.5cm]{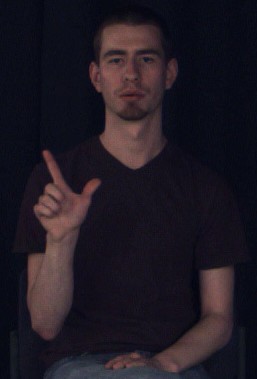}
\includegraphics[width=0.25\linewidth, height=1.5cm]{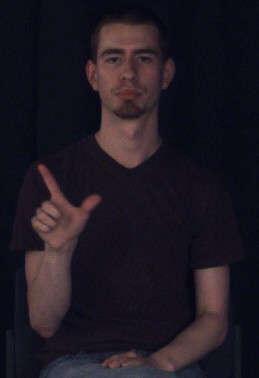}
\begin{minipage}{0.23\textwidth}
\vspace{1mm}
\tiny{Move the right \textcolor{red}{L} hand, palm facing forward, in a circle in front of the right shoulder. \\
\textbf{Handshape:}  \includegraphics[width=0.15\linewidth, height=0.5cm]{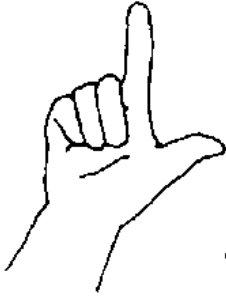} \quad L hand\\
\textbf{Orientation:} \includegraphics[width=0.15\linewidth, height=0.5cm]{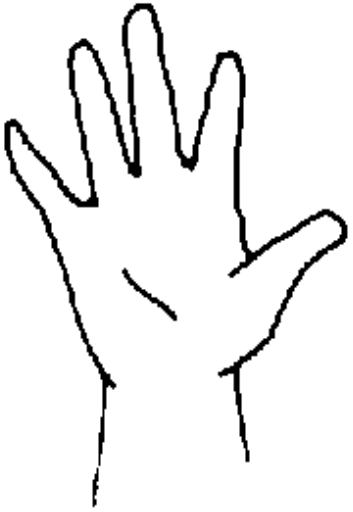} \quad Palm Out \\
\textbf{Location:} \qquad \includegraphics[width=0.15\linewidth, height=0.5cm]{images/handshapes/neutral_space.png} \enspace Neutral \\
\textbf{Movement:} \quad \includegraphics[width=0.15\linewidth, height=0.5cm]{images/handshapes/circular.png} \enspace Circular \\
\textbf{Others:} Repeats, One-hand sign \\}
\end{minipage}
\\
{\textit{MS-ZSSLR-W(ild)/C(lean)}} \\
\vspace{-4mm}
\scriptsize{\textbf{BOOK}} & 
\vspace{-4mm}
\scriptsize{\textbf{HORSE}} & 
\vspace{-4mm}
\scriptsize{\textbf{ORANGE}} & 
\vspace{-4mm}
\scriptsize{\textbf{FISH}} \\
\includegraphics[width=0.25\linewidth, height=1.5cm]{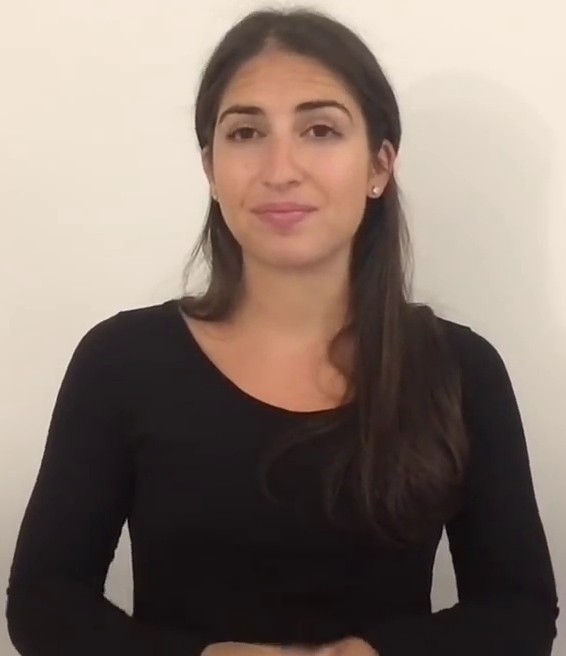}
\includegraphics[width=0.25\linewidth, height=1.5cm]{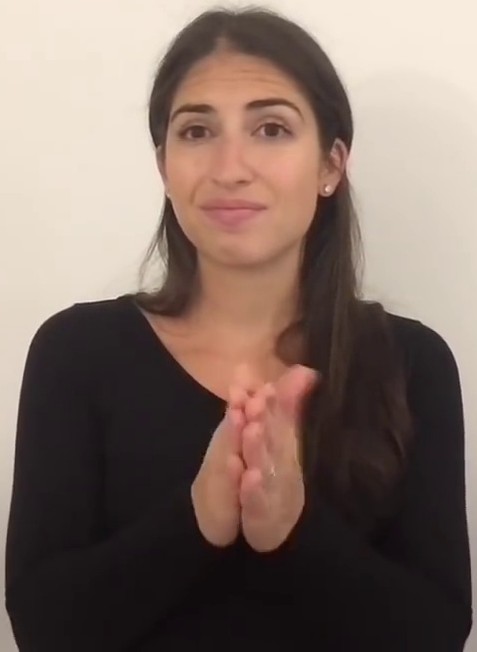}
\includegraphics[width=0.25\linewidth, height=1.5cm]{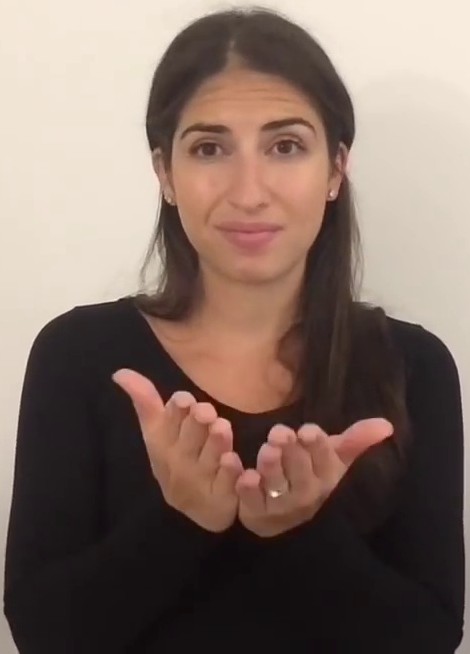}
\begin{minipage}{0.23\textwidth}
\vspace{1mm}
\tiny{Beginning with the palms of both open hands together in front of the chest, fingers angled forward, bring the hands apart at the top while keeping the little fingers together. \\
\textbf{Handshape:}  \includegraphics[width=0.15\linewidth, height=0.5cm]{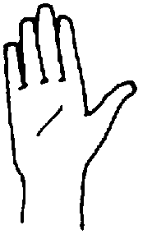} \quad Open B hand\\
\textbf{Orientation:} \includegraphics[width=0.15\linewidth, height=0.5cm]{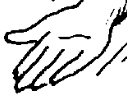} \quad Palm Up \\
\textbf{Location:} \qquad \includegraphics[width=0.15\linewidth, height=0.5cm]{images/handshapes/neutral_space.png} \enspace Neutral \\
\textbf{Movement:} \quad \includegraphics[width=0.15\linewidth, height=0.5cm]{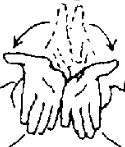} \enspace Wrist Movement \\
\textbf{Others:} No Repeat, Two-hand sign \\}
\end{minipage}
 & 
\includegraphics[width=0.25\linewidth, height=1.5cm]{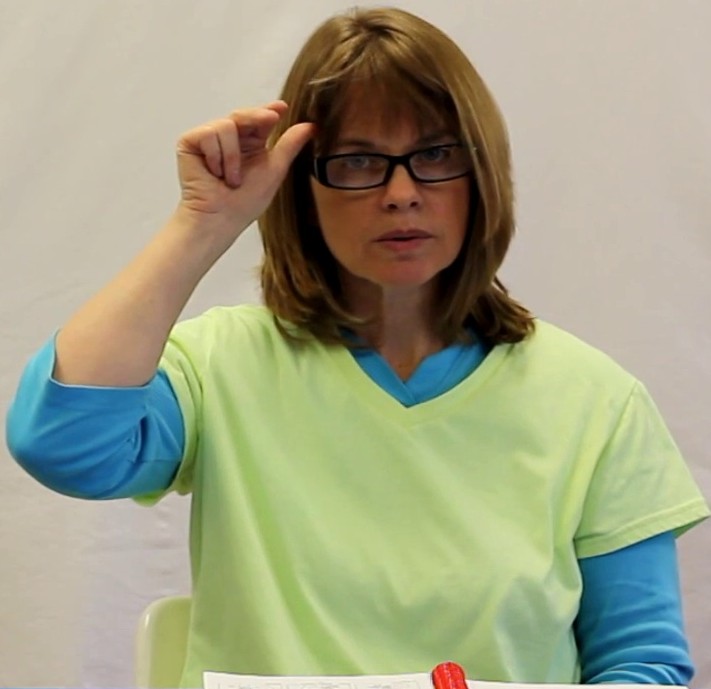}
\includegraphics[width=0.25\linewidth, height=1.5cm]{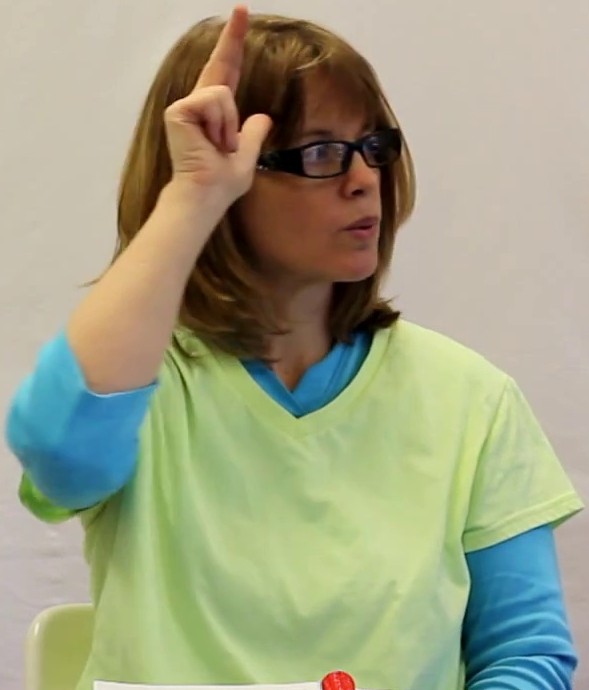}
\includegraphics[width=0.25\linewidth, height=1.5cm]{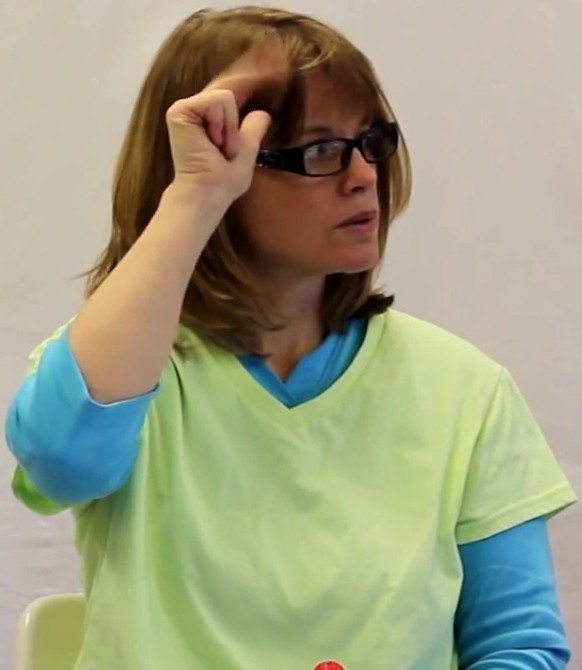}
\begin{minipage}{0.23\textwidth}
\vspace{1mm}
\tiny{With the extended thumb of the right \textcolor{red}{U} hand against the right side of the forehead, palm facing forward, bend the fingers of the \textcolor{red}{U} hand up and down with a double movement. \\
\textbf{Handshape:}  \includegraphics[width=0.15\linewidth, height=0.5cm]{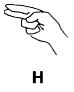} \quad H hand\\
\textbf{Orientation:} \includegraphics[width=0.15\linewidth, height=0.5cm]{images/handshapes/palm_out.png} \quad Palm Out \\
\textbf{Location:} \qquad \includegraphics[width=0.15\linewidth, height=0.5cm]{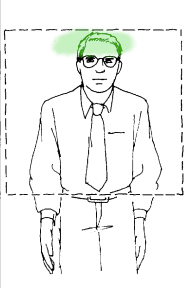} \enspace Temple or Ear \\
\textbf{Movement:} \quad \includegraphics[width=0.15\linewidth, height=0.5cm]{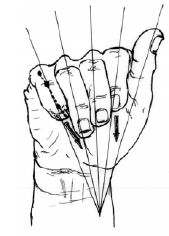} \enspace Finger Movement \\
\textbf{Others:} Repeats, One-hand sign \\}
\end{minipage} 
 & 
 \includegraphics[width=0.25\linewidth, height=1.5cm]{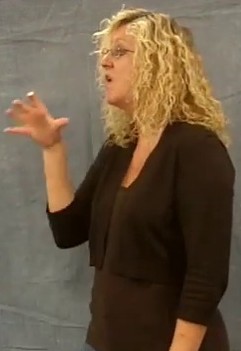}
\includegraphics[width=0.25\linewidth, height=1.5cm]{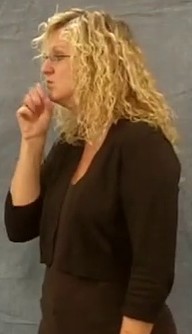}
\includegraphics[width=0.25\linewidth, height=1.5cm]{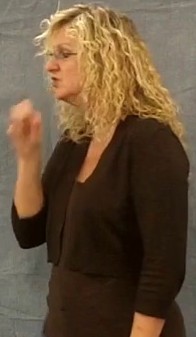}
\begin{minipage}{0.23\textwidth}
\vspace{1mm}
\tiny{Beginning with the right \textcolor{red}{C} hand in front of the mouth, palm facing left, squeeze the fingers open and closed with a repeated movement, forming an \textcolor{red}{S} hand each time.\\
\textbf{Handshape:}  \includegraphics[width=0.15\linewidth, height=0.5cm]{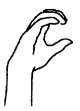} \includegraphics[width=0.15\linewidth, height=0.5cm]{images/handshapes/s.png} \quad C and S hand\\
\textbf{Orientation:} \includegraphics[width=0.15\linewidth, height=0.5cm]{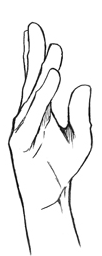} \quad \quad \quad \quad \quad Palm Left \\
\textbf{Location:} \qquad \includegraphics[width=0.15\linewidth, height=0.5cm]{images/handshapes/mouth.png} \enspace Cheek/Chin/Mouth/Nose \\
\textbf{Movement:} \quad \includegraphics[width=0.15\linewidth, height=0.5cm]{images/handshapes/finger_movement.png} \enspace Finger Movement \\
\textbf{Others:} Repeats, One-hand sign \\}
\end{minipage} 
& 
\includegraphics[width=0.25\linewidth, height=1.5cm]{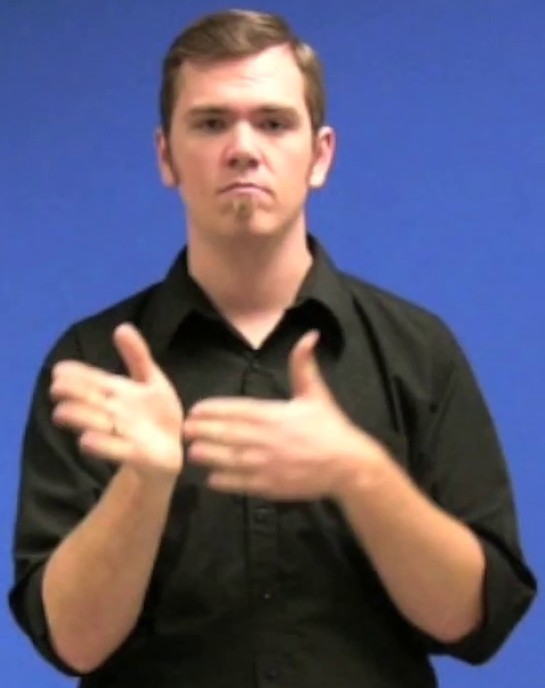}
\includegraphics[width=0.25\linewidth, height=1.5cm]{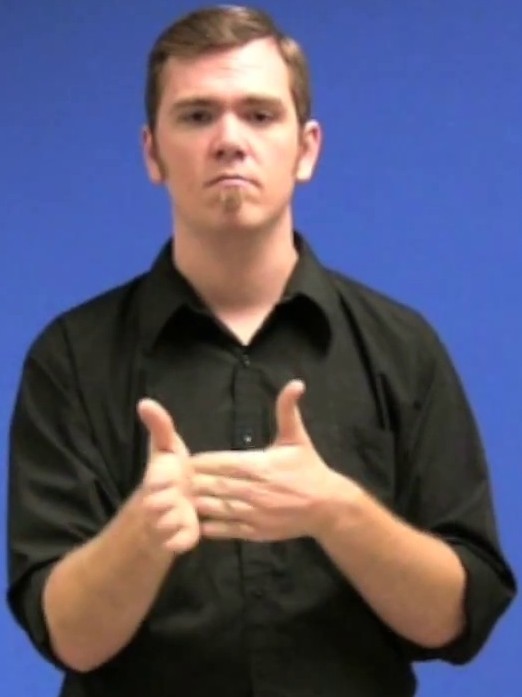}
\includegraphics[width=0.25\linewidth, height=1.5cm]{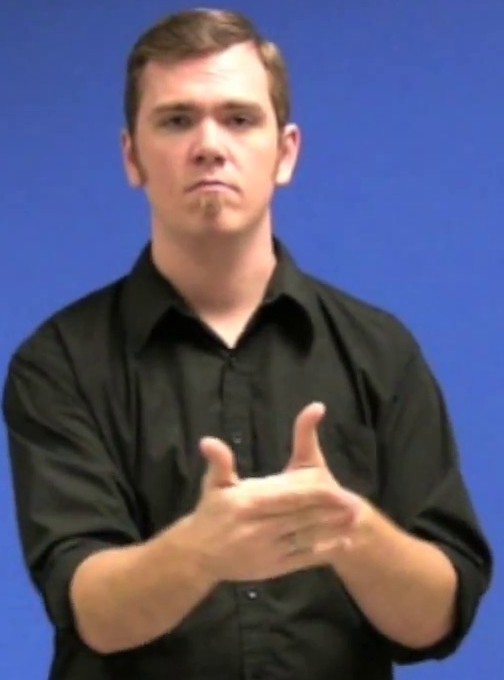}
\begin{minipage}{0.23\textwidth}
\vspace{1mm}
\tiny{While touching the wrist of the right open hand, palm facing left, with the extended left index finger, swing the right hand back and forth with a double movement.\\
\textbf{Handshape:}  \includegraphics[width=0.15\linewidth, height=0.5cm]{images/handshapes/open_B.png} \quad \quad \quad \quad \quad \quad Open-B hand\\
\textbf{Orientation:} \includegraphics[width=0.15\linewidth, height=0.5cm]{images/handshapes/palm_face-back.jpg} \includegraphics[width=0.15\linewidth, height=0.5cm]{images/handshapes/palm_left.png} \quad \quad Palm In and Palm Left \\
\textbf{Location:} \qquad \includegraphics[width=0.15\linewidth, height=0.5cm]{images/handshapes/neutral_space.png} \enspace \quad \quad \quad  \quad \quad Neutral \\
\textbf{Movement:} \quad \includegraphics[width=0.15\linewidth, height=0.5cm]{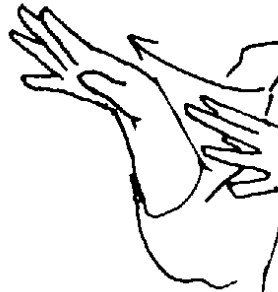} \includegraphics[width=0.15\linewidth, height=0.5cm]{images/handshapes/wrist_movement.png} \enspace Move Away and Wrist Movement \\
\textbf{Others:} No Repeat, Two-hand sign \\}
\end{minipage}\\
\bottomrule
\end{tabular}
\caption{Example sequences and corresponding textual descriptions from the ASL-Text (upper half) and MS-ZSSLR-W/C (bottom half) datasets. For visualization purposes, only the person regions of the videos are shown. }
\label{fig:dataset}
\vspace{-5mm}
\end{figure*}

\begin{table}[t]
{\centering
\caption{Statistics of the proposed ZSSLR benchmark datasets.}
\label{tab:statistics}
\scalebox{0.64}{\small{
\begin{tabular}{lcccccccc}
\toprule
Dataset & \makecell{Class \\ Count}  & \makecell{Train \\ Count} & \makecell{Val. \\ Count} & \makecell{Test \\ Count} & \makecell{Total \\ Count} & \makecell{Avg. per\\ Class}  & \makecell{Frame \\ Count }& \makecell{Con-\\trolled?} \\
\midrule
ASL-Text & 250 & 1188 & 151 & 259 & 1598 & 6.3 & 54,151 & \cmark \\
MS-ZSSLR-W  & 200 & 5862 & 1153 & 1846 & 8861 & 44.3 & 922,962 & \xmark \\
MS-ZSSLR-C  & 200 & 7303 & 1368 & 1779 & 10,450 & 52.2 & 688,118 & \xmark \\
\bottomrule
\end{tabular}
}}}
\end{table}

A summary of the dataset statistics is provided in Table~\ref{tab:statistics}. It can be seen that
ASL-Text has the highest class count ($250$) but the smallest number of samples per class on average
($6$). MS-ZSSLR-W/C datasets both have $200$ classes with $44$ and $52$ average sample counts per
class, respectively, which are approximately $7$ times higher than that of ASL-Text. Overall,
therefore, MS-ZSSLR-W/C provides a larger number of samples taken in an uncontrolled setting for
each class while ASL-Text provides fewer per-class samples taken in a laboratory environment, for a
relatively larger number of classes. The dataset splits are explained in Section~\ref{sec:implementationdetails}.

Figure~\ref{fig:dataset} presents example sign sample frames and the corresponding text and
attribute-based descriptions for the ASL-Text and MS-ZSSLR-W/C datasets. It can be observed how
textual descriptions and attributes can be complimentary to each other for characterizing signs in
terms of handshapes, hand locations, hand movements, palm orientations, combined with some
additional details. Our experimental results on ZSSLR and GZSSLR in Section~\ref{sec:exp} confirm this observation.

\subsection{Textual descriptions} 

We augment all datasets with the textual sign descriptions that are gathered from Webster American
Sign Language Dictionary \cite{costello1999random}.  The textual descriptions of the proposed
datasets include the detailed instructions of signs with an emphasis on four basic elements:
handshape ({\em A-hand}, {\em S-hand}, {\em 5-hand}, \etc),  the orientation of the palms ({\em forward,
backward, etc.}), movements of the hands ({\em right, left, etc.}), and the location of hands concerning to the body ({\em in front of the chest, each side of the body, right shoulder, etc.}). Some descriptions additionally include non-manual cues such as the facial expressions, head movement , and body posture. The average length of a textual description is 30 words with a vocabulary of 154
words and 29 words with a vocabulary of 274 words in the MS-ASL and MS-ZSSLR-W/C datasets, respectively.

Example textual descriptions can be found in Figure \ref{fig:dataset}. Note that hand shapes are described with a specialized vocabulary that involves the terms {\em F-hand}, {\em A-hand}, {\em S-hand}, {\em 5-hand}, {\em 8-hand}, {\em 10-hand}, {\em open-hand}, {\em bent-v hand}, {\em flattened-o hand} \cite{costello1999random}. 
From the example hand shapes shown in Figure \ref{fig:dataset}, it can be seen that the textual sign language descriptions are indeed quite indicative of the ongoing gesture. 

\subsection{Attribute descriptions}
\label{sec:attr_desc}
We further annotate the datasets with high-level attributes that are gathered from American Sign Language Hand Shape Dictionary \cite{tennant1998american}.
The attributes highlight four basic features; hand shape ({\tt A-hand}, {\tt S-hand}, {\tt 5-hand}, etc.), palm orientation ({\tt in, out, up, down, left, or right}), hand location ({\tt neutral, chest, on the shoulder, mouth/nose, on forehead/eyes, ear/temple}), and movement ({\tt up, down, left, right, inward, outward, circular, wrist movement, finger movement}). We additionally include two relevant attributes: {\tt repetition} indicates whether the hand movement repeats or not, and {\tt one-hand} indicates whether the sign is made by a single hand or both hands. In combination, we obtain a 53-dimensional attribute-based class representations. As can be seen in Figure~\ref{fig:dataset}, these selected attributes provide an expressive medium to represent sign characteristics. For instance, {\tt bicycle} sign is described to have {\tt S-hand} handshape with {\tt palms down}, {\tt natural} hand locations, and {\tt circular} movements. Additionally, {\tt circular} movement is repeated and both hands are used in the making of the sign.

\section{Methodology}
\label{sec:method}

In this section, we first give a formal definition of the problem and then explain the components of the proposed approach. Finally, we present our binary attribute analysis techniques. 

\newcommand{\domainv}{\mathbb{V}} 
\newcommand{\domaint}{\mathbb{T}} 
\newcommand{\clss}{\mathbb{C}_\text{s}} 
\newcommand{\clsu}{\mathbb{C}_\text{u}} 
\newcommand{\loss}{\ell} 

\myparagraphwithspace{Problem definition}
In ZSSLR, there are two sources of information: (i) the \textit{visual domain} $\domainv$, which consists of sign videos, and, (ii) the \textit{auxiliary information domain} $\domaint$, which includes the class embeddings based on textual descriptions and/or attributes.  At training time, the videos, labels and the sign descriptions are available only for the \textit{seen} classes, $\clss$. In ZSSLR, at test time, our goal is to correctly classify the examples of novel {\em unseen} classes, $\clsu$, which are distinct from the seen classes. GZSSLR is essentially the same as ZSSLR except that at test time, the goal expands to correctly classifying the novel examples of both {\em unseen} classes $\clsu$, and the {\em seen} classes $\clss$. In our explanations of the approach below, we focus on the ZSSLR case for simplicity.

The training set $S_{tr} = \{ (v_{i}, c_i )\}_{i=1}^{N}$ consists of $N$ samples where $v_{i}$ is the $i$-th training video and $c_i \in \clss$ is the corresponding sign class label. We assume that we have access to the textual and/or attribute descriptions of each class $c$, represented by $\tau(c)$. The goal is to learn a zero-shot classifier that can correctly assign  each test video to a class in $\clsu$, based on the auxiliary information. We aim to construct a label embedding based zero-shot classification model. For this purpose, we define a {\em compatibility function} $F(v,c)$ as a mapping from an input video and class pair to a score representing the confidence that the input video $v$ belongs to the class $c$. Given the compatibility function $F$, the test-time zero-shot classification function $f: \domainv \rightarrow \clsu$ is defined as:
\begin{equation}
f(v) = \argmax_{c \in \clsu} F(v,c) .
\label{eq:clsfunc}\end{equation}
In this way, we leverage the compatibility function to recognize novel signs at test time. In the case of GZSSLR, the $\argmax$ operator runs over $\clsu \cup \clss$ instead.

The performance of the resulting zero-shot sign recognition model directly depends on three factors: (i) video representation, (ii) class representation, and, (iii) the model used as the compatibility function $F$. An overview of our approach for addressing these issues is shown in Figure~\ref{fig:main}, and the details are presented in the following sections.

\begin{figure*}
\centering
\includegraphics[width=\textwidth]{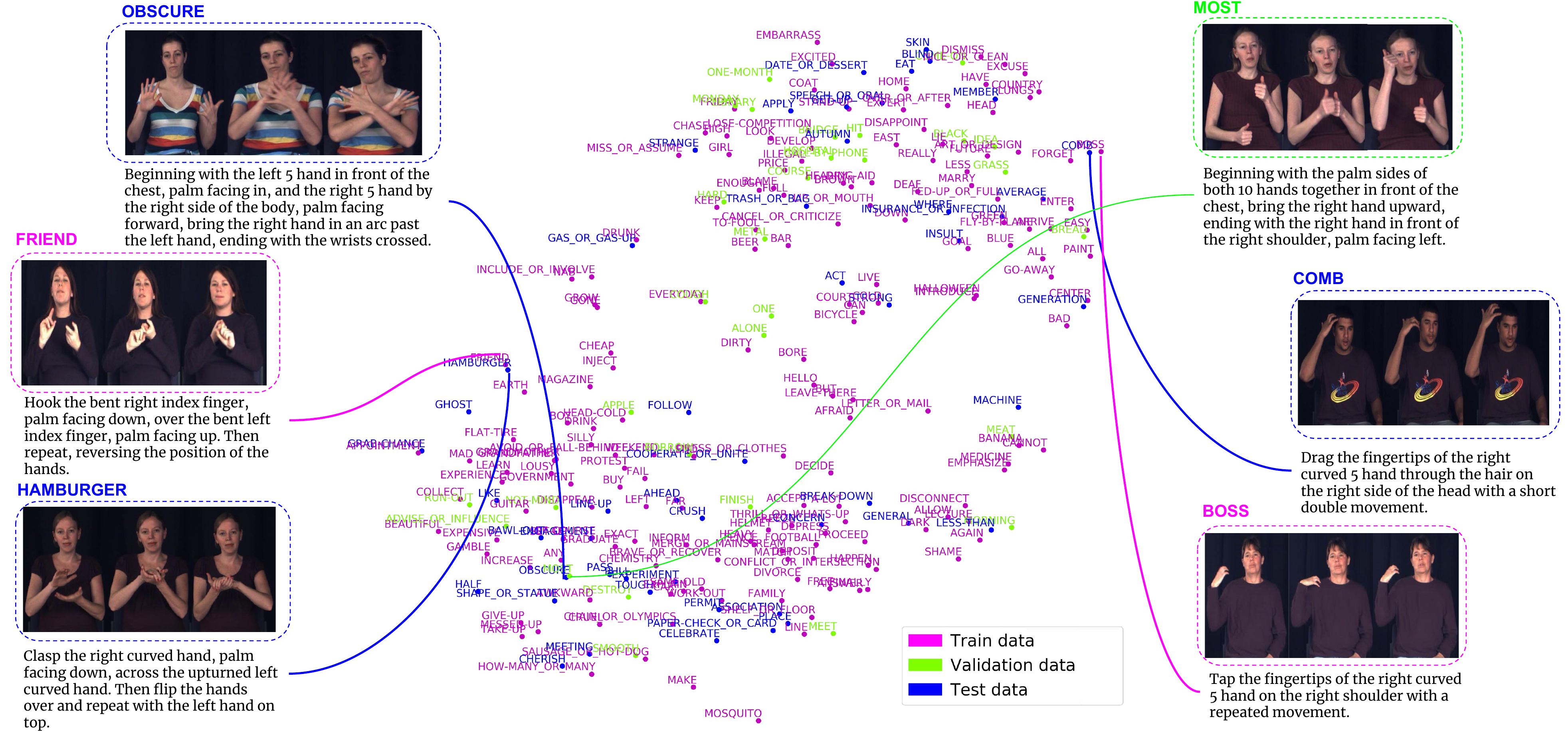}
\vspace{-.2cm}
\caption{t-SNE visualization of sign descriptions using BERT-\cite{devlin2018bert} embeddings. Nearby descriptions typically correspond to visually similar signs. Best viewed in color, with zoom.}
\label{fig:BERT}
\vspace{-.5cm}
\end{figure*}

\subsection{Spatiotemporal video embedding}
In this work, we explore two approaches for obtaining video representations. The first one is by extracting short-term spatiotemporal features using ConvNet features of the video snippets, and then capturing longer-term dynamics through recurrent models. The second approach is capturing spatiotemporal features through the recently proposed temporal shift module (TSM)\cite{lin2019tsm} in an end-to-end manner. We improve the video representation by extracting features in two separate streams: the full frames and hand regions.

\vspace{2mm}
\myparagraphwospace{Short-term spatiotemporal representation}
Short-term spatiotemporal representation is obtained by first splitting each video into 8 frames long snippets and then extracting their features using the pre-trained I3D model~\cite{carreira2017quo}, a state-of-the-art 
3D-ConvNet architecture. I3D model is obtained by adapting a pre-trained Inception model \cite{szegedy2015going} to the video domain and then fine-tuning on the Kinetics dataset. The final video representation is obtained by temporally average pooling the resulting snippet features.

\vspace{2mm}
\myparagraphwospace{Modeling longer-term dependencies}
Average pooling the 3D-CNN features is a well-performing technique for the recognition of non-complex (singleton) actions. Signs, on the contrary, tend to portray more complex 
patterns that are composed of sequences of multiple basic gestures. In order to capture the transition dynamics and longer-term dependencies across the snippets of a video, we use recurrent models that take an I3D representation sequence as input, and, provide an output embedding. For this purpose, we use a bidirectional LSTM (bi-LSTM) \cite{graves2005framewise} model, and, compare it against the average pooling, LSTM \cite{hochreiter1997long} and GRU \cite{cho2014learning} models.

\vspace{2mm}
\myparagraphwospace{Temporal shift module} Apart from modeling both short-term and longer-term dependencies in conjunction, we adapt Temporal Shift Module (TSM) \cite{lin2019tsm} to our proposed model as a spatiotemporal video representation module.  
TSM \cite{lin2019tsm} aims to achieve 3D CNN performance with 2D CNN complexity for large scale video understanding. The module basically shifts information on channels along the temporal dimension in forward and backward to aid information exchange with neighboring frames. For instance, a given 1-D input $X$ with infinite length and a kernel size of 3 with weights of $(w_1, w_2, w_3)$, the shift operation can be defined as:
\begin{equation}
    X_i^{-1} = X_{i-1}, X_i^0 = X_i, X_i^{+1}=X_{i+1}
\end{equation} by shifting the input by -1, 0, +1 in a bidirectional way. In order to get a spatiotemporal video representation, $\phi(v)$, with shifted input, the following multiply-accumulate operation is performed to acquire a visual representation;
\begin{equation}
    \phi(v) = w_1 \cdot X^{-1} + w_2 \cdot X^0 + w_3 \cdot X^{+1}
\end{equation}
Intuitively, the module first shifts the data and then perform 2D convolution operation. The advantage is that such a shift based approach can be incorporated into any 2D CNN model.

\vspace{2mm}
\myparagraphwospace{Two-stream video representation}
Hands play a central role in expressing signs. In order to encode details of the hand movement,
we detect and crop the hand regions using OpenPose \cite{cao2018openpose} and form a hand-only sequence
corresponding to each video snippet. As a result, two separate streams are defined, including either TSM\cite{lin2019tsm} or I3D and bi-LSTM networks, over whole video frames and hand regions. The resulting video representations coming from each stream are then concatenated to obtain the final video representation, as illustrated in Figure \ref{fig:main}. Note that, both streams are trained together with the compatibility function in an end-to-end fashion.

\subsection{Auxiliary knowledge modeling}
\label{sec:method:auxiliaryknowledge}
We use the auxiliary information needed for zero-shot recognition via two main sources; textual descriptions and attributes. The descriptions are both based on sign language dictionaries and include rich information about the signs. We explore the utilization of concatenated textual and attribute descriptions apart from individual usages of them.

\mypar{Text-based class embeddings} We extract contextualized language embeddings from textual sign descriptions using the state-of-the-art language representation model BERT \cite{devlin2018bert}. BERT architecture basically consists of a stack of encoders; specifically, multi-layer bidirectional transformers \cite{vaswani2017attention}. The model's main advantage over the word2vec~\cite{mikolov2013distributed} and glove \cite{pennington2014glove} representations is that BERT model is contextual and the extracted representations of the words change with respect to other words in a sentence. Figure \ref{fig:BERT} shows the t-SNE visualization of all sign class BERT embeddings. A close inspection to this feature space reveals that classes that appear closer in t-SNE embeddings have indeed similar descriptions.
For instance, \textit{friend} and \textit{hamburger} signs are composed of similar motions with different handshapes, \textit{obscure} and \textit{most} signs have similar hand movements but different hand shapes and directions,
and, \textit{comb} and \textit{boss} signs include the same repeated movement with different handshape and locations with respect to the body.

\mypar{Attribute-based class representation} As the attribute based class representations, we use binary attributes such that each class $c$ has fixed-length binary attribute vectors; $\alpha(c) = [a_1^{c}, a_2^{c}, ..., a_{53}^{c}]$ to describe the classes. These attributes are defined based on a handshape dictionary for each sign class that emphasizes four points; handshapes, hand location, palm orientation, hand movement to describe sign making. Figure \ref{fig:dataset} presents sample attribute based representations. To the best of our knowledge, this is the first study that uses attributes as a common space for sign language recognition.

We also utilize both attributes and textual descriptions in conjunction as the auxiliary information source.
\begin{finalquote} To have a more compact representation, we first transform $768$ dimensional textual feature vectors into a $d_t$-dimensional embedding space using a linear layer.  We then concatenate the $53$ dimensional attribute and $d_t$ dimensional textual feature vectors into $k+53$ dimensional vectors to represent sign classes. We fix $d_t=64$ based on the validation set results of ASL-Text dataset. In Section \ref{sec:expRes}, we evaluate the corresponding ablation studies using these different information sources and provide details of the $d_t$ tuning experiments. \end{finalquote} 

\subsection{Zero-shot learning model}
In our work, we adapt a label embedding \cite{akata2013label,weston2010large} based formulation to tackle the ZSSLR and GZSSLR problems. More specifically, we use a bi-linear compatibility function that evaluates a given pair of video and class representations:
\begin{equation}
    F(v,c) = \phi(v)^T W \rho(c)
\end{equation}
where $\phi(v)$ is the $d$-dimensional embedding of the video $v$, $W$ is the $d{\times}t$ compatibility matrix, and $\rho(c)$ is the $t$-dimensional class embedding. The class embedding
can be the per-class attribute vector, reduced BERT embedding or their concatenation.

In order to learn this compatibility matrix, we use cross entropy loss with $\ell_2$-regularization:
\vspace{-1mm}
\begin{equation}
    \min_{W} - \dfrac{1}{N} \sum_{i=1}^{N} \log \frac{\exp \{ F(v_i,c_i) \} }{ \sum_{c_j \in \clss} \exp \{ F(v_i,c_j) \} } + \lambda \| W \|^2
\end{equation}

\noindent where $\lambda$ is the regularization weight. This core formulation is also used in \cite{sumbul2018fine}, in a completely different ZSL problem. Since the objective function is analogous to the logistic regression classifier, we refer to this approach as {\em logistic label embedding} (LLE).

In addition to LLE, we also adapt the {\em embarrassingly simple zero-shot learning} (ESZSL) \cite{romera2015embarrassingly} and {\em semantic auto-encoder} (SAE) \cite{kodirov2017semantic} formulations as baselines. \begin{finalquote} The ESZSL formulation utilizes a regression-based loss function in combination with multiple ZSL-specific regularization terms to learn a bi-linear compatibility model between the image and class embeddings. The SAE approach embraces an auto-encoding principle that aims to project image embeddings to the semantic space, and reconstruct back from it. Compared to these two alternatives, LLE takes a more discriminative learning approach simply based on regularized cross-entropy loss over the compatibility scores. \end{finalquote}

\subsection{Analysing binary attribute based class definitions}
\label{sec:methodanalysis}

In our experimental results, presented in the next section, we observe that the auxiliary knowledge
based class representations greatly affect the ZSSLR results. To better understand the role of
auxiliary knowledge in ZSSLR, binary attributes are particularly valuable for being visually and semantically
distinct entities~\cite{ferrari2007learning}. 
\begin{finalquote}
We note that the case of using dictionary-based binary attributes in ZSSLR differs from the commonly studied case in zero-shot image classification benchmarks, \eg aPY, AWA1, AWA2, CUB and SUN datasets~\cite{xian2017zero}, where the class attributes are typically continuous and obtained by averaging per-image manual annotations for experimental purposes. In contrast, we focus on a more realistic setting where the attribute annotations are derived directly from the canonical definitions of signs, without relying on visual sample annotations (of unseen classes). 
\end{finalquote}
To this end, we aim to evaluate how associating (or not) a class with a binary attribute affects the 
unseen class predictions. For this purpose, below, we present two
techniques 
for analysing correct and incorrect zero-shot recognition results. 

\mypar{Analysing correct confidence scores} On correct predictions, we aim to 
understand the relation between the confidence scores and the binary attribute based definitions of classes.
Had attributes been continuous entities, we could straightforwardly evaluate the impact of the relation between the $k$-th attribute and $c$-th class in terms of partial derivatives:
\begin{equation}
    \frac{d p(c|v)}{d a_{c,k}}
\end{equation} 
where $a_{c,k}$ is the representation of class $c$ in terms of the $k$-th attribute and $p(c|v)$ is the posterior probability of class $c$ for the input $v$. We note that
$F$, used in the definition of $p(c|v)$, depends on $a_{c,k}$ through the use of class embedding function $\rho(c)$.
However, since
we are interested in binary attribute definitions of sign classes, we cannot directly 
compute partial derivatives. As a remedy, we define the {\em flip-difference operator} $\nabla_{c,k}$,
analogous to the backward difference based derivative approximations:
\begin{equation}
    \nabla_{c,k}[p(v|c)] = p(c|v) - p(c|v;\overline{a_{c,k}}) 
    \label{eq:flip_prob}
\end{equation}
where we use the notation $p(c|v;\overline{a_{c,k}})$ to express posterior probability of class $c$ for input $v$ when the $k$-th attribute of the $c$-th class is flipped (positive to negative, or vice versa) at inference time. 

Intuitively, we expect a significant drop in posterior probability when an informative attribute is
flipped. Therefore, a relatively large $\nabla_{c,k}[p(v|c)]$ value can be interpreted as the higher importance
of the corresponding class attribute relation for some particular input $v$. 
To estimate the class-level importance,
we take the correctly classified test examples $y$ of some unseen class $c$, and, average the result
of Eq.~\ref{eq:flip_prob} over the test samples for each attribute separately, resulting in
estimated importance scores of attributes in making correct zero-shot predictions.

\mypar{Analysing zero-shot misclassifications} To better understand incorrect zero-shot predictions,
we again use a flip-difference operator based approach to estimate the effect of each attribute. 
However, in this case instead of simply computing the influence of attributes on posterior probabilities (of wrongly
predicted classes), we want to more explicitly answer the question why some other class is being predicted 
instead of the correct one. For this purpose, we focus on log-derivative of class posterior probabilities 
given by the ZSL model. More specifically, analogous to partial derivative of log ratios with 
respect to attributes, we apply the flip-difference operator on the ratios:
\begin{equation}
    \nabla_{c,k}[r(c^\star,c^\text{o},v)] = r(c^\star,c^\text{o},v) - r(c^\star,c^\text{o},v;\overline{a_{c^\star,k}})
    \label{eq:flip_ratio}
\end{equation}
where $r(c^\star,c^\text{o},v)$ denotes the log-ratio of some (incorrect) class $c^\star$ probability over the
correct class $c^\text{o}$ for the input $v$:
\begin{equation}
    r(c^\star,c^\text{o},v) = \log \frac{p(c^\star|v)}{p(c^\text{o}|v)} .
\end{equation}
Similarly, $r(c^\star,c^\text{o},v;\overline{a_{c^\star,k}})$ denotes the log ratio when posterior probabilities are
computed by flipping the k-th attribute in the definition of the $c$-th class:
\begin{equation}
    r(c^\star,c^\text{o},v;\overline{a_{c^\star,k}}) = \log \frac{p(c^\star|v;\overline{a_{c^\star,k}})}{p(c^\text{o}|v;\overline{a_{c^\star,k}})} .
\end{equation}
Overall, by definition, a large drop in log-ratio value indicates that the probability will shift more quickly from the
incorrectly predicted class to the correct one. Therefore, larger influence values in this definition can be interpreted as the 
dominant sources of corresponding misclassifications. To obtain confusion-level attribute influence scores, we find the most commonly occurring 
class confusions, and average the values obtained via Eq.~\ref{eq:flip_ratio} over the test samples of each
predicted and ground-truth class tuple.

\section{Experiments}
\label{sec:exp}

In this section, we first describe the implementation details and the experimental setup.~\footnote{Datasets and source code is available at \url{https://ycbilge.github.io/towardszslsign.html}.} Then we present and discuss our experimental results on (G)ZSSLR. Finally, we present our attribute analysis results in Section~\ref{sub:attr_analysis}. 

\begin{table}[t]
\centering
\caption{Comparison of ZSL models on {\em ASL-Text}. I3D \cite{carreira2017quo} features are extracted over the whole frames (\ie body stream) only. \label{tab:zsl_comparison}}
\footnotesize{
\begin{tabular}{llllll}
\toprule
Method & visual rep. &Val (30 Classes) & \multicolumn{3}{c}{Test (50 Classes)} \\
       & &  top- 1                 & top-1 & top-2 & top-5 \\
\midrule
Random & - & 3.3                    &  2.0 & 4.0 & 10.0  \\ \hline
SAE  & body &10.6 &  8.0 & 12.0 & 16.0  \\ 
ESZSL  & body & 12.0  & \textbf{16.9} & \textbf{26.0} & \textbf{44.4}  \\ 
LLE      & body & \textbf{14.1} & 11.4 & 21.2 & 41.1 \\ 
\bottomrule
\end{tabular}
}
\end{table}

\subsection{Experimental setup and implementation details}
\label{sec:implementationdetails}

We evaluate our models on the proposed zero-shot sign language datasets; ASL-Text, MS-ZSSLR-W, and
MS-ZSSLR-C. In our experiments, we fix the number of video frames of each sign video to 32.  For
LSTMs, we extract 1024-d features from the last average pooling layer of the I3D model  using a
stride of 4 using every consecutive 8 frames. When modeling the longer temporal context, we set
LSTM's or bi-LSTM's initial hidden and cell states to the average pool of each sequence during training.

In this work, we opt to utilize a bidirectional-TSM module with the ResNet-34 backbone \cite{he2016deep}
pretrained on ImageNet. We utilize TSM architecture by removing its last fully-connected layer and
use the output of its preceding the average-pooling layer as the video representation.  One of the
trade-offs of TSM is that data movement highly increases memory usage, and therefore, we reduce each
video sample frame count to 4 frames.  We train the TSM model in conjunction with our label
embedding approach in an end-to-end manner. 

As our text representation, we use the $\mathrm{BERT_{BASE}}$ model \cite{devlin2018bert} and
extract 768-dimensional sentence-based features. Following the description in \cite{devlin2018bert},
we concatenate the features from the last four layers of the pretrained Transformer of
$\mathrm{BERT_{BASE}}$ and $l_2$-normalize them. 

We evaluate the models in both the zero-shot learning (ZSL) and generalized zero-shot learning
(GZSL) settings. To create dataset splits, we use the largest classes, ranked by the number of
in-class samples, for training and the smallest ones for testing. In the ZSL setting, we use $170$
classes of ASL-Text and $120$ classes of MS-ZSSLR-W/C as the training classes. In all three
datasets, the validation and test sets contain $30$ and $50$ classes, respectively. In
the GZSL setting of ASL-Text, train, validation, and test splits contain $170$, $200$ and $250$
classes, respectively. Similarly, the train, validation, and test splits of $MS-ZSSLR-W/C$ datasets
contain $120$, $150$, and $200$ classes in the GZSL setting, respectively.

We repeat each experiment $5$ times and report the average results.  We use top-n accuracy scores
normalized by the class sizes. We compute random baseline scores based on 10000 random prediction trials.

\subsection{Experimental results}
\label{sec:expRes}

In the following, we first present our main ZSSLR results, including a through evaluation with
ablative studies on the ASL-Text and MS-ZSSLR-C datasets. In Section~\ref{sub:ms_zsslrwc_exp}, we
then present additional results on the use of {\em in-the-wild} benchmark MS-ZSSLR-W(ild) for test-only
or train and test purposes, with comparisons to the results on the cleaned version MS-ZSSLR-C(lean).
Finally, in Section~\ref{sub:gzssslr_exp}, we present our generalized zero-shot learning results.

\subsubsection{Main results and ablative studies}
\label{sub:asl-text_exp}

\myparagraphwospace{ZSL formulation} We evaluate three simple and widely used ZSL formulations, namely
SAE~\cite{kodirov2017semantic}, ESZSL~\cite{romera2015embarrassingly}, and LLE.
In this comparison, for simplicity, we use the average pooled 3D-CNN features of
complete frames only as our the video representation and use the ASL-Text dataset. 
We compare methods in terms of top-1 validation accuracy and top-1, top-2, and top-5 test
accuracy scores. 

The results are presented in Table~\ref{tab:zsl_comparison}. We observe that the best performing 
formulations on the validation and test sets are LLE and ESZSL, respectively. SAE performs worse 
in all metrics, which can possibly be explained by the fact that SAE incorporates an auto-encoding loss function that aims to reconstruct from video to semantic space with the purpose of reconstructing
back from semantic space to video. Such a reconstruction based formulation possibly behaves sub-optimally 
in the case of high in-class variance and few per-class train samples. Based on these results, 
we drop SAE from our following experiments.

\begin{table}[t]
\centering
\caption{Evaluation of two-stream spatiotemporal representation on ASL-Text.
\textit{body} denotes full frame stream, \textit{hand} denotes estimated hand stream.
    Average pooling is used for aggregating short-term video representations. \label{tab:two_stream_asl}}
\footnotesize{
\begin{tabular}{llllll}
\toprule
Method & visual rep. & Val (30 Classes) & \multicolumn{3}{c}{Test (50 Classes)} \\
       &             & top- 1                 & top-1 & top-2 & top-5 \\
\midrule
Random & - & 3.3                    &  2.0 & 4.0 & 10.0  \\
\midrule
\multirow{3}{*}{ESZSL} &  body  & 12.0  & 16.9 & 26.0 & \textbf{44.4} \\
                            & hand & 13.3  & 11.6 & 19.6 & 33.7 \\
                           & body + hand & 14.6  &  17.1 & 25.7 & 43.0  \\ 
\midrule
\multirow{3}{*}{LLE} & body  & 14.1   & 11.4 & 21.2 & 41.1 \\
                            & hand  &  15.0  & 12.6 & 19.8 & 37.8 \\
                           & body + hand  &  \textbf{16.2}   &  \textbf{18.0} & \textbf{27.4} & 43.8 \\ 
\bottomrule
\end{tabular}
}
\end{table}

\mypar{Two stream representation} In our next experiments, we introduce the second stream based on estimated hand positions and evaluate the effect of using a two-stream representation.
For this purpose, we compare three representations obtained by using 
\textit{body}, \ie, the full-frame input stream, \textit{hand}, or both.
The results on the ASL-Text dataset given in Table~\ref{tab:two_stream_asl} show that 
using the two streams in conjunction improves the top-1 scores on validation
and test sets for both ESZSL and LLE formulations. We also observe that LLE outperforms ESZSL 
using the two-stream representation consistently in all metrics. Due to its greater
validation set performance, we fix our ZSL formulation to LLE in the remainder of our experiments.

\begin{finalquote} In Table~\ref{tab:two_stream_asl}, we also observe that using the body-only representation, while LLE performs better on the validation set, ESZSL outperforms on the test set. This is in contrast to the consistently better performance of LLE over the combined body and hand representation. We believe that this is primarily due to the nature of ZSSLR, where the characteristics of the sign classes can greatly vary between the train, validation, and test sets, which consist of mutually exclusive classes. Despite such performance fluctuations across the ZSL formulations, we, in principle, expect higher correct recognition rates over richer visual representations due to fine-grained distinctions across many sign classes. For example, \texttt{friend} and \texttt{hamburger} signs are very similar in terms of their overall motion; the hand shape is the only factor that differentiates these two signs. In this respect, richer visual representation based on separate body and hand encoding is expected to be beneficial, which may also contribute to more stable ZSL results. \end{finalquote} 

\begin{table}[t]
\centering
\caption{Evaluation of two-stream spatiotemporal representation on MS-ZSSLR-C. 
\textit{body} denotes full frame stream, \textit{hand} denotes estimated hand stream.
Average pooling is used for aggregating short-term video representations.
\label{fig:twostream_ms_c}}
\footnotesize{
\begin{tabular}{llllll}
\toprule
Method & visual rep.  & Val (30 Classes) & \multicolumn{3}{c}{Test (50 Classes)}  \\
       &     & top - 1    & top-1 & top-2 & top-5   \\
\midrule
Random  & \hspace{0.2cm} -   &  3.3  & 2.0 & 4.0 & 10.0  \\ \hline
\multirow{3}{*}{LLE} & body  & 10.8   & 5.54 & 9.55 & 19.5 \\
                            & hand  &  10.0  & 4.05 & 7.12 & 15.7 \\
                           & body + hand  &   \textbf{11.7 } & \textbf{5.74} & \textbf{10.0} & \textbf{19.7} \\ 

\bottomrule
\end{tabular}
}
\end{table}

In Table \ref{fig:twostream_ms_c}, we present the evaluation of two-steam representation on the
MS-ZSSLR-C dataset. The results consistently show that using body and hand stream representations
jointly improves the results over body-only and hand-only results, on both validation and test sets.
Overall, the results on both datasets confirm that hand and body streams provide complementary
information for zero-shot recognition of signs.

\begin{table}[t]
\centering
\caption{Comparison of different temporal units with LLE method on ASL-Text and MS-ZSSLR-C datasets. Body and hand streams are used in conjunction, and the textual descriptions are used as auxiliary data.}
\label{tab:temporal_rep_comparison}
\small{
\scalebox{0.92}{
\begin{tabular}{lllllllll}
\toprule
\multirow{2}{*}{\shortstack[l]{Temporal}} &  \multicolumn{3}{c}{\textit{ASL-Text Dataset}} & \multicolumn{3}{c}{\textit{MS-ZSSLR-C Dataset}}\\
      & top-1 & top-2 & top-5 &top-1 & top-2 & top-5\\
\midrule
 AvePool   &  18.0  & 27.4  & 43.8 & 5.74 & 10.0 & 19.7  \\
  LSTM     &  18.2 & 28  & 47.2 & 6.02 & 10.1 & 19.7  \\
  GRU   & 19.7  & 31.8  & 50.0 & 6.04 & 10.5 & 19.2 \\
  bi-LSTM    &  \textbf{20.9} & \textbf{32.5}  & \textbf{51.4} & 6.65 & 10.4 & 19.9
 \\
  TSM   & 16.8 & 23.2 &  38.2     & \textbf{7.01} & \textbf{11.5} & \textbf{22.1} \\
\bottomrule
\end{tabular}
}}
\end{table}

\mypar{Temporal models} We further evaluate the effect of sequential temporal modeling using
recurrent architectures and the TSM model~\cite{lin2019tsm}, and compare to the simple temporal
average pooling scheme. As the recurrent model alternatives, we experiment with
LSTM\cite{hochreiter1997long}, GRU\cite{cho2014learning}, and bi-LSTM\cite{graves2005framewise} based
temporal models. For all cases, we use both hand and body video streams, textual description based
class embeddings, and the LLE model.

The results for both ASL-Text and MS-ZSSLR-C are presented in
Table~\ref{tab:temporal_rep_comparison}. Overall, we observe that, compared to temporal average
pooling, the framework benefits from the introduction of explicit temporal models.  On ASL-Text, the
best performing model is bi-LSTM, and TSM-based temporal model does not generalize well. In
contrast, on MS-ZSSLR-C, we observe that TSM performs the best in all metrics. \begin{finalquote} These results suggest
that TSM performs better in the presence of a larger training set, 
which is probably related to the fact that the TSM representation is learned in an end-to-end fashion, without utilizing a pre-trained I3D model as in bi-LSTM. \end{finalquote}

Our overall proposed
framework reaches a top-1 normalized accuracy of $20.9$ and $7.01$, and top-5 normalized accuracy of
51.4 and 19.9 on the test set of the ASL-Text and MS-ZSSLR-C datasets, respectively. 

\begin{table}[t]
\centering
    \caption{Evaluation of auxiliary knowledge sources on ASL-Text and MS-ZSSLR-C datasets, using LLE model with body and hand streams.\label{tab:attr_text_comparison}}
\small{
\scalebox{0.8}{
\begin{tabular}{llllllllll}
\toprule
\multirow{2}{*}{\shortstack[l]{Temporal}} & \multirow{2}{*}{\shortstack[l]{Auxiliary\\knowledge}} & \multicolumn{3}{c}{\textbf{ASL-Text}} & \multicolumn{3}{c}{\textbf{MS-ZSSLR-C}}\\

     &  & top-1 & top-2 & top-5 &top-1 & top-2 & top-5\\
\midrule
  \multirow{3}{*}{bi-LSTM}  & Text &  20.9 & 32.5  & 51.4 & 6.65 & 10.4 & 19.9 \\
   & Attr & 23.7 & 38.6 & 59.2 & 12.9 & 22.4 & 41.8\\
  & Attr + Text & \textbf{31.3} & \textbf{46.8} & \textbf{66.0} &  13.5   & 24.1 & \textbf{45.2} \\
  \midrule
   \multirow{3}{*}{TSM}    & Text & 16.8 & 23.2 &  38.2     & 7.01 & 11.5 &22.1 \\
       & Attr & 21.8 & 32.2 &  55.7     & 13.1 & 23.3 & 41.4 \\
    & Attr + Text & 25.6 & 37.8 & 57.8 & \textbf{14.7} & \textbf{24.6} & 43.4\\

\bottomrule
\end{tabular}
}}
\end{table}

\mypar{Auxiliary class knowledge} In our final analysis, we explore the options for auxiliary class
knowledge sources. We evaluate models using textual descriptions, attributes, or both jointly as
class embeddings. We evaluate each option with bi-LSTM and
TSM temporal models, following our observations in the preceding experiments. 

\begin{figure} 
\centering
   \includegraphics[width=\linewidth]{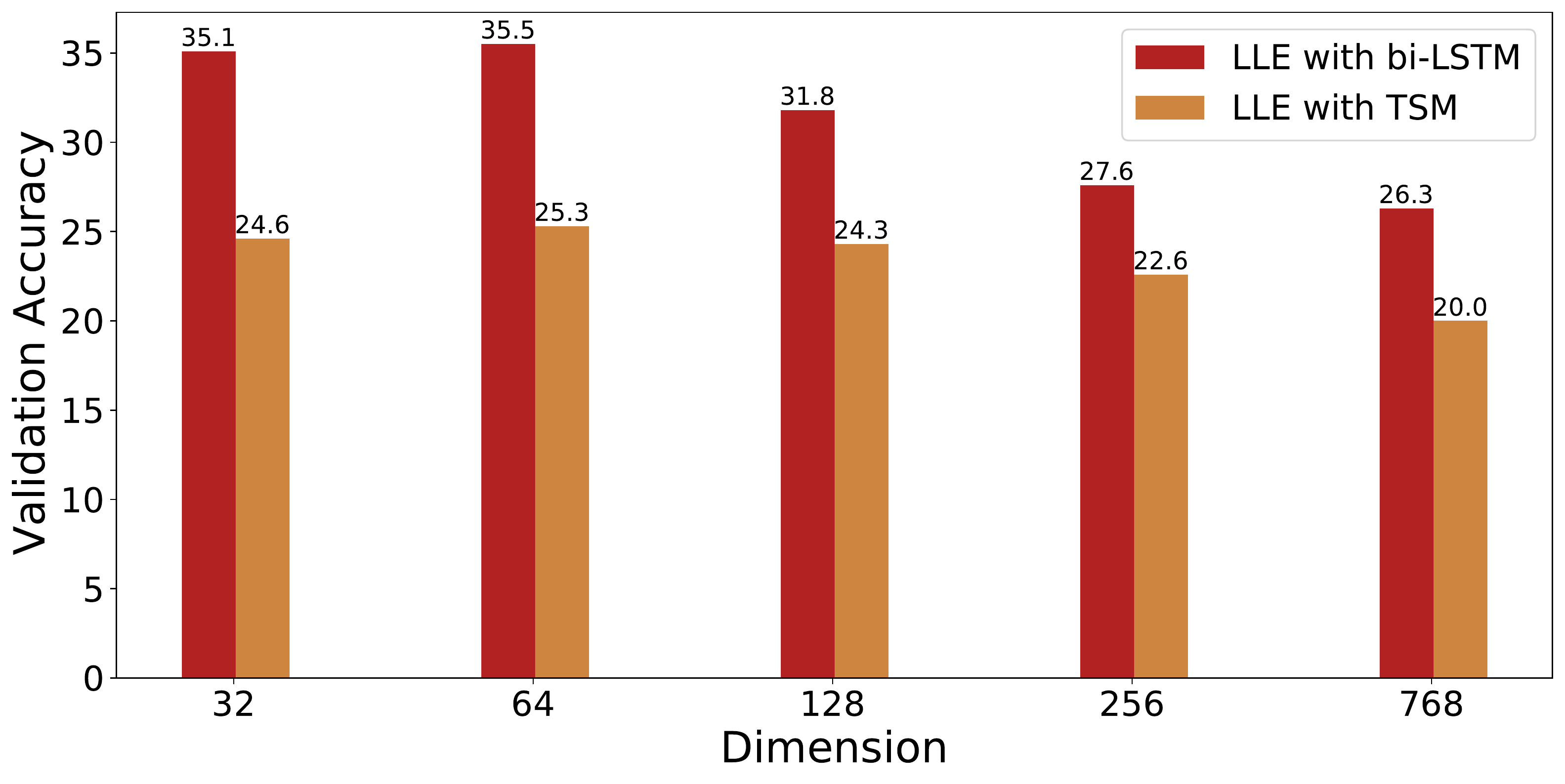}
    \caption{ASL-Text validation set accuracy as a function of textual embedding dimensionality, in the case of combined text and attribute class embeddings. The $768$ bin corresponds using the original textual embeddings, without applying a reduction layer. \label{fig:dim-reduction-val-set}}
\end{figure}

The empirical comparison of class embedding options is presented in
Table~\ref{tab:attr_text_comparison}.  We first observe that attribute embeddings perform
significantly better than the textual descriptions, for both bi-LSTM and TSM based results on
both datasets, in all metrics.  We also observe that using attributes and textual descriptions
jointly improves attribute-only and text-only results significantly in all cases, reaching a top-1
score of $31.3$ on ASL-TEXT using bi-LSTM and $14.7$ on MS-ZSSLR-C using TSM. Similarly, using joint
embeddings instead of text-only embeddings, the best top-5 scores impressively improves from $51.4$
to $66.0$ on ASL-Text and $19.9$ to $45.2$ on MS-ZSSLR-C. 

These results indicate that the sign class attributes that we define, carry valuable information and
two auxiliary knowledge sources are complementary to each other. The results also show the central
importance of class embeddings in zero-shot recognition. Therefore, a promising research direction,
in addition to representation and recognition modeling, can be the exploration of new ways for
obtaining and leveraging auxiliary knowledge sources.

\begin{figure}[!t]
\centering
\scalebox{0.9}{
\begin{tabular}{>{\raggedright\arraybackslash}m{0.35\linewidth}>{\raggedright\arraybackslash}m{0.6\linewidth}}
\toprule
\textit{ASL-Text} & \\
\includegraphics[width=0.3\linewidth, height=1.5cm]{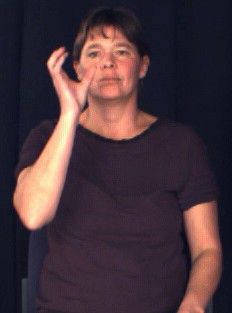}
\includegraphics[width=0.3\linewidth, height=1.5cm]{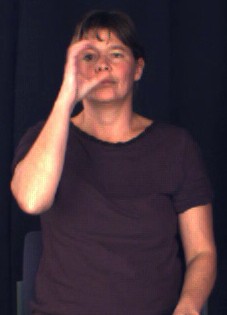}
\includegraphics[width=0.3\linewidth, height=1.5cm]{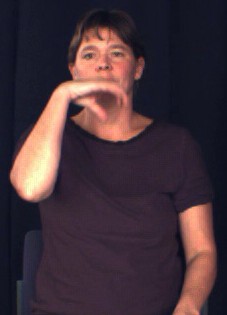} 
&
\makecell[t]{\begin{minipage}{0.3\textwidth}
\scriptsize{\vspace{-3mm}\textbf{Correctly Predicted Label: STRANGE}}\\
\scriptsize{Move the right \textcolor{red}{C} hand from near the right side of the face, palm facing left, downward in an arc in front of the face, ending near the left side of the chin, palm facing down.}
\end{minipage}} \\


\includegraphics[width=0.3\linewidth, height=1.5cm]{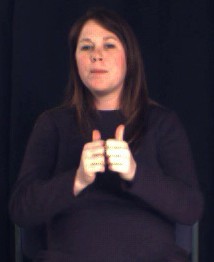}
\includegraphics[width=0.3\linewidth, height=1.5cm]{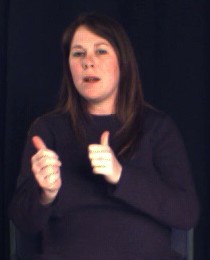}
\includegraphics[width=0.3\linewidth, height=1.5cm]{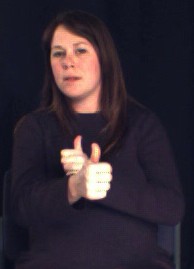} 
 & 
\makecell[t]{\begin{minipage}{0.3\textwidth}
    \scriptsize{\vspace{-2mm}\textbf{Correctly Predicted Label: AHEAD}\\
Beginning with the palm sides of both \textcolor{red}{A} hands together, move the right hand forward in a small arc.}
\end{minipage}}  \\


\vspace{1mm}
MS-ZSSLR-C & \\
\vspace{-4mm}
\includegraphics[width=0.3\linewidth,height=1.5cm]{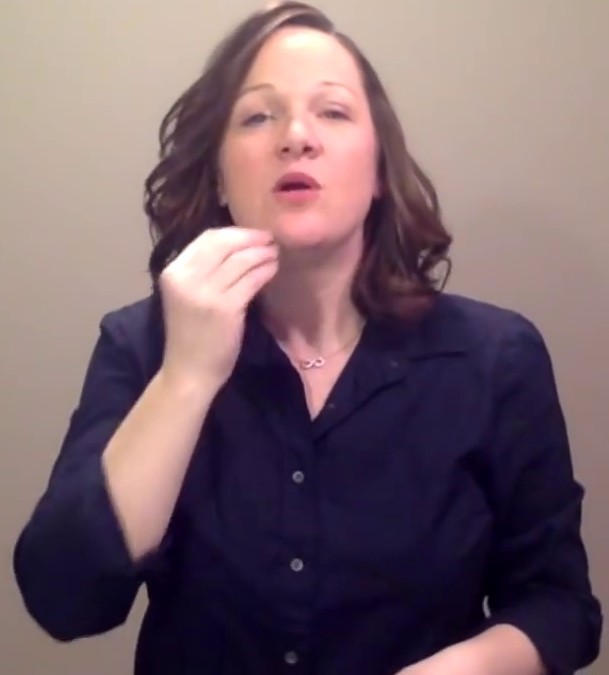}
\includegraphics[width=0.3\linewidth,height=1.5cm]{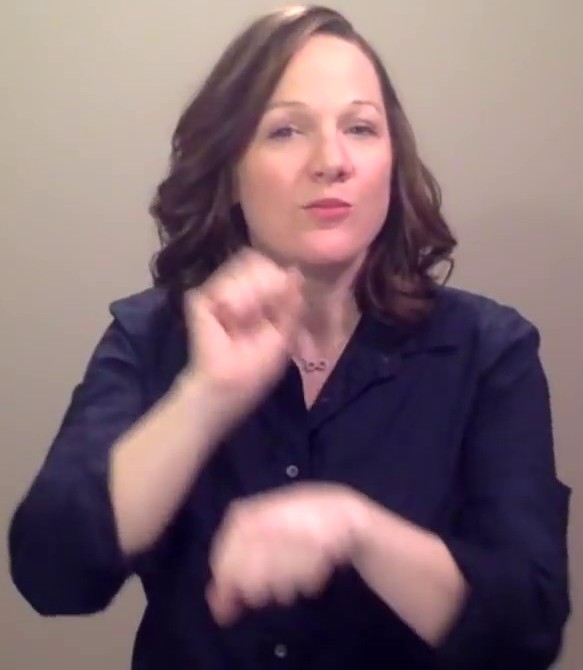}
\includegraphics[width=0.3\linewidth,height=1.5cm]{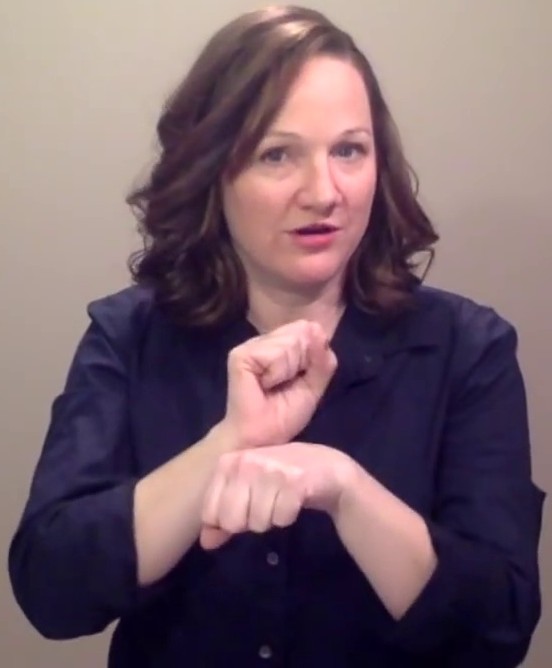} 
 & 
\makecell[t]{\begin{minipage}{0.3\textwidth}
\scriptsize{\vspace{-4mm}\textbf{Correctly Predicted Label: HOMEWORK}\\
Touch the fingertips of the right flattened  \textcolor{red}{O} hand to the right cheek, palm facing down. Then move the right hand down while changing into an  \textcolor{red}{S} hand and tap the base of the right S hand on the back of the left S hand held in front of the chest, palm facing down.}
\end{minipage}} \\


\includegraphics[width=0.3\linewidth,height=1.5cm]{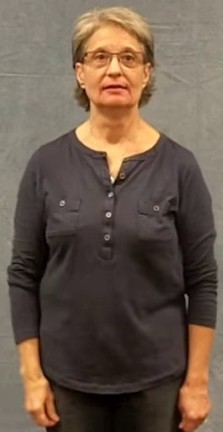}
\includegraphics[width=0.3\linewidth,height=1.5cm]{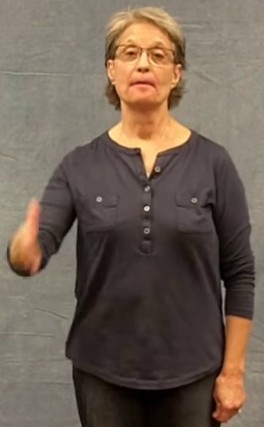}
\includegraphics[width=0.3\linewidth,height=1.5cm]{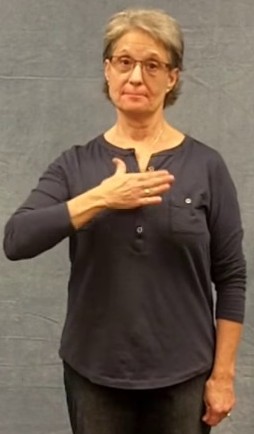} 
 & 
\makecell[t]{\begin{minipage}{0.3\textwidth}
\scriptsize{\vspace{-2mm}\textbf{Correctly Predicted Label: MY}\\
Place the palm of the right \textcolor{red}{open} hand on the chest, fingers pointing left.}
\end{minipage}} \\

\midrule

ASL-Text \\
\includegraphics[width=0.3\linewidth,height=1.5cm]{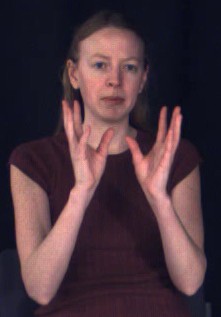}
\includegraphics[width=0.3\linewidth,height=1.5cm]{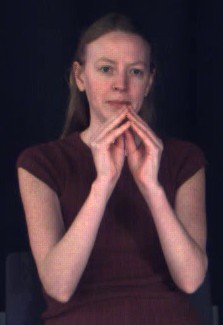}
\includegraphics[width=0.3\linewidth,height=1.5cm]{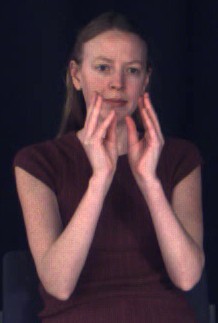}
 & 
\begin{minipage}{0.3\textwidth}
\scriptsize{\textbf{Predicted Label: BREAK-DOWN} \\
Beginning with the fingertips of both curved \textcolor{red}{5} hands touching in front of the chest, palms facing each other, allow the fingers to loosely drop, ending with the palms facing down.\\
\textbf{Correct Label: MEETING} \\
Beginning with both open hands in front of the chest, palms facing each other and fingers pointing up, close the fingers with a double movement into flattened \textcolor{red}{O} hands while moving the hands together.
}\end{minipage} \\

MS-ZSSLR-C \\
\includegraphics[width=0.3\linewidth,height=1.5cm]{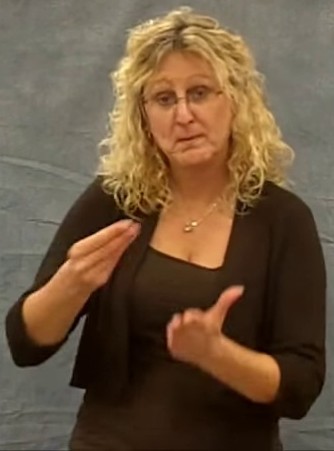}
\includegraphics[width=0.3\linewidth,height=1.5cm]{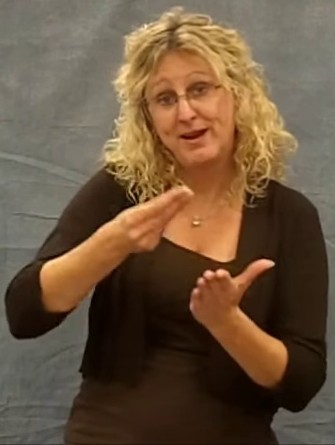}
\includegraphics[width=0.3\linewidth,height=1.5cm]{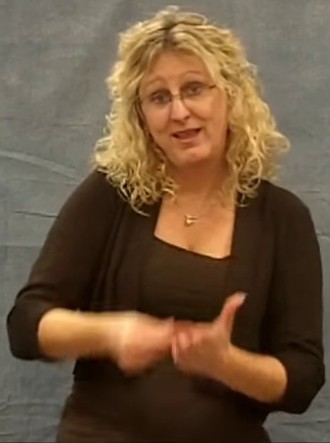}
& 
\begin{minipage}{0.3\textwidth}
\scriptsize{\textbf{Predicted Label: MONEY} \\
Tap the back of the right flattened \textcolor{red}{O} hand, palm facing up, with a double movement against the palm of the left open hand, palm facing up.\\
\textbf{Correct Label: COOKIE} \\
Touch the fingertips of the right \textcolor{red}{C} hand, palm facing down, on the upturned palm of the left open hand. Then twist the right hand and touch the left palm again.
}
\end{minipage} 
\\
\bottomrule
\end{tabular}}
\caption{Correct (first four rows) and incorrect (last two rows) zero-shot prediction examples on the ASL-Text and MS-ZSSLR-C datasets. The textual descriptions of the predicted and ground-truth (separately for incorrect predictions) classes are shown for each case. \label{fig:prediction}}
\vspace{-5mm}
\end{figure}

\begin{table*}
\begin{center}
\caption{Evaluation over the MS-ZSSLR-C and MS-ZSSLR-W dataset combinations.}
\label{fig:datasetsettings} 
\vspace{-3mm}
\label{avgpoolvsatp}
\resizebox{\textwidth}{!}{%
\begin{tabular}{p{1cm}p{1cm}p{1cm}p{1cm}p{1cm}p{1cm}p{1cm}p{1cm}p{1cm}p{1cm}p{1cm}p{1cm}}
\toprule
\multicolumn{3}{c}{MS-ZSSLR-C} & \multicolumn{3}{c}{MS-ZSSLR-C $\cap$ MS-ZSSLR-W} & \multicolumn{3}{c}{MS-ZSSLR-W} & \multicolumn{3}{c}{MS-ZSSLR-W $\cap$ MS-ZSSLR-C}\\
\multicolumn{3}{c}{Clean Train w/ Clean Evaluation} &\multicolumn{3}{c}{Clean Train w/ Wild Evaluation} &\multicolumn{3}{c}{Wild Train w/ Wild Evaluation} & \multicolumn{3}{c}{Wild Train w/ Clean Evaluation}\\ 
\toprule
\hfil top-1 & \hfil top-2 & \hfil top-5 & \hfil top-1 & \hfil  top-2 & \hfil top-5 & \hfil  top-1 & \hfil top-2 & \hfil top-5 & \hfil  top-1 & \hfil  top-2 & \hfil top-5 \\
\cmidrule(lr){1-3}\cmidrule(lr){4-6}\cmidrule(lr){7-9}\cmidrule(lr){10-12}
\hfil \textbf{14.7} & \hfil  \textbf{24.6} & \hfil \textbf{43.4} & \hfil 10.3 & \hfil 17.8 & \hfil 32.6 & \hfil 8.7 & \hfil  15.7 & \hfil  30.3 & \hfil  10.3 & \hfil  18.6 & \hfil 36.0\\
\bottomrule
\end{tabular}%
}
\vspace{-5mm}
\end{center}
\end{table*}

\begin{finalquote}
An important detail in the use combined class embeddings is the trainable linear layer that we use to transform the original $768$-dimensional textual embeddings into smaller $d_t$ dimensional vectors, before concatenating with the attribute embeddings. Figure~\ref{fig:dim-reduction-val-set} shows the effect of $d_t$ choice on the ASL-Text validation scores, based on which we set $d_t=64$. In the figure, the $768$ bin corresponds using the original textual embeddings directly. The results show that the both
LLE with bi-LSTM and LLE with TSM models clearly benefit from the introduction of the transformation layer and both models achieve the highest validation results with $d_t=64$. With this setting, the accuracy scores of the bi-LSTM and TSM based models improve from $26.3$ to $35.5$, and from $20.0$ to $25.3$, respectively.
\end{finalquote}

\mypar{Qualitative results} Figure \ref{fig:prediction} presents several correct and incorrect
zero-shot prediction examples for the ASL-Text and MS-ZSSLR-C datasets. Through the correct prediction
examples, we observe that the model is able to perform well on sign classes with both simple and complex
patterns and descriptions. On the incorrect prediction examples, we observe that the descriptions of
the confused class pairs are similar to each other, highlighting the difficulty of the tackled problem.

\subsubsection{Exploration of MS-ZSSLR-W(ild)}
\label{sub:ms_zsslrwc_exp}

As explained in Section~\ref{sec:dataset}, we obtain MS-ZSSLR-C(lean) by manually filtering samples
from the MS-ZSSLR-W(ild) dataset, according to their adherence to their canonical sign definitions.
While the variations of signs deserve an alternative definitions of the signs for a reliable analysis
of the ZSSLR models, MS-ZSSLR-W still provides an interesting test bed for understanding the effects
of such deviations from the canonical sign definitions. For this reason, we explore MS-ZSSLR-W
through three new experimental protocols where we use MS-ZSSLR-W samples for training only, testing only or for both
training and testing, and compare against the clean-only setting.  For these experiments, we use the configuration that yields the highest results on 
MS-ZSSLR-C, \ie, LLE zero-shot learning formulation using TSM as the visual representation and attributes+text embeddings as the class embeddings.

The corresponding results are presented in Table~\ref{fig:datasetsettings}. We observe that comparing \textit{Clean Train w/ Clean Evaluation} results to \textit{Clean Train w/Wild Evaluation} results, the model trained on clean samples, yields a lower accuracy on wild samples. This is not surprising, since the wild samples includes different dialects that are incompatible with the textual class definitions. Note that the results in Clean vs Wild evaluation setups are not directly comparable, since these tests sets are not identical. When we compare the \textit{Clean Train w/Clean Evaluation} versus \textit{Wild Train w/ Clean Evaluation}, \ie the training set is switched to wild setting over the same test set of clean examples, we observe that incorporating dialects into the training harms the recognition performance and the model trained on MS-ZSSLR-C dataset is more successful. The results over the \textit{Wild Evaluation} follows a similar pattern, and the model trained on the clean samples yields better performance.

These results show the importance of compatibility across visual data and auxiliary
representations for both training and testing purposes. The samples incompatible with the class
definitions not only reduce the recognition success rates when they appear during the test time, but
they also degrade the model quality when used as training samples.  As pointed out by these results, 
modeling sign languages with dialects, especially in zero-shot setting, is an important
open problem. We believe that the MS-ZSSLR-C/W dataset pair provides a valuable benchmark for
future research in this direction.

\begin{table}[t]
\centering
\caption{Generalized zero-shot learning (GZSL) results on ASL-Text.}
\label{tab:gzsl_asl-text}
\small{
\scalebox{0.7}{
\begin{tabular}{llcccc}
\toprule
Method  & Auxiliary &Val (200 Classes) & \multicolumn{3}{c}{Test (250 Classes)} \\
\midrule
& & top-1 & top-1 & top-2 & top-5 \\  \cmidrule(lr){3-3}\cmidrule{4-6}
Random   & - &  0.66  & 0.5 & 1.0 & 2.5  \\ \midrule
\multirow{3}{*}{f-clswgan \cite{xian2018feature}}  & Text & 18.4 &  11.6 & 17.9  & 34.1  \\
 & Attribute & 27.8 &  21.5 & 29.4  & 44.9  \\
 & Attribute + Text & 28.2 & 24.8 & 37.0  & 50.5  \\ \midrule
\multirow{3}{*}{tfvaegan \cite{narayan2020latent}}  & Text & 22.0 &  14.8 & 19.3  & 31.6  \\
 & Attribute & 25.4 &  24.1 & 31.8  & 46.4  \\
 & Attribute + Text & 26.8 & 26.2 & 34.9  & 53.0  \\ \midrule
\multirow{3}{*}{LLE with bi-LSTM}  & Text & 29.7 &  22.5 & 32.5  & 45.6  \\
 & Attribute & 30.5 &  23.5 & 33.8  & 48.4  \\
 & Attribute + Text & \textbf{38.4} & \textbf{26.9} & \textbf{39.5}  & \textbf{55.8}  \\
\bottomrule
\end{tabular}
}}
\vspace{-2mm}
\end{table}

\begin{table}[t]
\centering
\caption{Generalized zero-shot learning (GZSL) results on MS-ZSSLR-C.}
\label{tab:gzsl_ms-zsslr}
\scriptsize{
\scalebox{0.9}{
\begin{tabular}{llcccc}
\toprule
Method  & Auxiliary  & Val (150 Classes) & \multicolumn{3}{c}{Test (200 Classes)} \\
\midrule
& & top-1 & top-1 & top-2 & top-5 \\ \cmidrule(lr){3-3}\cmidrule{4-6}
Random   & - &  0.66  & 0.5 & 1.0 & 2.5  \\ \midrule
 \multirow{3}{*}{f-clswgan \cite{xian2018feature}}     & Text &  28.9  & 21.5 & 30.4 & 42.0 \\
  & Attribute &  32.0  & 29.8 & 38.0 & 49.3 \\
  & Attribute + Text &  33.4  & 31.0 & 39.5 & 51.6 \\ \midrule
    \multirow{3}{*}{tfvaegan \cite{narayan2020latent}}     & Text &  27.0  & 22.7 & 28.3 & 38.6 \\
  & Attribute &  33.5  & 32.6 & 40.4 & 51.7 \\
  & Attribute + Text &  34.5  & 34.2 & 41.9 & 52.2 \\ \midrule
    \multirow{3}{*}{LLE with TSM}     & Text &  45.8  & 33.4 & 40.5 & 48.8 \\
  & Attribute &  45.7  & 33.9 & 42.2 & 52.9 \\
  & Attribute + Text &  \textbf{46.3}  & \textbf{34.7} & \textbf{42.6} & \textbf{53.4} \\
\bottomrule
\end{tabular}
}}
\vspace{-2mm}
\end{table}

\begin{table*}
\centering
\caption{Generalized zero-shot learning accuracy statistics on ASL-Text and MS-ZSSLR-C test sets.}
\label{tab:gzsl_harmonic}
\small{
\scalebox{0.8}{
\begin{tabular}{llccccccccccccc}
\toprule
Method  & Dataset & Overall Acc. & \multicolumn{3}{c}{Seen} & ~ & \multicolumn{3}{c}{Unseen} & ~ &  \multicolumn{3}{c}{Harmonic}\\
\midrule
& & top-1 & top-1 & top-2 & top-5  & & top-1 & top-2 & top-5 &  & top-1 & top-2 & top-5 \\ \cmidrule(lr){3-3} \cmidrule{4-6} \cmidrule{8-10} \cmidrule{12-14}
f-clswgan \cite{xian2018feature}  & \multirow{3}{*}{ASL-Text} & 24.8 &  33.3 & 48.2  & 64.6 &  &  6.7 & 13.2 & 20.7 & & 11.1 & 20.7 & 31.3 \\

tfvaegan \cite{narayan2020latent}  &  & 26.2 &  35.2 & 45.8  & 66.8 &  &  7.1 & 11.5 & 23.5 &  & 11.8 & 18.3 & 34.7 \\

LLE with bi-LSTM  &  & 26.9 &  37.0 & 53.3  & 72.4 &  & 5.5 & 10.4 & 20.3 & &  9.5 & 17.4 & 31.7 \\
\midrule
f-clswgan \cite{xian2018feature}  & \multirow{3}{*}{MS-ZSSLR-C} & 31.0 &  49.2 & 60.9  & 75.1 &  & 3.8 & 7.4 & 16.2 & & 7.0 & 13.1 & 26.6 \\

tfvaegan \cite{narayan2020latent}  &  & 34.2 &  54.2 & 63.8  & 74.4 & & 4.3 & 8.9 & 18.8 & & 7.9 & 15.6 & 30.0 \\

LLE with TSM~~~~~ &  & 34.7 &  54.6 & 64.6  & 76.2 & & 4.8 & 9.7 & 19.1 & & 8.8 & 16.8 & 30.5 \\
\bottomrule
\end{tabular}
}}
\end{table*}

\subsubsection{Generalized-ZSSLR results}
\label{sub:gzssslr_exp}

Finally, we present the results for generalized zero-shot learning on the ASL-Text and MS-ZSSLR-C datasets. On both datasets, we use the best-performing spatiotemporal representation according to the 
ZSL results, \ie, bi-LSTM for ASL-Text and TSM for MS-ZSSLR-C, with LLE formulation. On both datasets, 
we evaluate attribute-only, text-only and attribute-text combination based class embeddings. \begin{finalquote} In addition, we also adapt and evaluate two recent generative GZSL approaches,   f-clswgan~\cite{xian2018feature} and  tfvaegan~\cite{narayan2020latent}, as generative approaches tend to yield more comparable seen and unseen class scores for GZSL. The f-clswgan~\cite{xian2018feature} approach combines a Wasserstein GAN~\cite{wgan} with a classification loss to learn a class embedding conditional feature generating model. The tfvaegan~\cite{narayan2020latent} approach involves learning a variational auto-encoder~\cite{vae1} with adversarial training and a feedback mechanism, and similarly learns a conditional feature generation model. To adapt both approaches, we use our pre-trained, best-performing models to extract spatio-temporal representations, and train the conditional generative models using these features. To obtain the final classifier, we train supervised models over real and generated examples following the implementation details of  \cite{xian2018feature} and \cite{narayan2020latent}.   \end{finalquote} 
 
 \begin{finalquote} 
The GZSL results in terms of top-k accuracy metrics for ASL-Text and MS-ZSSLR-C are presented in Table~\ref{tab:gzsl_asl-text} and
Table~\ref{tab:gzsl_ms-zsslr}, respectively. On ASL-Text dataset, when using only attribute based class embeddings, tfvaegan \cite{narayan2020latent} yields the highest accuracy scores. With text-only, and combined embeddings, the LLE model outperforms the generative approaches. Using the LLE formulation,
we observe that text and attribute-based class representations yield comparable performance on the validation and test sets of both datasets, with relatively better results
for the attribute-based class embeddings. We observe significant performance gains, especially on the ASL-Text dataset, when two embeddings are used in combination in all formulations. \end{finalquote} 

\begin{finalquote} 
In the GZSL setting, the separate accuracy values on the {\em seen} and {\em unseen} class samples are of interest. For this reason, we present the seen and unseen top-k class accuracy scores, and their harmonic means (adapted from \cite{xian2017zero}) on ASL-Text and MS-ZSSLR-C test sets in Table~\ref{tab:gzsl_harmonic}. In these results, we observe that unseen class accuracy values are much lower that the seen ones, similar to the case in zero-shot image classification benchmarks~\cite{xian2017zero}. In contrast to the image classification benchmarks, however, we do not see a clear advantage of the generative approaches in terms of unseen class recognition performances, with mixed results across the datasets. These results suggest that the ZSSLR problem has unique challenges, requiring dedicated efforts for developing representation and recognition models.
\end{finalquote}

\begin{figure*} 
\centering
   \includegraphics[width=0.9\textwidth]{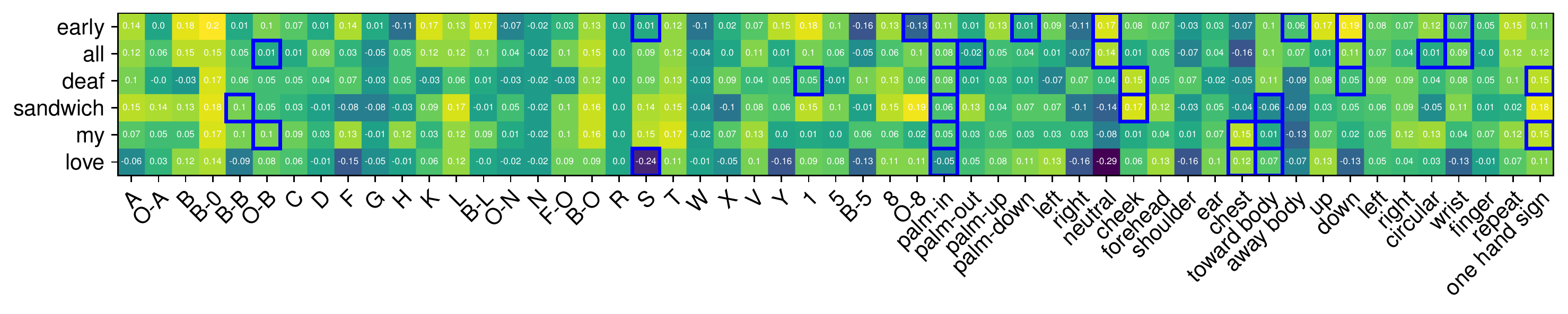}
   \vspace{-5mm}
    \caption{Influence of the attributes on the confidence scores of correct classified samples of five randomly chosen classes. Each row corresponds to a class and each column corresponds to an attribute. Shown numerical values indicate the average influence of an attribute on the corresponding class confidence scores. Thick boxes show the positive attribute relations according to the attribute based class definitions. Lighter colors correspond to higher values. Best viewed in color with zoom.\label{fig:attr-analysis-1}}
    \vspace{-2mm}
\end{figure*}

\begin{figure}
\centering
\includegraphics[width=.95\linewidth]{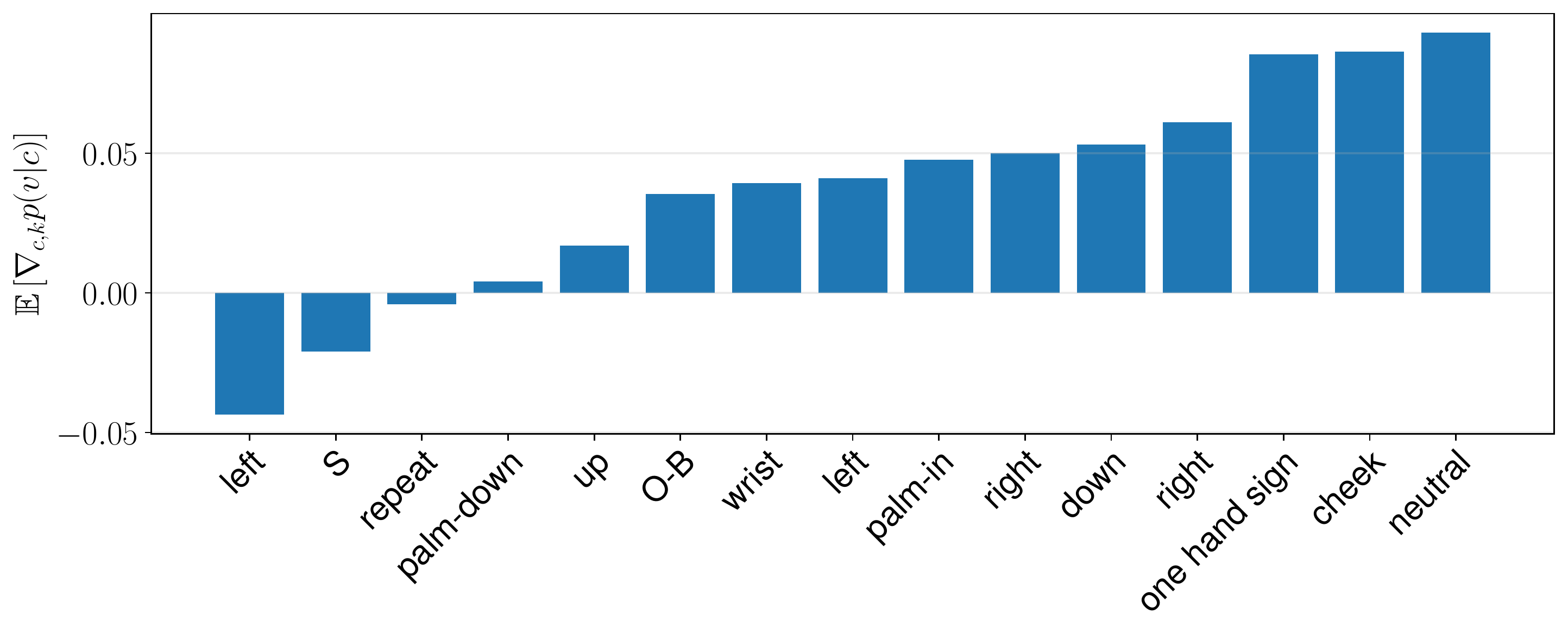}
    \caption{Influence of positive attributes on the confidence scores of correctly classified test samples, averaged over classes. Only the attributes positively associated with at least $10$ unseen classes are shown. \label{fig:att_avg_importance}}
    \vspace{-2mm}
\end{figure}

\begin{figure*}
\centering
   \includegraphics[width=\textwidth,trim=0 0 0 .5cm,clip]{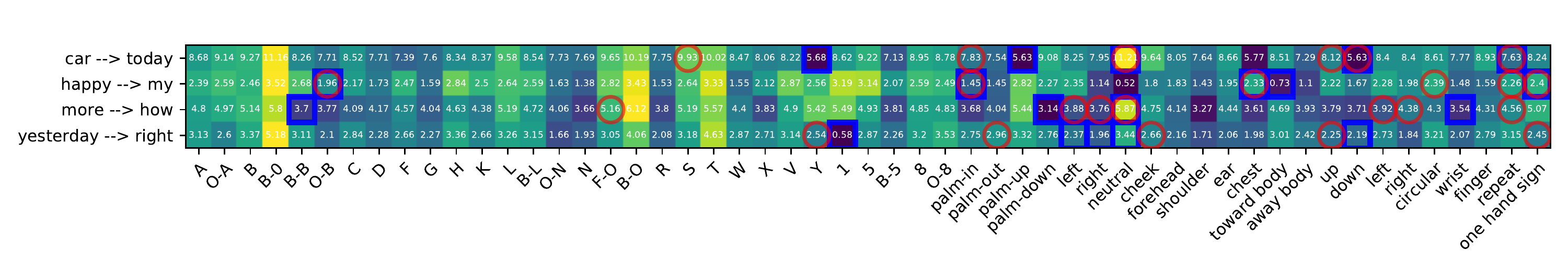}
    \vspace{-5mm}
    \caption{Influence of the attributes on the confidence scores of zero-shot misclassifications. Each row corresponds to a misclassification case, denoted as {\em ground-truth class} $\rightarrow$ {\em predicted class}. Shown numerical values indicate the average influence score according to Eq.~\ref{eq:flip_ratio}. Thick boxes and circles indicate that the corresponding attribute belongs to the corresponding predicted and ground-truth classes, respectively. Lighter colors indicate to higher values, colors in each row are predicted independently for better visualization. Best viewed in color with zoom. \label{fig:attr-analysis-2}}
\end{figure*}

\subsection{Attribute usage analysis} 
\label{sub:attr_analysis}

Our experimental results show that the auxiliary class representation has a major influence on the
success of zero-shot sign language recognition models. In this section, we present the results of
our attribute analysis techniques defined in Section \ref{sec:methodanalysis}.  Throughout this
section, we use the test set predictions on the MS-ZSSLR-C dataset using LLE model based on TSM
features and attribute-only class representations. We present our observations for correct and
incorrect predictions in the following paragraphs.

\mypar{Observations on correct zero-shot predictions} We present binary attribute confidence scores for five randomly chosen unseen classes in Figure~\ref{fig:attr-analysis-1}. Here, we take the correctly classified samples of these classes and compute the average influence of each attribute according to Eq.~\ref{eq:flip_prob}. Each row shows the influence scores of the attributes for one class. By definition, the influence scores can be interpreted as the degree of association between the attribute and class pairs.

From the results, we observe that our model strongly associates certain attributes with the relevant classes. For example, \textit{chest} attribute appears to have strong positive effects on the \textit{my} and \textit{love} class predictions, and, negative effects on the other classes, all of which are consistent with the auxiliary knowledge on class-attribute relations. Similarly, \textit{cheek} and \textit{neutral} attributes influence scores show consistent patterns with the class-attribute relation definitions. 
The influence scores also point to certain cases where the model seems to make incorrect, or unexpected, associations between the visual patterns and class attributes. For example, 
\textit{O-8} (stands for \textit{Open-8}) and \textit{8} appears to be negatively correlated with the attribute definitions of the \textit{early} class. We believe that this is due to
subtle differences, where the model is prone to making incorrect generalizations.

To further understand the overall impact of the attributes, for each attribute, we choose the unseen classes that are positively affiliated with that attribute and average the attribute influence values over these classes. In Figure~\ref{fig:att_avg_importance}, we show the results for the subset of attributes where there exists at least 10 such positively affiliated class according to the class attribute definitions. Consistent with our aforementioned observations, we observe that \textit{neutral} appears to be a strong positive contributor to the related correct class confidence scores, similar to many other attributes such as \textit{cheek} (stands for \textit{cheek/chin/mouth/nose}) and \textit{one hand sign}. While we would ideally expect to see only such positive values for all attributes, few attributes, namely \textit{left} and \textit{S} (and marginally \textit{repeat}), negatively affect the predictions on average. One of the factors in poor modeling of these attributes is possibly their rarity among the training examples, as \textit{left} and \textit{S} are affiliated respectively with only 8 and 10 classes, in contrast to 66 affiliations of \textit{neutral}. Overall, these analysis results show that the ZSSLR model tends to learn meaningful relations, but there is room for improvement in building more explainable and robust models.

\mypar{Observations on zero-shot misclassifications}
Finally, we look into the influence of class-attribute definitions on zero-shot misclassifications.
For this analysis, we find four most commonly confused ground-truth - predicted class tuples, and
compute the attribute influence scores based on log-ratio of class posteriors, according to
Eq.~\ref{eq:flip_ratio}. We present the results in Figure~\ref{fig:attr-analysis-2}, where each row
corresponds to a particular class confusion case. Thick boxes and circles indicate that the
corresponding attribute belongs to the corresponding predicted and ground-truth classes,
respectively. We note that the influence scores here are based on the rate of probability transition
from incorrectly predicted classes to the correct ones. Therefore, a larger value
indicates a relatively bigger role of an attribute on the resulting misclassification.

In Figure~\ref{fig:attr-analysis-2}, we observe large influence scores in some attributes that are positively defined for both the predicted
and the ground-truth class, as expected. This is particularly the case 
with the \textit{neutral} attribute, which seems to be modelled accurately according to our preceding observations, as well.
However, this is not always the case, which suggests that the model suffers from poor (and unclear) attribute
recognitions in these problematic misclassification cases. It is also noticeable that \textit{B-0}, which stands for \textit{Baby 0}, is inactive for all classes, 
yet activating this attribute in the definition of the assigned classes would yield great drops. It turns out that this attribute 
is positively defined for only one training class, explaining why the model might be behaving instable
as a function of this attribute. Overall, these results suggest that building ZSL models with more robust (implicit) attribute predictors can potentially be a fruitful future research direction.

\vspace{-3mm}
\section{Conclusion}
\label{sec:conclusions}

This paper explores the problem of ZSSLR and GZSSLR. We present three benchmark datasets for this novel problem by augmenting two large ASL dataset with sign language dictionary descriptions and attributes. Our proposed framework builds upon the idea of using these auxiliary texts and attributes as additional sources of information to recognize unseen signs. We propose an end-to-end trainable ZSSLR method that focuses hand and full body regions via several temporal modeling approaches and learns a compatibility function via label embedding. Overall, the experiments yield promising results on zero-shot recognition of signs. Nevertheless, the acquired accuracy levels are relatively low compared to other ZSL domains, pinpointing a substantial need for further exploration of this direction. The observations based on our binary attribute analysis methodologies suggest that one of the fundamental challenges in ZSSLR is to build models that accurately relate visual patterns to the elements of class definitions, instead of relying on spurious correlations, in making zero-shot predictions.

In our work, we observe two fundamental challenges inherent to the (G)ZSSLR task. First, some of the differences across sign descriptions are subtle, both visually and textually. Second, many dialects exists; even the same sign can be expressed in many different forms. Future work should try to address these issues, possibly by developing visual representations and recognition models that can better capture fine details. Incorporation of additional visual cues, \eg facial expressions and pose information, are also likely to be beneficial. We believe that (G)ZSSLR is an important research direction towards building large-vocabulary sign language recognition systems.

\ifCLASSOPTIONcaptionsoff
  \newpage
\fi

\bibliographystyle{IEEEtran}
\bibliography{IEEEabrv,egbib}

\vspace{-5mm}
\vskip -1.5\baselineskip plus -1fil
\begin{IEEEbiography}[{\includegraphics[width=1in,height=1.25in,clip,keepaspectratio]{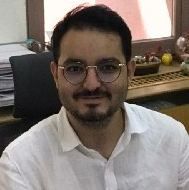}}]{Yunus Can Bilge}
received his BSc degrees from Izmir University of Economics, Turkey and received his MEngSc degree from University of New South Wales, Australia. He is a PhD candidate at Hacettepe University, Turkey. His research interests include computer vision and machine learning with minimal supervision.
\end{IEEEbiography}
\vskip -1\baselineskip plus -1fil
\begin{IEEEbiography}[{\includegraphics[width=1in,height=1.25in,clip,keepaspectratio]{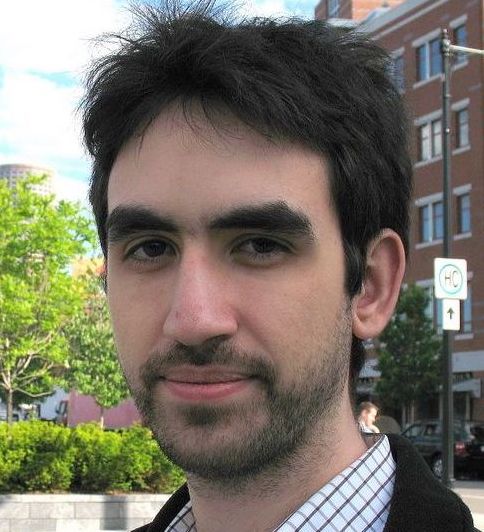}}]{Ramazan Gokberk Cinbis}
    received his Bsc degree from Bilkent University, Turkey, and received his M.A. degree from Boston University, USA.
    He was a doctoral student at INRIA Grenoble between 2010-2014, and received a PhD degree
    from Universit\'{e} de Grenoble, France,
    in 2014. He is currently a faculty member at METU, Ankara, Turkey. His research interests include machine learning with limited and incomplete supervision.
\end{IEEEbiography}
\vspace{-8mm}
\begin{IEEEbiography}[{\includegraphics[width=1in,height=1.25in,clip,keepaspectratio]{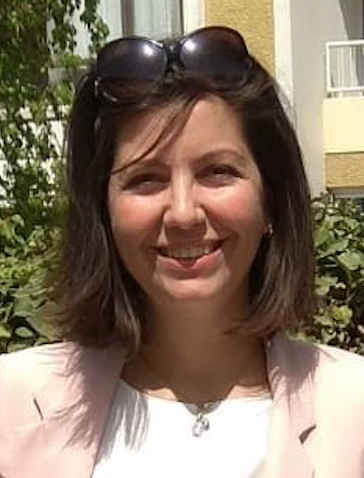}}]{Nazli Ikizler-Cinbis}
received her BSc and MSc degrees from  Department of Computer Engineering at Bilkent University. During  2005-2006, she was a visiting scholar at University of Illinois at  Urbana-Champaign (UIUC). After receiving her PhD degree from Bilkent  University, she worked as a post-doctoral research associate  at Boston University (USA). Since 2011, she is a faculty member at Hacettepe University. Her research areas are computer vision and machine  learning, specifically focusing on video processing, human action and interaction recognition, and zero-shot learning. 
\end{IEEEbiography}

\vspace{1cm}

\appendix
\vspace{.5cm}

\section*{MS-ZSSLR-(W)ild Dataset}
MS-ASL~\cite{vaezijoze2019ms-asl} dataset contains significant variations due to various dialects in sign languages, and the sign language dictionaries do not comprehensively provide the dialect descriptions, to the best of our knowledge. Based on the MS-ASL~\cite{vaezijoze2019ms-asl} dataset, we have therefore created two separate benchmarks: MS-ZSSLR-W(ild) and MS-ZSSLR-C(lean). MS-ZSSLR-W contains the dialects, and MS-ZSSLR-C contains only the samples where the textual descriptions are inline with the visual descriptions of signs, as also explained in the main paper.

\begin{figure*}
\centering
\begin{tabular}{>{\raggedright\arraybackslash}p{0.235\linewidth}@{$\;\;$}>{\raggedright\arraybackslash}p{0.235\linewidth}@{$\;\;$}>{\raggedright\arraybackslash}p{0.235\linewidth}@{$\;\;$}>{\raggedright\arraybackslash}p{0.235\linewidth}}
\toprule
{\textit{MS-ZSSLR-W(ild)}} \\
\scriptsize{\textbf{COMPUTER}} & 
\vspace{-4mm}
\scriptsize{\textbf{Variation 1}} & 
\vspace{-4mm}
\scriptsize{\textbf{Variation 2}} & 
\vspace{-4mm}
\scriptsize{\textbf{Variation 3}} 
 \\
\includegraphics[width=0.25\linewidth, height=1.5cm]{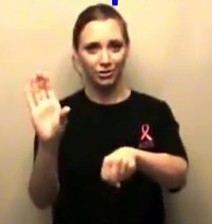}
\includegraphics[width=0.25\linewidth, height=1.5cm]{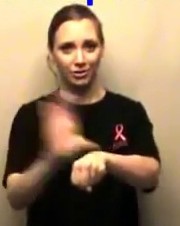}
\includegraphics[width=0.25\linewidth, height=1.5cm]{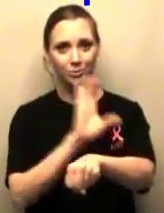} 
\begin{minipage}{0.23\textwidth}
\vspace{1mm}
\tiny{Move the thumb side of the right \textcolor{red}{C} hand, palm facing left, from touching the lower part of extended left arm upward to touch the upper arm with a repeated movement.\\ }
\end{minipage} 
 & 
\includegraphics[width=0.25\linewidth, height=1.5cm]{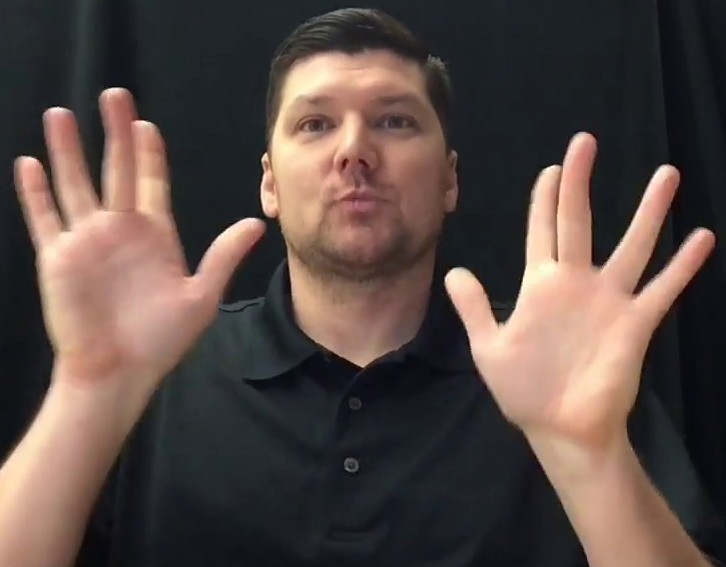}
\includegraphics[width=0.25\linewidth, height=1.5cm]{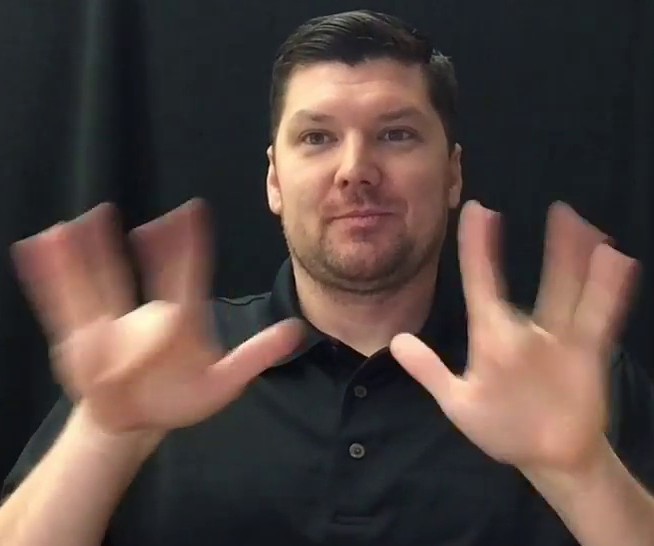}
\includegraphics[width=0.25\linewidth, height=1.5cm]{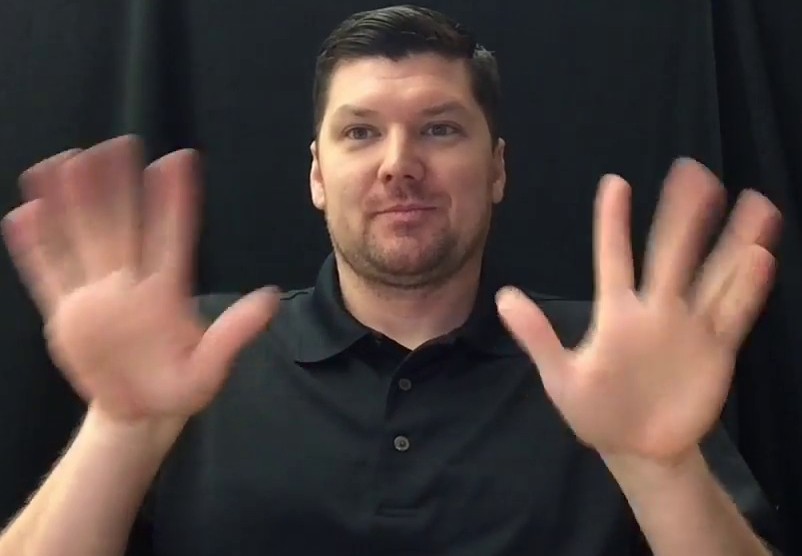}
\begin{minipage}{0.23\textwidth}
\tiny{Different handshape and movement.\\ }
\vspace{1mm}
\end{minipage} 
& 
\includegraphics[width=0.25\linewidth, height=1.5cm]{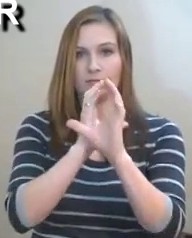}
\includegraphics[width=0.25\linewidth, height=1.5cm]{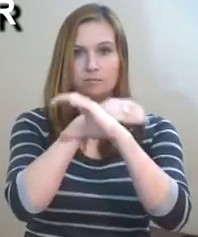}
\includegraphics[width=0.25\linewidth, height=1.5cm]{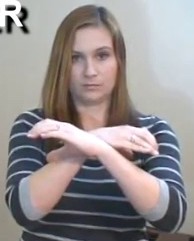}
\begin{minipage}{0.23\textwidth}
\tiny{Mainly wrist movement is executed.}
\vspace{1mm}

\end{minipage} 
& 
\includegraphics[width=0.25\linewidth, height=1.5cm]{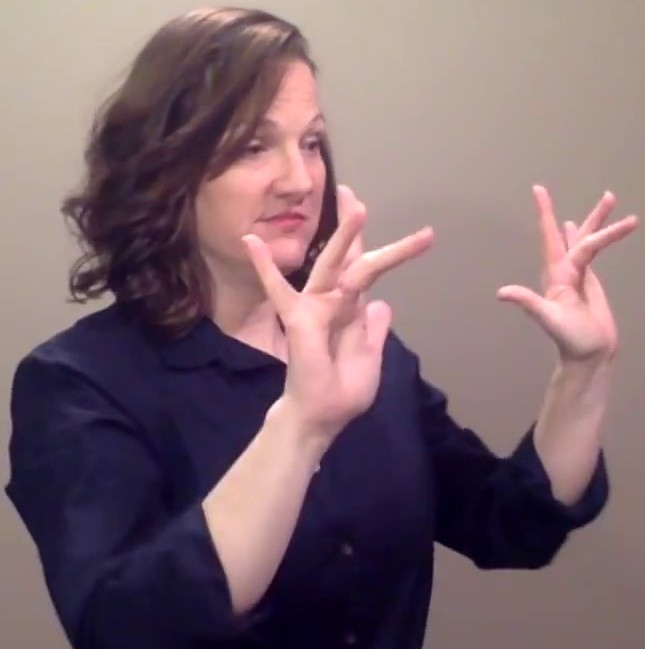}
\includegraphics[width=0.25\linewidth, height=1.5cm]{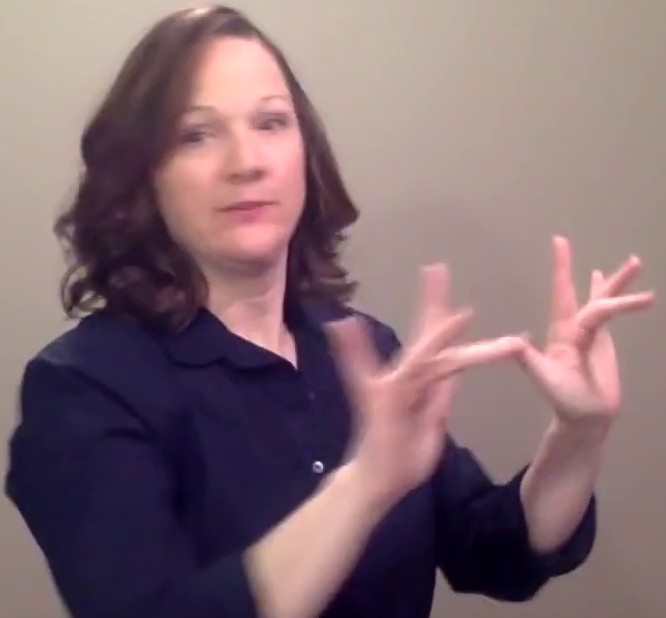}
\includegraphics[width=0.25\linewidth, height=1.5cm]{images/appendix_images/computer_v3/03767.jpg}
\begin{minipage}{0.23\textwidth}
\tiny{Different handshape and movement.\\ }
\vspace{1mm}
\vspace{1mm}

\end{minipage}
\\
\vspace{-4mm}
\scriptsize{\textbf{COUSIN}} & 
\vspace{-4mm}
\scriptsize{\textbf{Variation 1}} & 
\vspace{-4mm}
\scriptsize{\textbf{Variation 2}} & 
\vspace{-4mm}
\scriptsize{\textbf{Variation 3}} \\
\includegraphics[width=0.25\linewidth, height=1.5cm]{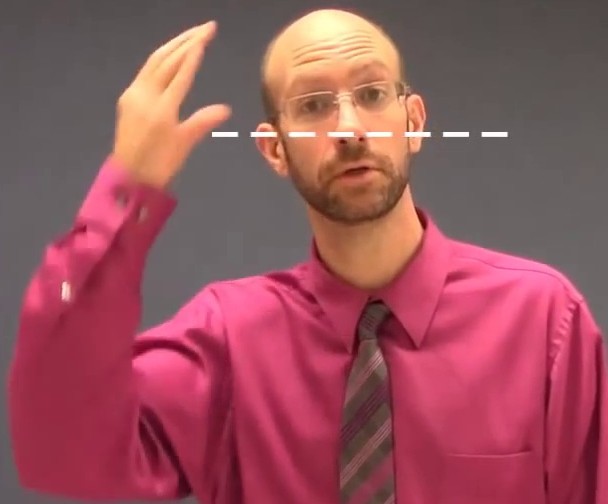}
\includegraphics[width=0.25\linewidth, height=1.5cm]{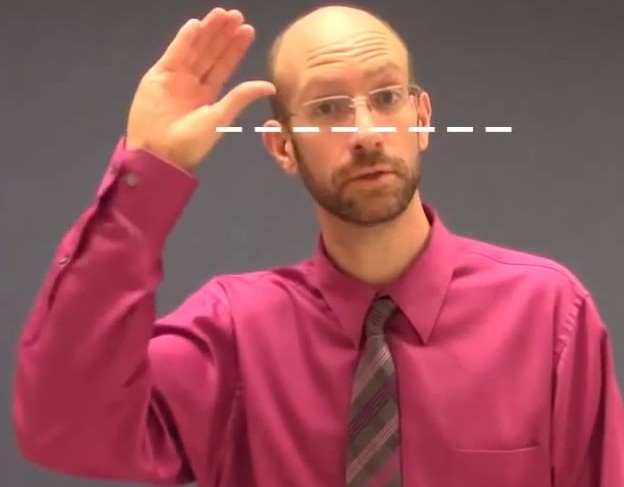}
\includegraphics[width=0.25\linewidth, height=1.5cm]{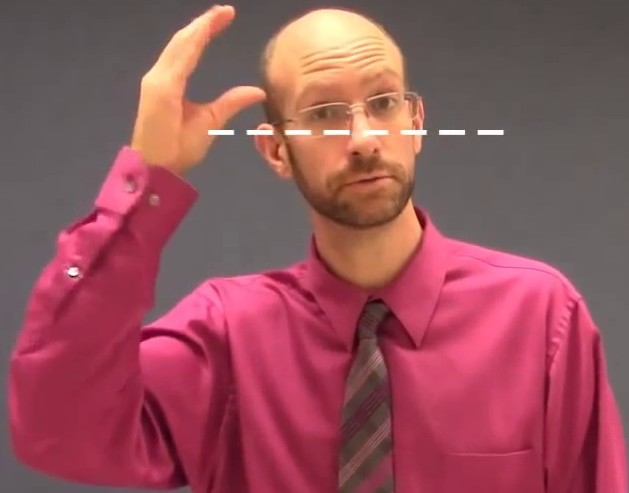}
\begin{minipage}{0.23\textwidth}
\vspace{1mm}
\tiny{Move the right C hand, palm facing left, with a shaking movement near the right side of the forehead. \\}
\end{minipage}
 & 
\includegraphics[width=0.25\linewidth, height=1.5cm]{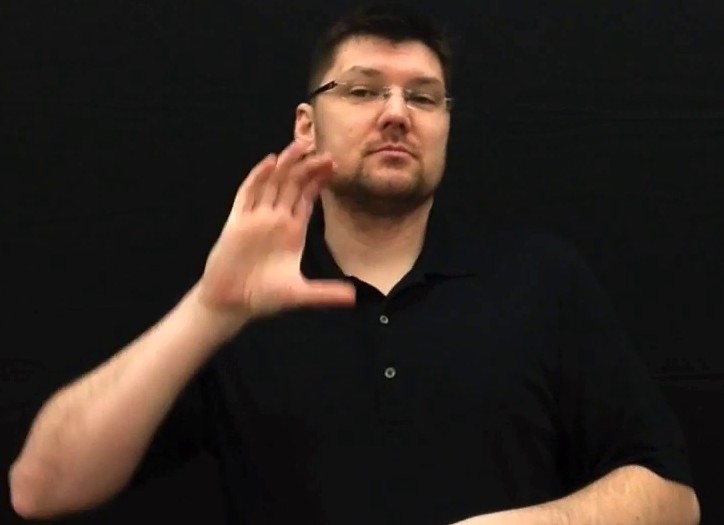}
\includegraphics[width=0.25\linewidth, height=1.5cm]{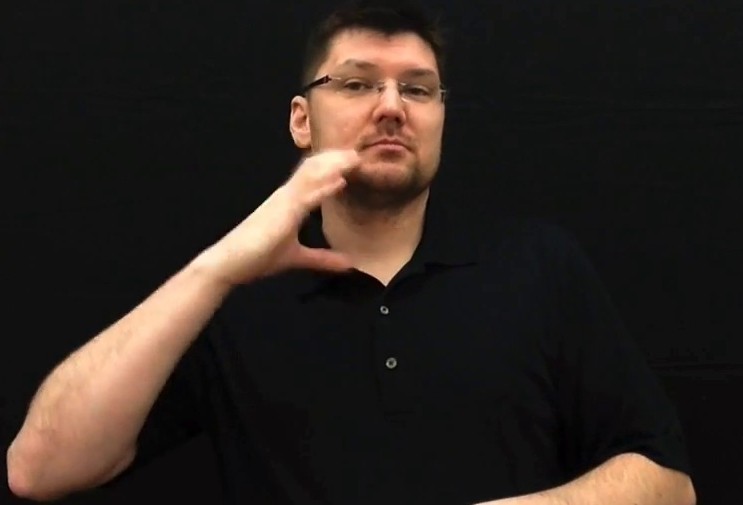}
\includegraphics[width=0.25\linewidth, height=1.5cm]{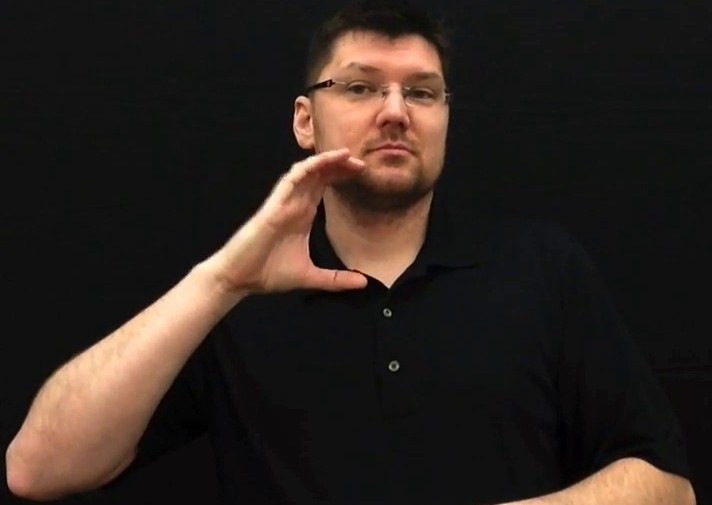}
\begin{minipage}{0.23\textwidth}
\tiny{The sign is executed on the chin.}
\vspace{1mm}

\end{minipage} 
 & 
 \includegraphics[width=0.25\linewidth, height=1.5cm]{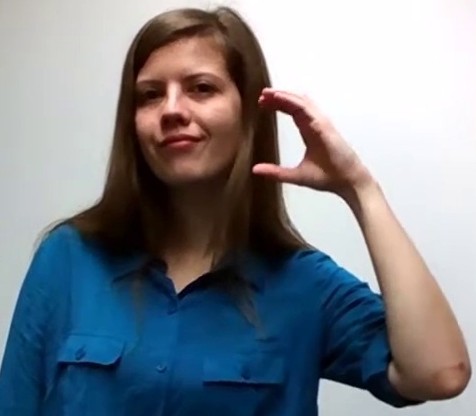}
\includegraphics[width=0.25\linewidth, height=1.5cm]{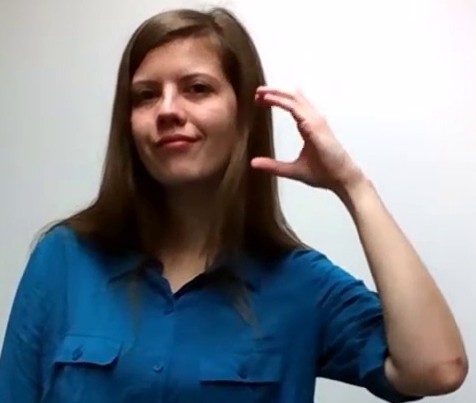}
\includegraphics[width=0.25\linewidth, height=1.5cm]{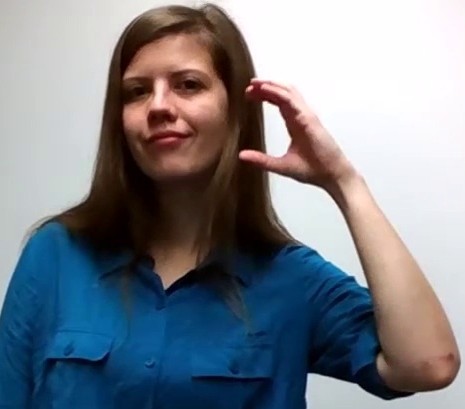}
\begin{minipage}{0.23\textwidth}
\tiny{Left hand is used.}
\vspace{1mm}
\end{minipage} 
& 
\includegraphics[width=0.25\linewidth, height=1.5cm]{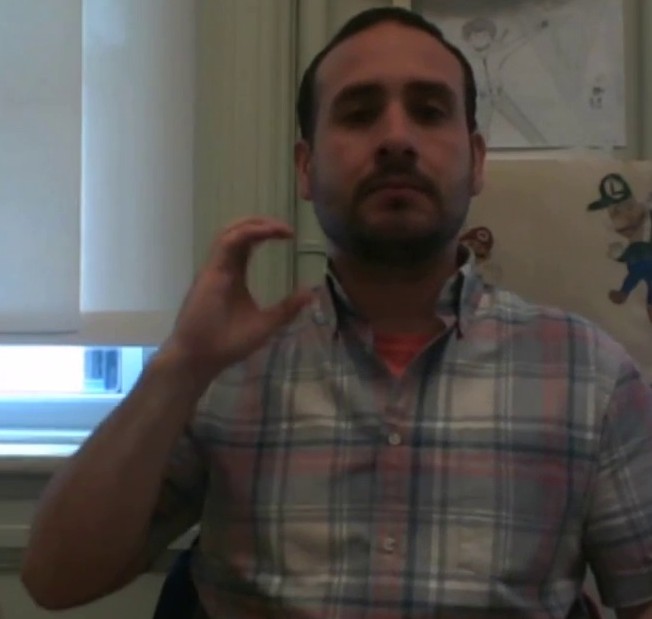}
\includegraphics[width=0.25\linewidth, height=1.5cm]{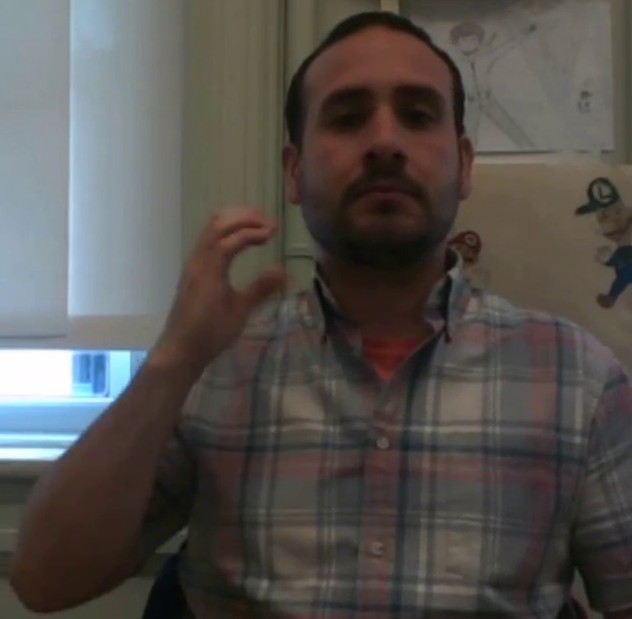}
\includegraphics[width=0.25\linewidth, height=1.5cm]{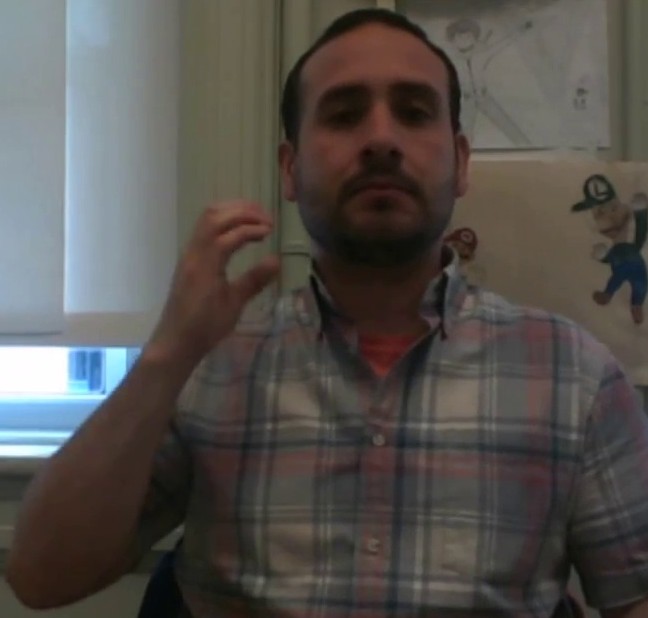}
\begin{minipage}{0.23\textwidth}
\tiny{The sign is again executed on the chin.}
\vspace{1mm}
\end{minipage}\\

\\
\vspace{-4mm}
\scriptsize{\textbf{HORSE}} & 
\vspace{-4mm}
\scriptsize{\textbf{Variation 1}} & 
\vspace{-4mm}
\scriptsize{\textbf{Variation 2}} & 
\vspace{-4mm}
\scriptsize{\textbf{Variation 3}} \\
\includegraphics[width=0.25\linewidth, height=1.5cm]{images/horse/horse_0046.jpg}
\includegraphics[width=0.25\linewidth, height=1.5cm]{images/horse/horse_0068.jpg}
\includegraphics[width=0.25\linewidth, height=1.5cm]{images/horse/horse_0090.jpg}
\begin{minipage}{0.23\textwidth}
\vspace{1mm}
\tiny{With the extended thumb of the right U hand against the right side of the forehead, palm facing forward, bend the fingers of the U hand up and down with a double movement. \\}
\end{minipage}
 & 
\includegraphics[width=0.25\linewidth, height=1.5cm]{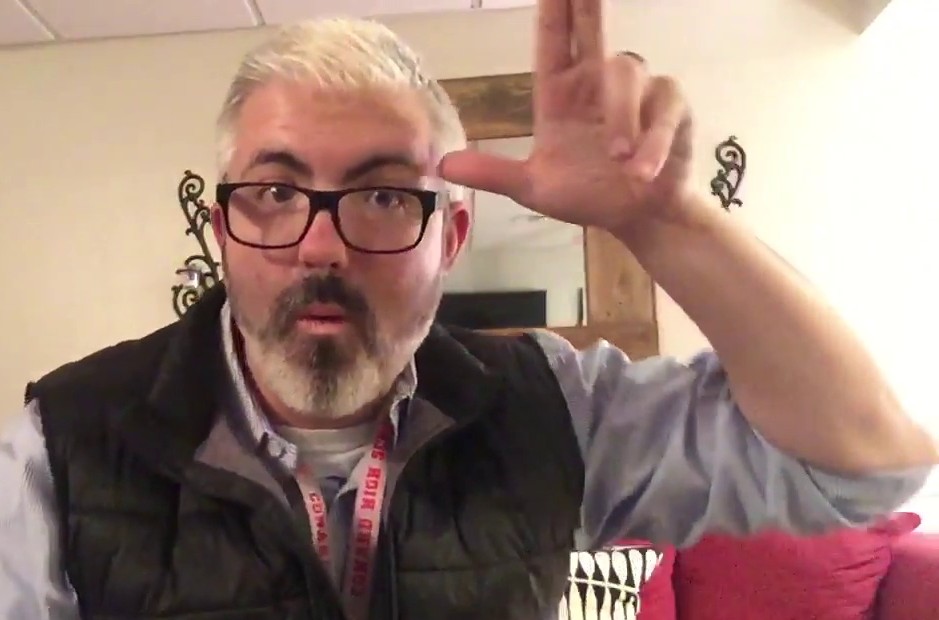}
\includegraphics[width=0.25\linewidth, height=1.5cm]{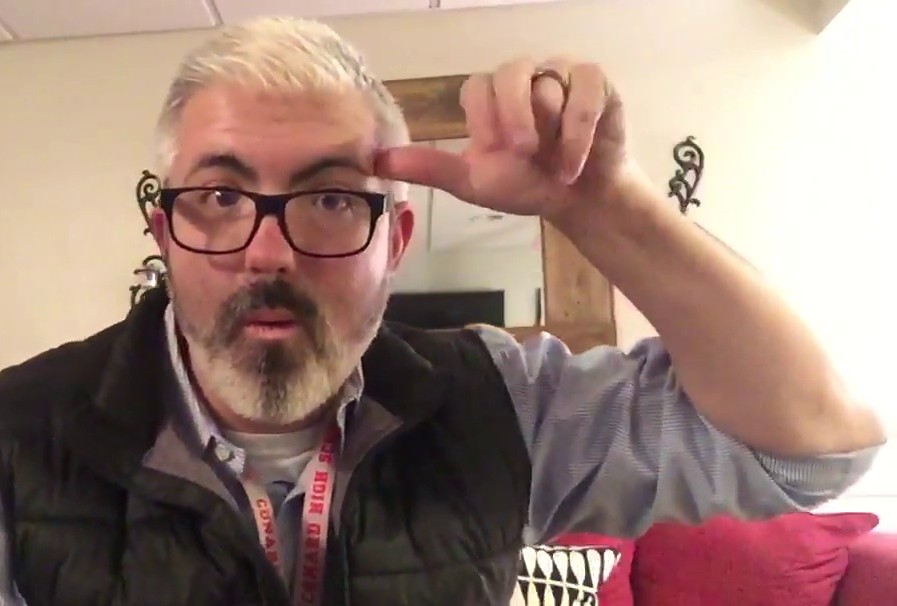}
\includegraphics[width=0.25\linewidth, height=1.5cm]{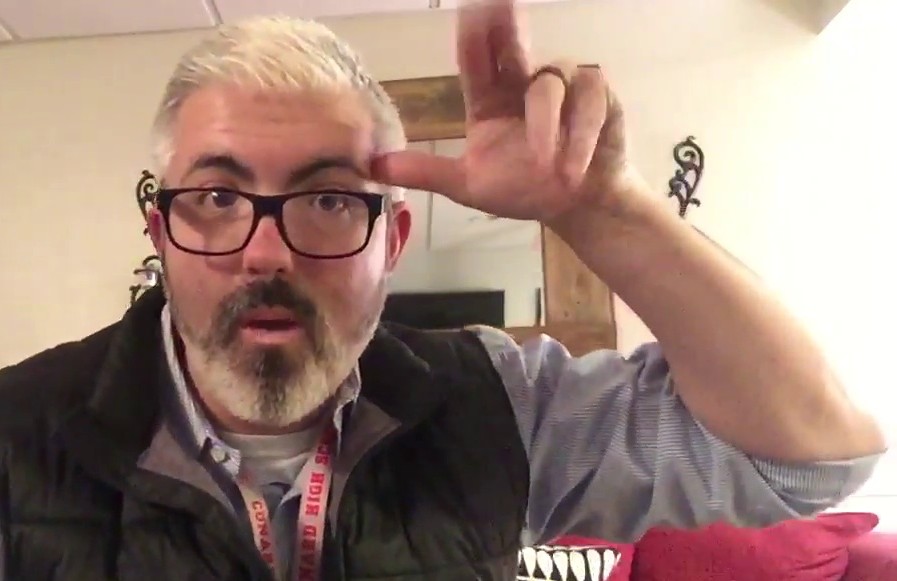}
\begin{minipage}{0.23\textwidth}
\tiny{Left hand is used.\\}
\vspace{1mm}

\end{minipage} 
 & 
 \includegraphics[width=0.25\linewidth, height=1.5cm]{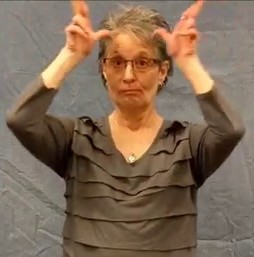}
\includegraphics[width=0.25\linewidth, height=1.5cm]{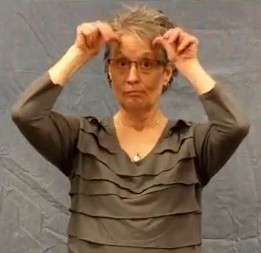}
\includegraphics[width=0.25\linewidth, height=1.5cm]{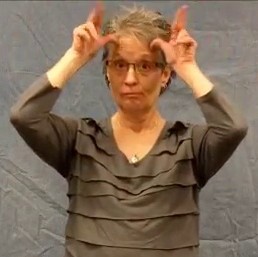}
\begin{minipage}{0.23\textwidth}
\vspace{1mm}
\tiny{Both hands are used.\\}
\end{minipage} 
& 
\includegraphics[width=0.25\linewidth, height=1.5cm]{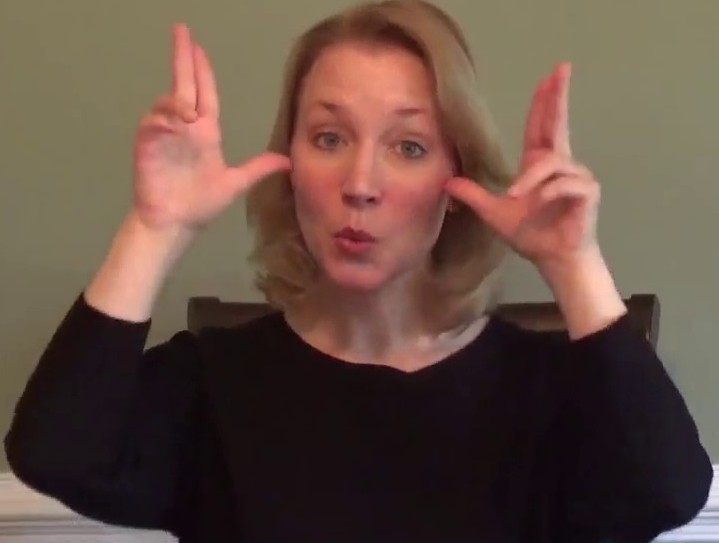}
\includegraphics[width=0.25\linewidth, height=1.5cm]{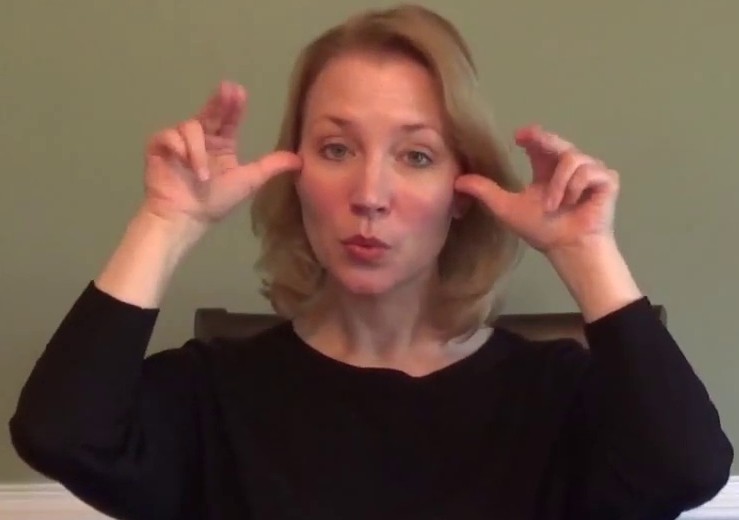}
\includegraphics[width=0.25\linewidth, height=1.5cm]{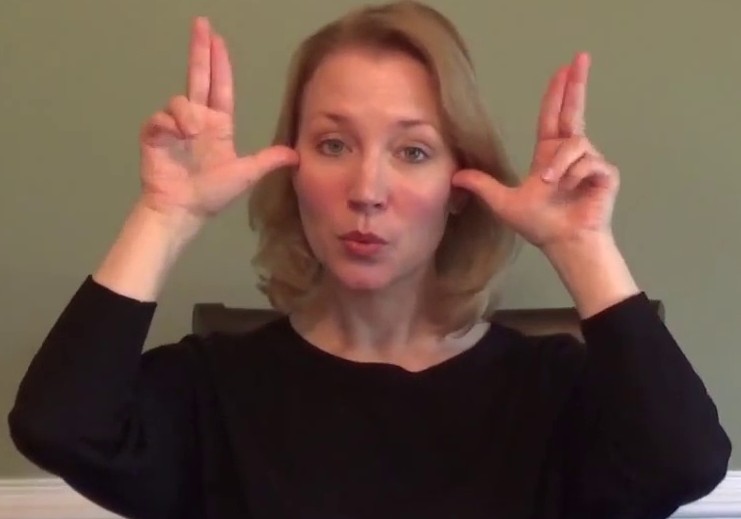}
\begin{minipage}{0.23\textwidth}
\tiny{The sign is executed on the cheek.\\}
\vspace{1mm}
\end{minipage}\\

\\
\vspace{-4mm}
\scriptsize{\textbf{LOST}} & 
\vspace{-4mm}
\scriptsize{\textbf{Variation 1}} & 
\vspace{-4mm}
\scriptsize{\textbf{Variation 2}} & 
\vspace{-4mm}
\scriptsize{\textbf{Variation 3}} \\
\includegraphics[width=0.25\linewidth, height=1.5cm]{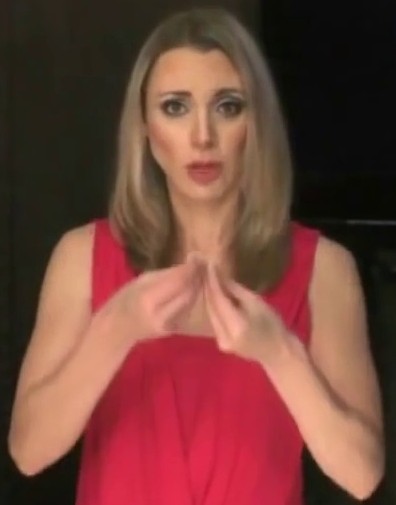}
\includegraphics[width=0.25\linewidth, height=1.5cm]{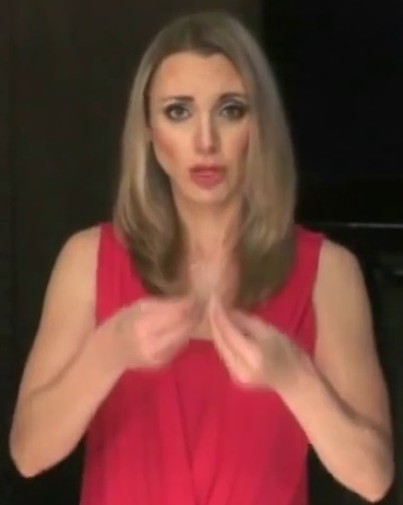}
\includegraphics[width=0.25\linewidth, height=1.5cm]{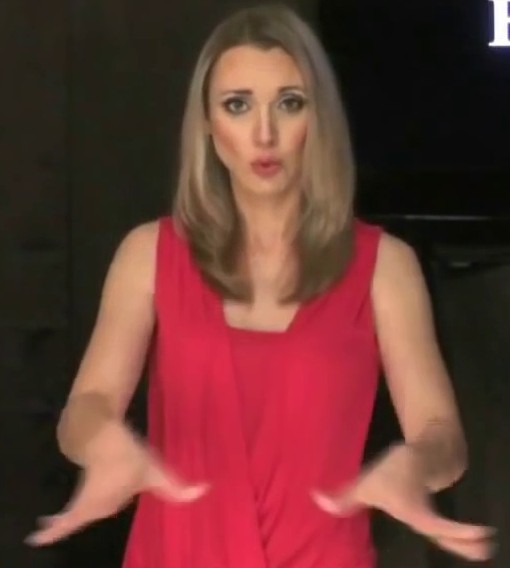}
\begin{minipage}{0.23\textwidth}
\vspace{1mm}
\tiny{Beginning with the fingertips of both flattened O hands touching in front of the body, palms facing up, drop the fingers quickly downward and away from each other while opening into 5 hands, ending with both palms and fingers angled downward. \\}
\end{minipage}
 & 
\includegraphics[width=0.25\linewidth, height=1.5cm]{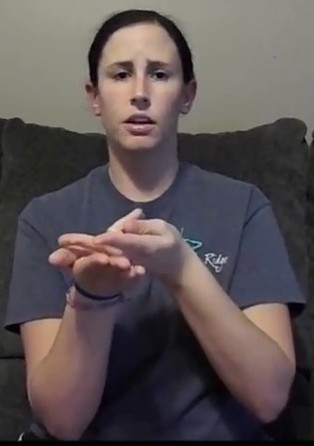}
\includegraphics[width=0.25\linewidth, height=1.5cm]{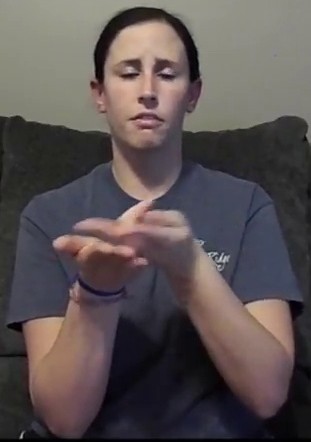}
\includegraphics[width=0.25\linewidth, height=1.5cm]{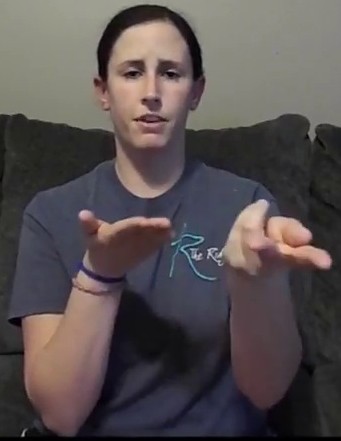}
\begin{minipage}{0.23\textwidth}
\tiny{Different handshape and movement. \\}
\vspace{1mm}

\end{minipage} 
 & 
 \includegraphics[width=0.25\linewidth, height=1.5cm]{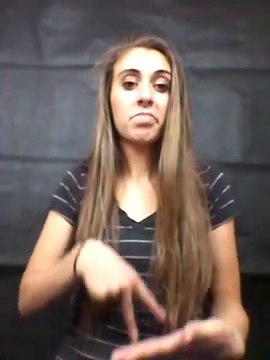}
\includegraphics[width=0.25\linewidth, height=1.5cm]{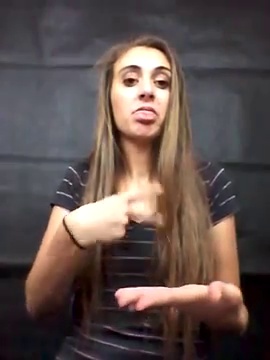}
\includegraphics[width=0.25\linewidth, height=1.5cm]{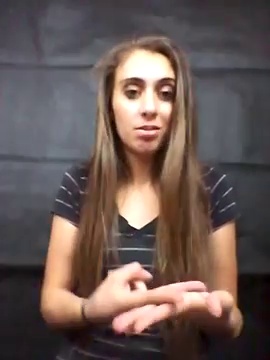}
\begin{minipage}{0.23\textwidth}
\tiny{Different handshape and movement. \\}
\vspace{1mm}
\end{minipage} 
& 
\includegraphics[width=0.25\linewidth, height=1.5cm]{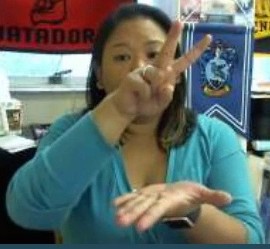}
\includegraphics[width=0.25\linewidth, height=1.5cm]{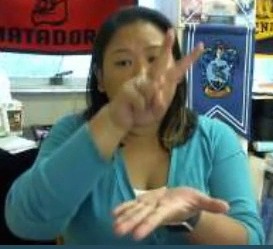}
\includegraphics[width=0.25\linewidth, height=1.5cm]{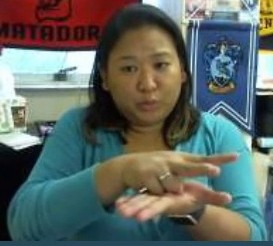}
\begin{minipage}{0.23\textwidth}
\tiny{Different handshape and movement. \\}
\vspace{1mm}
\end{minipage}\\

\\
\vspace{-4mm}
\scriptsize{\textbf{NOTHING}} & 
\vspace{-4mm}
\scriptsize{\textbf{Variation 1}} & 
\vspace{-4mm}
\scriptsize{\textbf{Variation 2}} & 
\vspace{-4mm}
\scriptsize{\textbf{Variation 3}} \\
\includegraphics[width=0.25\linewidth, height=1.5cm]{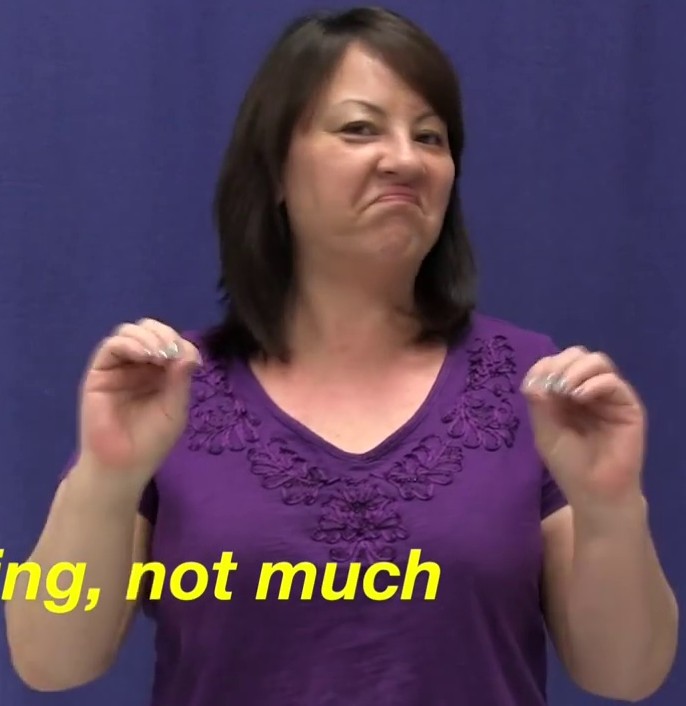}
\includegraphics[width=0.25\linewidth, height=1.5cm]{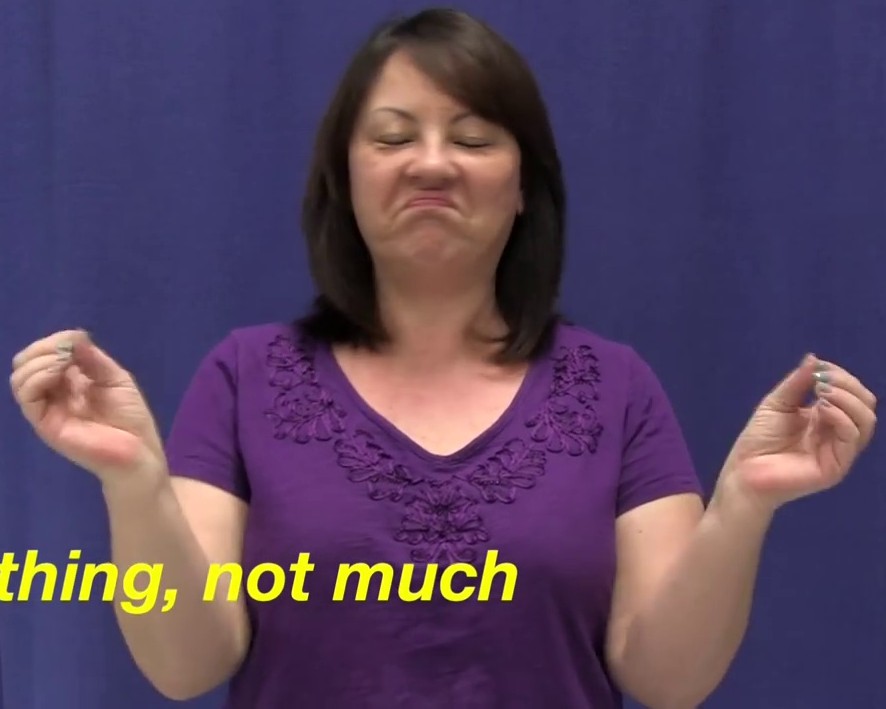}
\includegraphics[width=0.25\linewidth, height=1.5cm]{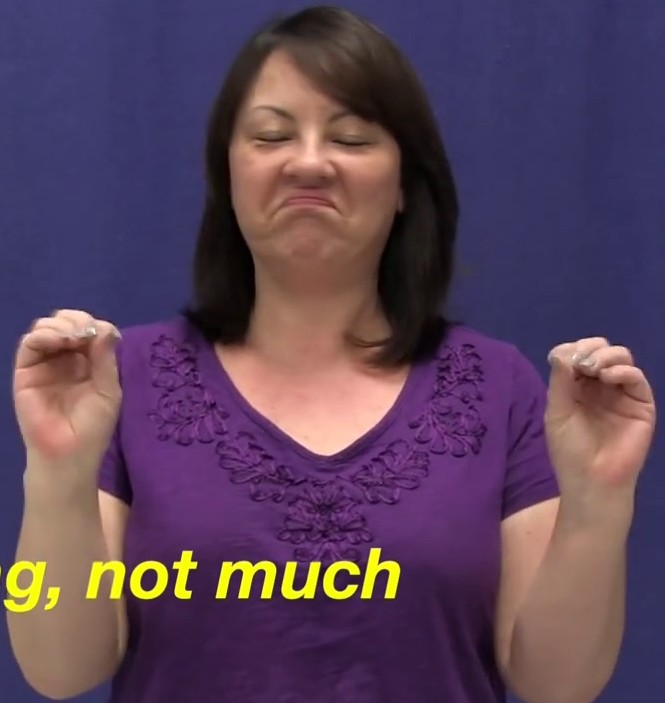}
\begin{minipage}{0.23\textwidth}
\vspace{1mm}
\tiny{Move both flattened O hands, palms facing forward, from side to side with repeated movement in front of each side of the chest. \\}
\end{minipage}
 & 
\includegraphics[width=0.25\linewidth, height=1.5cm]{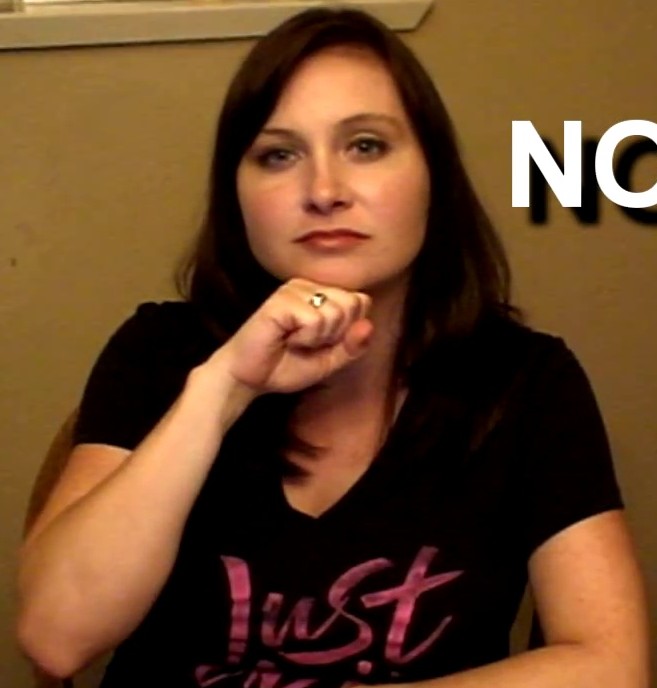}
\includegraphics[width=0.25\linewidth, height=1.5cm]{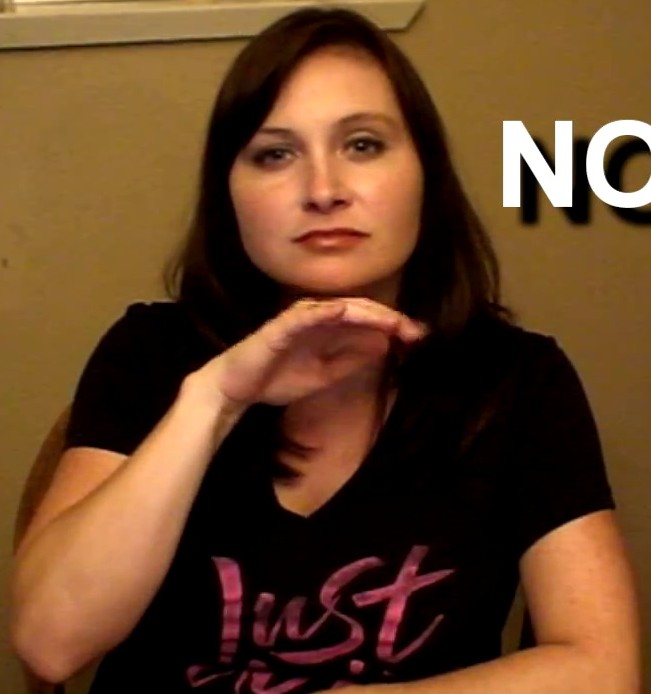}
\includegraphics[width=0.25\linewidth, height=1.5cm]{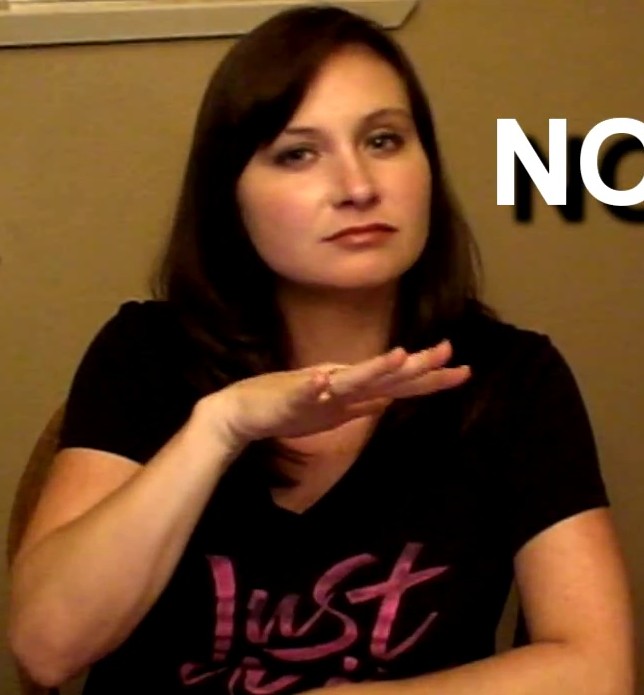}
\begin{minipage}{0.23\textwidth}
\tiny{Sing hand sign, different handshape and movement.\\}
\vspace{1mm}

\end{minipage} 
 & 
 \includegraphics[width=0.25\linewidth, height=1.5cm]{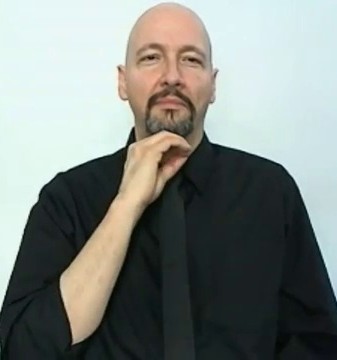}
\includegraphics[width=0.25\linewidth, height=1.5cm]{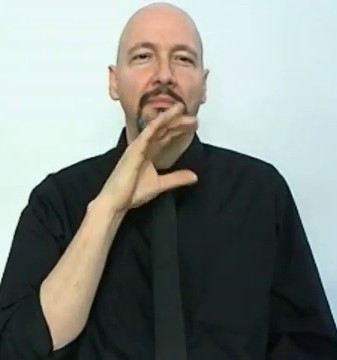}
\includegraphics[width=0.25\linewidth, height=1.5cm]{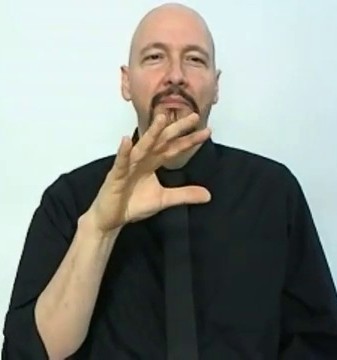}
\begin{minipage}{0.23\textwidth}
\tiny{Sing hand sign, different handshape and movement. \\}
\vspace{1mm}
\end{minipage} 
& 
\includegraphics[width=0.25\linewidth, height=1.5cm]{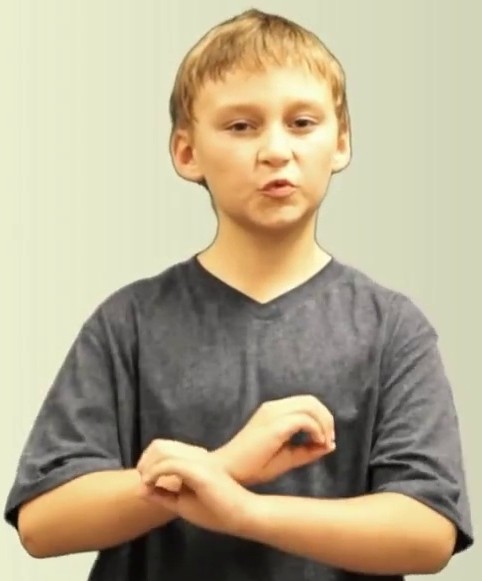}
\includegraphics[width=0.25\linewidth, height=1.5cm]{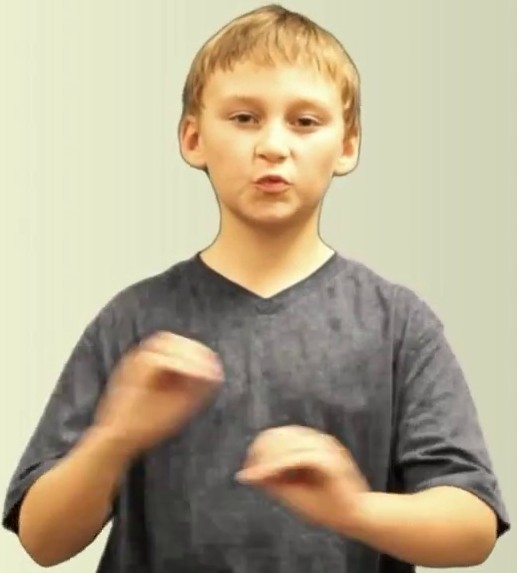}
\includegraphics[width=0.25\linewidth, height=1.5cm]{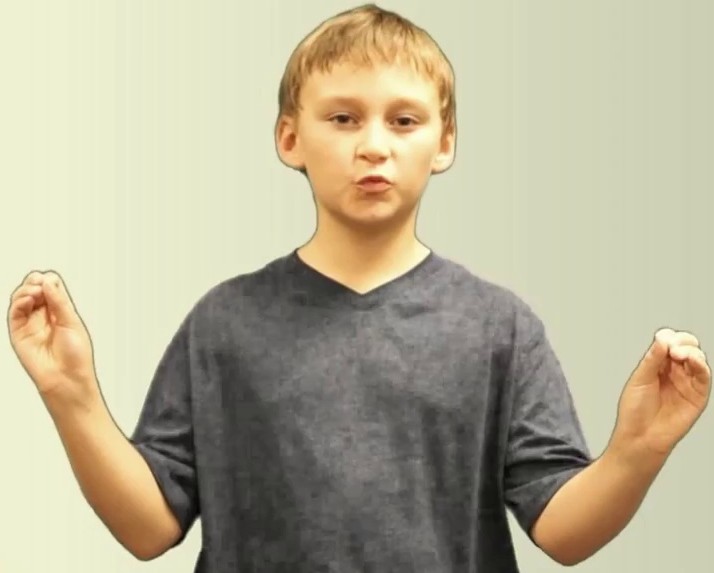}
\begin{minipage}{0.23\textwidth}
\tiny{Different handshape and movement.}\\
\vspace{1mm}
\end{minipage}\\
\bottomrule
\end{tabular}
\caption{Example in-class variations and corresponding textual descriptions from MS-ZSSLR-W(ild) dataset. The first column shows an example video that is inline with the dictionary description. The remaining columns show different visual variations. }
\label{fig:appendix_dataset}

\end{figure*}

For our MS-ZSSLR-C/W benchmarks, we collect text and attribute based class definitions from Webster American Sign Language Dictionary~\cite{costello1999random} and American Sign Language Handshape Dictionary~\cite{tennant1998american}. Figure~\ref{fig:appendix_dataset} shows examples for demonstrating the aforementioned variances of dialects, and the problem of visual sign and definition mismatches.
The first column shows an example video that is inline with the dictionary description. The remaining columns show different visual variations. We observe that the differences are mostly due to the variations in handshape, hand location with respect to the body and hand movements in sign making.  Some signers execute the sign with two hands while the sign is supposed to be executed with one hand according to the textual definition or vice-versa. We also observe variations due to left-handed signers in the dataset. 

\section*{Attributes}
Attribute-based representations are defined with respect to four main parameters: \textit{handshape}, \textit{palm orientation}, \textit{hand location}, \textit{hand movement}. There are two other attributes describing whether the sign includes repeated movements, and whether the sign is executed with a single hand. Below is the list of attributes defined for ASL-Text dataset:
\begin{itemize}
    \item \textbf{Handshape:} {\tt A}, {\tt Open A}, {\tt B}, {\tt Baby 0}, {\tt Bent B}, {\tt V}, {\tt Bent V}, {\tt Open B}, {\tt C}, {\tt D}, {\tt E}, {\tt F}, {\tt G}, {\tt H}, {\tt I}, {\tt K}, {\tt L}, {\tt M}, {\tt Flattened O}, {\tt S}, {\tt X}, {\tt V}, {\tt Y}, {\tt 1}, {\tt 3}, {\tt 4}, {\tt 5}, {\tt Bent 5}, {\tt 8}, {\tt Open 8}.
    \item \textbf{Palm Orientation: } {\tt in}, {\tt out}, {\tt up}, {\tt down}, {\tt left}, {\tt right}.
    \item \textbf{Hand Location: } {\tt neutral}, {\tt chest}, {\tt ear/temple}, {\tt cheek/chin/mouth/nose}, {\tt forehead/eyes}, {\tt on shoulder}.
    \item \textbf{Hand Movement: } {\tt move towards to the body}, {\tt move away from the body}, {\tt up}, {\tt down}, {\tt left}, {\tt right}, {\tt circular}, 
    {\tt internal movement at the wrist}, {\tt internal movement at the fingers}.
    \item \textbf{Others: } {\tt movement repeats?}, {\tt one-hand sign?}.
\end{itemize}

As MS-ZSSLR-C/W datasets include different sign classes, handshape attributes are slightly altered to define new sign classes. The corresponding handshape attributes that are used in MS-ZSSLR-C/W datasets are as follows:
\begin{itemize}
    \item \textbf{Handshape:} {\tt A}, {\tt Open A}, {\tt B}, {\tt Baby 0}, {\tt Bent B}, {\tt Open B}, {\tt C}, {\tt D}, {\tt F}, {\tt G}, {\tt H}, {\tt K}, {\tt L}, {\tt Bent L}, {\tt Open N}, {\tt N}, {\tt Flattened O}, {\tt Baby O}, {\tt R}, {\tt S}, {\tt T}, {\tt W}, {\tt X}, {\tt V} , {\tt Y}, {\tt 1}, {\tt 5}, {\tt Bent 5}, {\tt 8}, {\tt Open 8}.
\end{itemize}

\begin{figure} 
\centering
   \includegraphics[height=55mm]{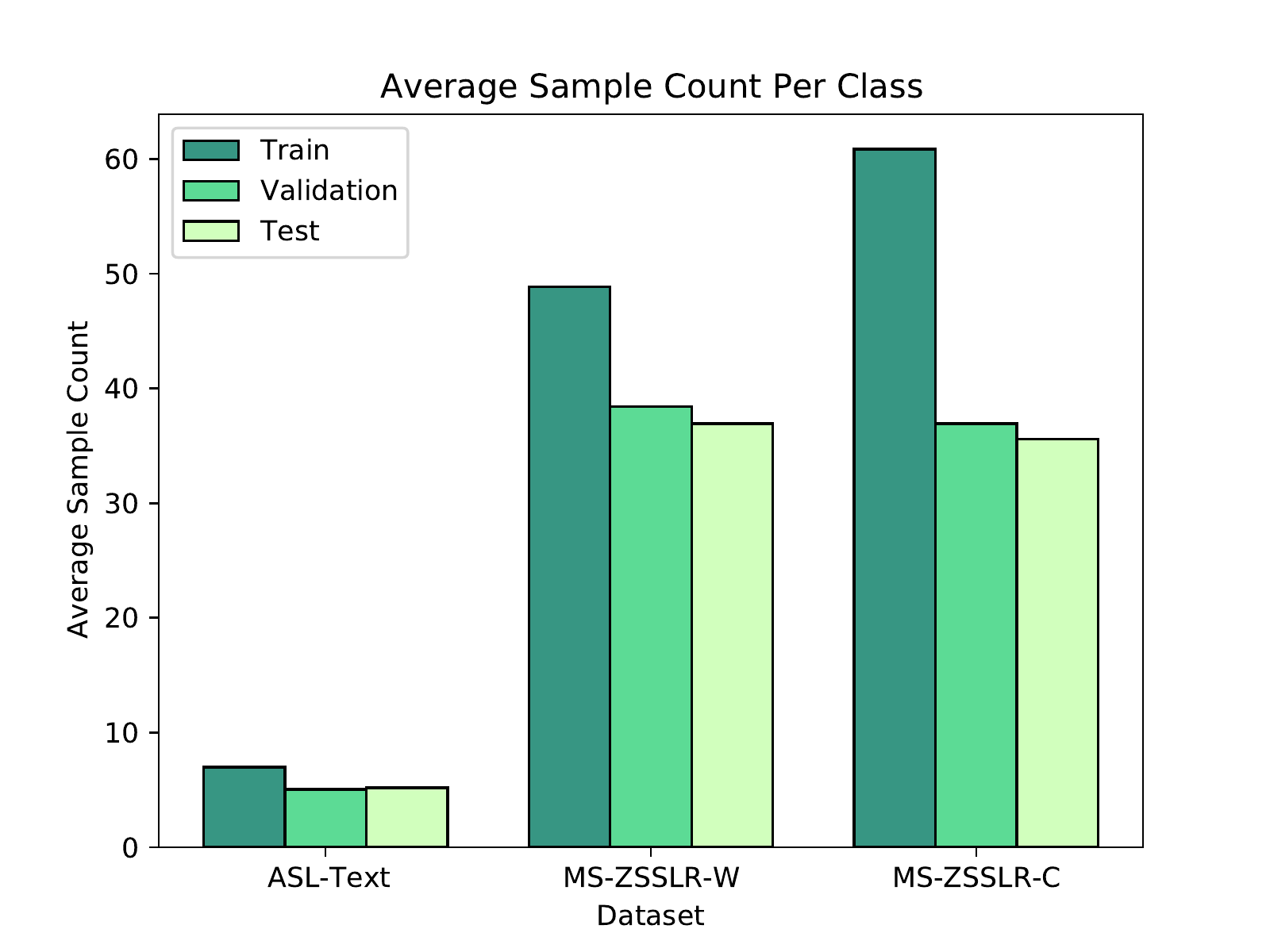}
    \caption{Average sample count per train, validation and test sets for ASL-Text, MS-ZSSLR-C, and MS-ZSSLR-W datasets. \label{fig:supp_distribution}}
\end{figure}

\section*{Dataset splits}

Figure \ref{fig:supp_distribution} shows average sample count per train, validation and test sets of ASL-Text, MS-ZSSLR-W and MS-ZSSLR-C datasets. 
We provide per-dataset split details in the following subsections.

\subsection*{ASL-Text Dataset}
ASL-Text dataset contains 170, 30, and 50 mutually exclusive classes in train, validation and test sets. The class names are listed below.

\noindent \textbf{Training classes:} Excuse, Include/Involve, Shelf/Floor, Answer, Boss, Dress/Clothes,  Marry, Stand-Up, Disappoint, Expert, Cancel/Criticize, Fed-Up/Full, Guitar, Emphasize, Government, Look, Afraid, Court, Medicine, Hello, Conflict/Intersection, Less, Of-Course, Dismiss, Dark, Silly,  Home, Blue, Appointment, Disconnect, A-Lot, Enter, Mad, Cold, Decide, Arrive, Inform, Proceed,  Miss/Assume, Letter/Mail, Keep, Fly-By-Plane, Deaf, High, Beautiful, Again, Happen, Depress,  Embarrass, Deposit, Drunk, Develop, Over/After, Brave/Recover, Avoid/Fall-Behind, Full, Blame,  Goal, Art/Design, Allow, Live, Future, Boy, Nice/Clean, Dry, Have, Take-Up, Heavy, Grow, Earth,  Friday, Down, Bore, Center, Cheap, Everyday, Divorce, Forget, Awkward, Grandfather, Cruel,  Graduate, Beer, Leave-There, Any, Chemistry, Brown, Friend, Left, Free, Freeze, Cannot, All, East,  Give-Up, Family, Bad, Green, Can, Learn, Coat, Drink, Head, Football, Lousy, Buy, Excited, Price,  Enough, Grandmother, Lie, Sausage/Hot-Dog, Thrill/Whats-Up, Shame, Same-Old, Messed-Up, Match,  Bar, Car, Helmet, Illegal, Merge/Mainstream, Chase, Work-Out, Weekend, Dirty, How-Many/Many, Gone,  Far, Head-Cold, Chain/Olympics, Line, Go-Away, Collect, Set-Up, Country, Really, Protest,  Flat-Tire, Lungs, Paint, Inject, Easy, Lip/Mouth, Nab, Fail, Fence, To-Fool, Gamble, Banana,  Introduce, Mosquito, Lend, Finally, Halloween, Exact, Hearing-Aid, Explain, Lecture, Bicycle,  Magazine, Increase, Disappear, Make, Lose-Competition, Experience, Expensive, Girl, Accept, But.

\noindent \textbf{Validation classes:} Meet, Finish, Advise/Influence, Course, Destroy, Cough, Alone,  Bridge, Call-By-Phone, Hard, Idea, Apple, Hospital, One-Month, Black, Grass, Borrow, Run-Out,  Bread, Monday, Library, One, Metal, Morning, Hit, Most, Meat, Come-On, Not-Mind, Smooth.

\noindent \textbf{Test classes:} Generation, Half, Engagement, Break-Down, Apply, Date/Dessert,  Shape/Statue, Pass, Insult, Hamburger, Obscure, Like, Crush, Less-Than, Bawl-Out, Blind,  Paper-Check/Card, Get-Up, Place, Cooperate/Unite, Insurance/Infection, Follow, Meeting, General,  Autumn, Comb, Experiment, Line-Up, Gas/Gas-Up, Grab-Chance, Permit, Eat, Tough, Trash/Bag,  Speech/Oral, Cherish, Strange, Association, Pull, Member, Ghost, Machine, Average, Act, Ahead,  Celebrate, Skin,Strong, Where, Concern.

\subsection*{MS-ZSSLR-C/W Dataset Zero-Shot Settings}
MS-ZSSLR-C/W datasets contain 120, 30, and 50 mutually exclusive classes in train, validation and test sets, respectively. The corresponding class names are listed below.

\textbf{Train classes:} Again, Apple, Aunt, Bad, Bathroom, Beautiful, Bird, Black, Blue, Book, Boring, Boy, Bread, Brother, Brown, But, Call, Can, Cat, Cheese, Coffee, College, Color, Cook,  Dance, Daughter, Day, Doctor, Draw, Drink, Eat, Family, Father, Fine, Finish, Fish, Forget, France,  Friend, Future, Girl, Go, Good, Grandfather, Grandmother, Green, Have, Hearing, Hello, Help, Home,  Horse, How-Many, Hungry, Hurt, Jacket, Know, Learn, Like, Man, Me, Milk, Mother, Must, Name, Nice,  Night, No, Not, Nothing, Now, Nurse, Old, Orange, Paper, Pencil, Pink, Play, Please, Purple, Read,  Red, Room, Sad, School, See, Sell, Shirt, Shoes, Sick, Sister, Sit, Slow, Sorry, Spring, Student,  Sunday, Table, Teach, Teacher, Thank-You, Time, Tired, Tomorrow, Understand, Walk, Want, Water,  What, When, Where, White, Who, Woman, Work, Write, Wrong, Yellow, Yes, You.

\textbf{Validation classes:} Afraid, Bicycle, Buy, Class, Cousin, Door, English, Hot, Late, Lost, Mad, Meet, Movie, Nephew, Not-Know, Not-Like, Remember, Restaurant, Run, Same, Sign, Son, Start, Tea, Week, Which, Why, Wife, Year, Your.

\textbf{Test classes:} All, Baby, Bed, Big, Boyfriend, Candy, Car, Cold, Computer, Cookie, Deaf,  Dentist, Divorce, Dog, Early, Enjoy, Excited, Favorite, Germany, Happy, Here, Hotdog, Homework, Hour, House, How, Kitchen, Library, Live, Love, Mean, Money, More, My, Party, Practice, Ready,  Right, Sandwich, Saturday, Soccer, Soda, Study, Today, Turkey, Ugly, Uncle, Watch, Wednesday,  Yesterday.

\begin{figure}
\scalebox{0.9}{
\begin{tabular}{>{\raggedright\arraybackslash}m{0.5\linewidth}>{\raggedright\arraybackslash}m{0.5\linewidth}}
\toprule
\textit{MS-ZSSLR-C} & \\
\includegraphics[width=0.3\linewidth, height=1.5cm]{images/my/my_0015.jpg}
\includegraphics[width=0.3\linewidth, height=1.5cm]{images/my/my_0030.jpg}
\includegraphics[width=0.3\linewidth, height=1.5cm]{images/my/my_0045.jpg} 
&
\makecell[t]{\begin{minipage}{0.25\textwidth}
\scriptsize{\vspace{-3mm}\textbf{MY}\\
Place the palm of the right open hand on the chest, fingers pointing left.}
\end{minipage}} \\

\includegraphics[width=0.3\linewidth, height=1.5cm]{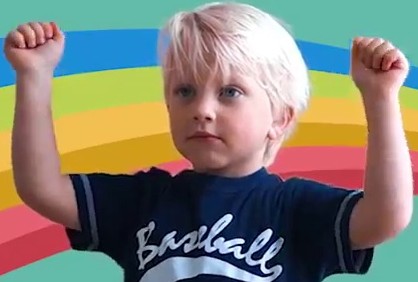}
\includegraphics[width=0.3\linewidth, height=1.5cm]{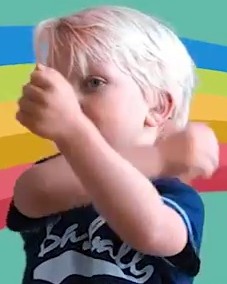}
\includegraphics[width=0.3\linewidth, height=1.5cm]{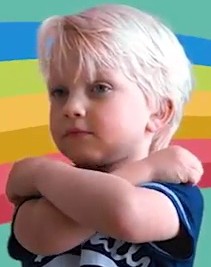} 
&
\makecell[t]{\begin{minipage}{0.25\textwidth}
\scriptsize{\vspace{-3mm}\textbf{LOVE}\\
With the wrists of both \textcolor{red}{S} hands crossed in front of the chest, palms facing in, bring the arms back against the chest.}
\end{minipage}} \\
\includegraphics[width=0.3\linewidth, height=1.5cm]{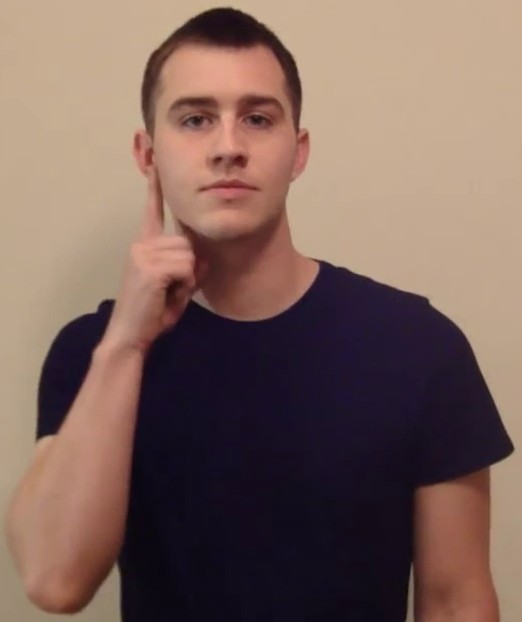}
\includegraphics[width=0.3\linewidth, height=1.5cm]{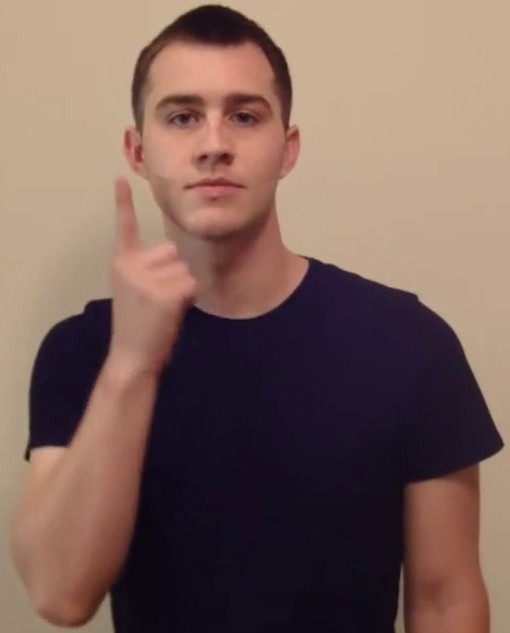}
\includegraphics[width=0.3\linewidth, height=1.5cm]{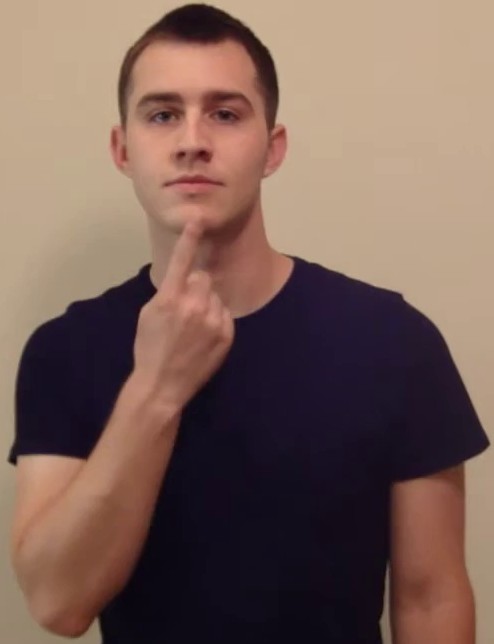} 
&
\makecell[t]{\begin{minipage}{0.25\textwidth}
\scriptsize{\vspace{-3mm}\textbf{DEAF}\\
Touch the extended right index finger first to near the right ear and then to near the right side of the mouth.}
\end{minipage}} \\

\midrule

\vspace{2mm}
\includegraphics[width=0.3\linewidth, height=1.5cm]{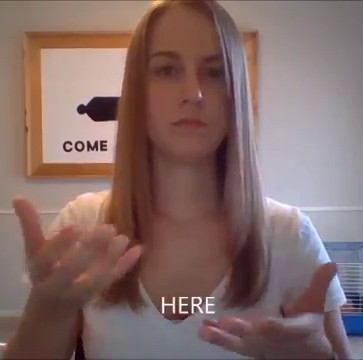}
\includegraphics[width=0.3\linewidth, height=1.5cm]{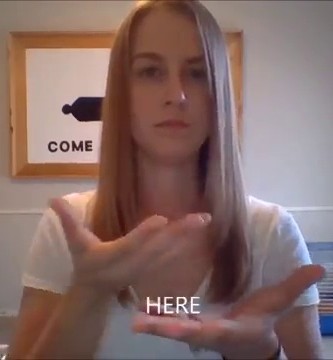}
\includegraphics[width=0.3\linewidth, height=1.5cm]{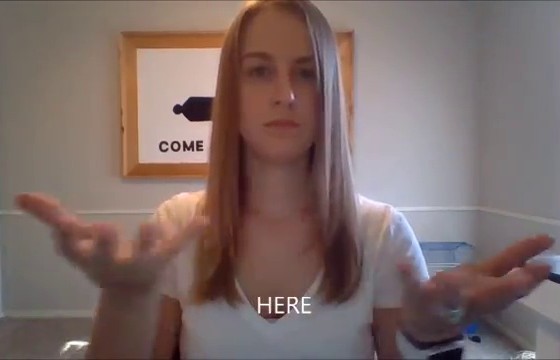} 
&
\makecell[t]{
\vspace{2mm}
\includegraphics[width=0.3\linewidth, height=1.5cm]{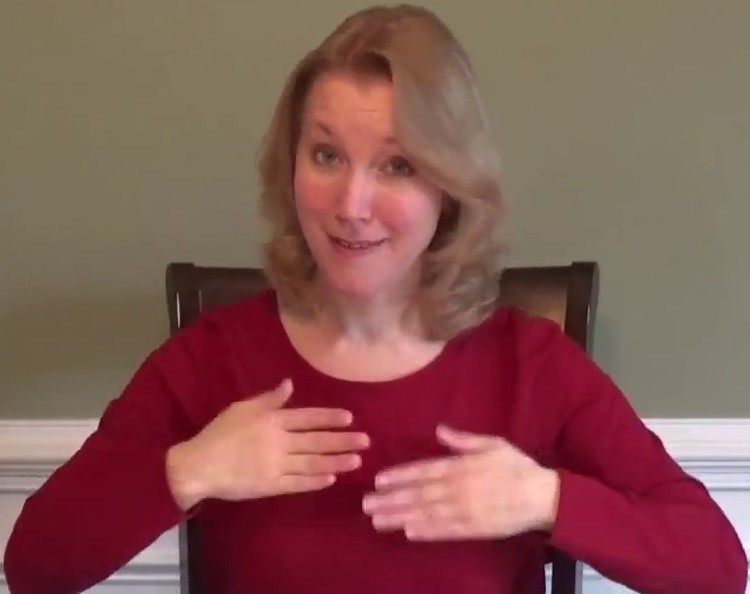}
\includegraphics[width=0.3\linewidth, height=1.5cm]{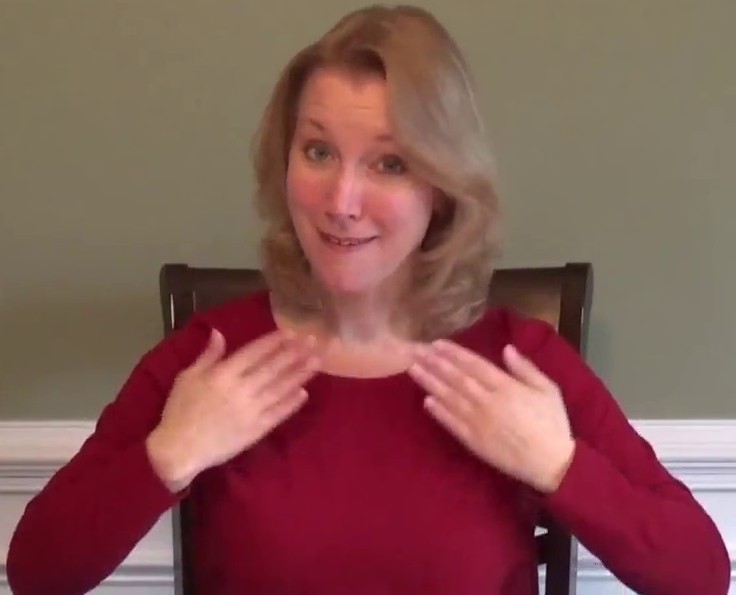}
\includegraphics[width=0.3\linewidth, height=1.5cm]{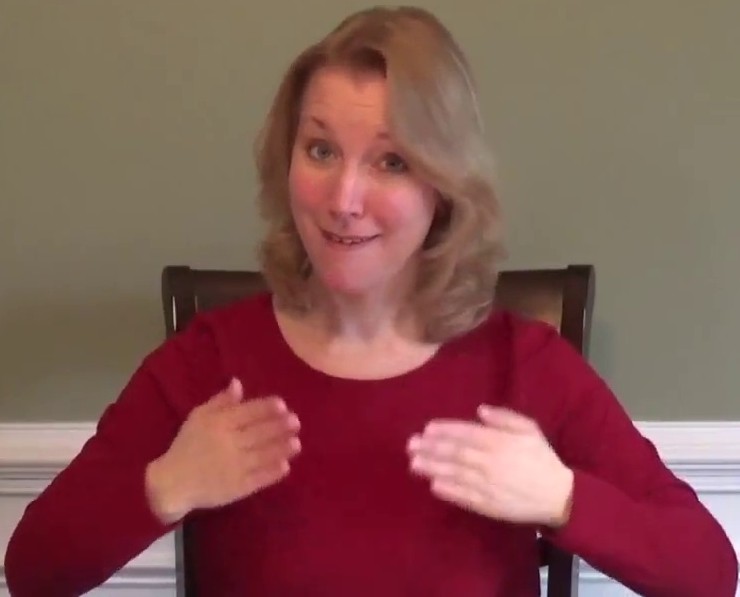}
} \\

\begin{minipage}{0.25\textwidth}
\scriptsize{\vspace{-3mm}\textbf{HERE}}\\
\scriptsize{Beginning with both curved hands in  front of  each side of the body, palms  facing up, move the hands toward  each other in repeated flat circles.}
\end{minipage}
 & 
\makecell[t]{\begin{minipage}{0.25\textwidth}
    \scriptsize{\vspace{-2mm}\textbf{HAPPY}\\
Brush the fingers of the right open  hand, palm facing in and fingers  pointing left, upward in a repeated  circular movement on the chest.}
\end{minipage}}  \\


\includegraphics[width=0.3\linewidth, height=1.5cm]{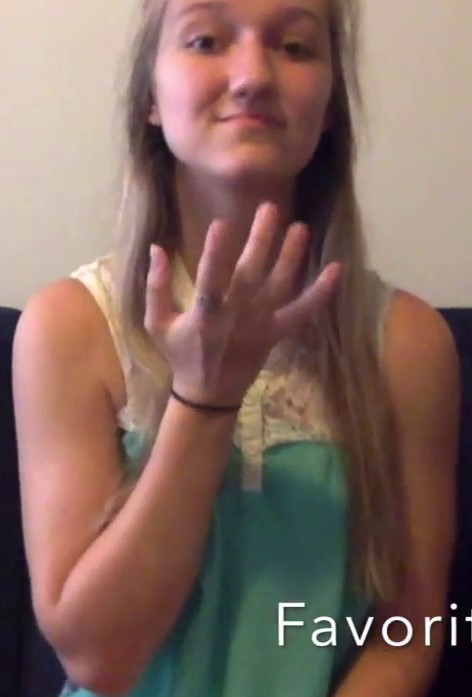}
\includegraphics[width=0.3\linewidth, height=1.5cm]{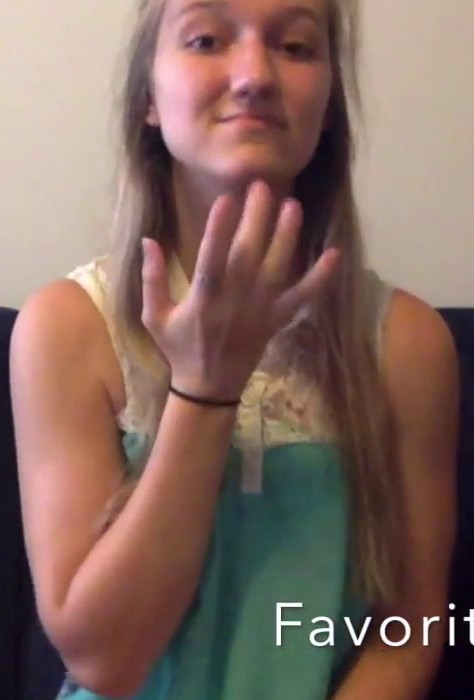}
\includegraphics[width=0.3\linewidth, height=1.5cm]{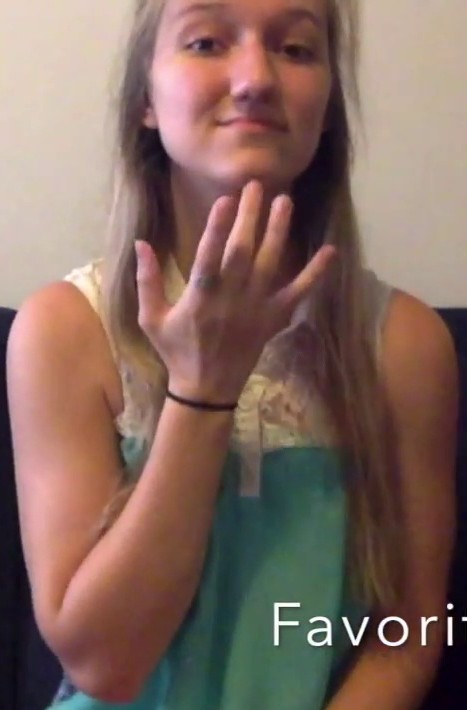} 
&
\makecell[t]{
\includegraphics[width=0.3\linewidth, height=1.5cm]{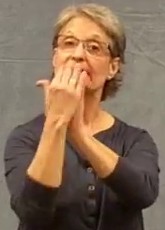}
\includegraphics[width=0.3\linewidth, height=1.5cm]{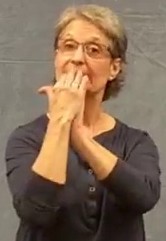}
\includegraphics[width=0.3\linewidth, height=1.5cm]{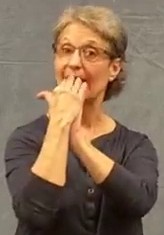}
} \\


\begin{minipage}{0.25\textwidth}
\scriptsize{\vspace{-1mm}\textbf{FAVORITE}}\\
\scriptsize{Touch the bent middle finger of the  right \textcolor{red}{5} hand, palm facing in, to the  chin with a double movement.}
\end{minipage}
 & 
\makecell[t]{\begin{minipage}{0.25\textwidth}
    \scriptsize{\vspace{-2mm}\textbf{SANDWICH}\\
    With the palms of both open hands  together, right hand above left, bring the  fingers back toward the mouth with a  short double movement.}
\end{minipage}}  \\

\includegraphics[width=0.3\linewidth, height=1.5cm]{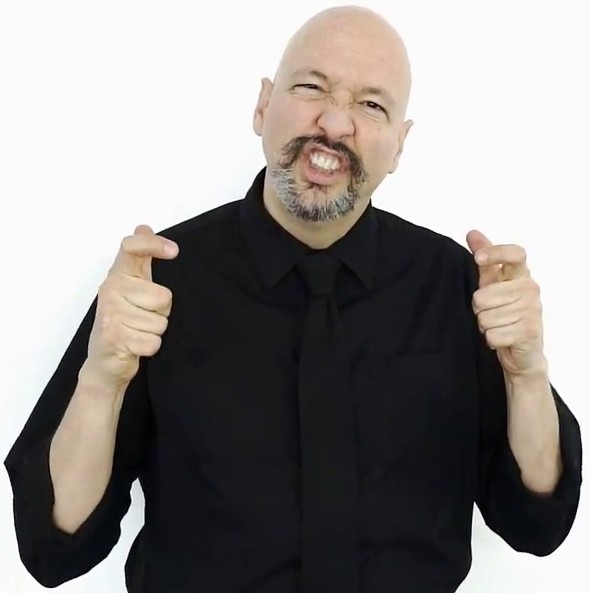}
\includegraphics[width=0.3\linewidth, height=1.5cm]{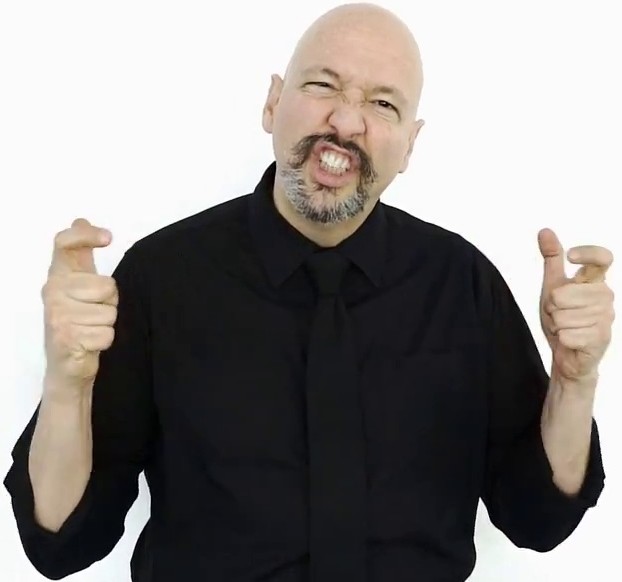}
\includegraphics[width=0.3\linewidth, height=1.5cm]{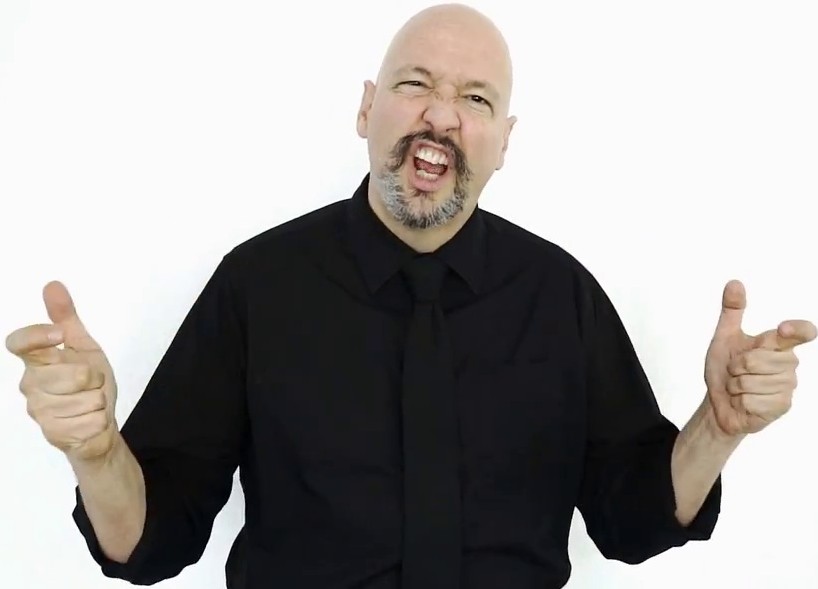} 
&
\makecell[t]{
\includegraphics[width=0.3\linewidth, height=1.5cm]{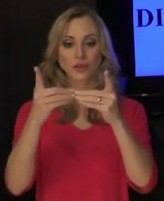}
\includegraphics[width=0.3\linewidth, height=1.5cm]{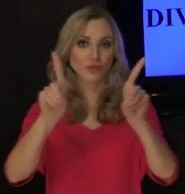}
\includegraphics[width=0.3\linewidth, height=1.5cm]{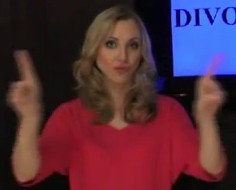}
} \\


\begin{minipage}{0.25\textwidth}
\scriptsize{\vspace{-9mm}\textbf{BIG}}\\
\scriptsize{Move both \textcolor{red}{L} hands from in front of  each side of the chest, palms facing  each other, in large arcs beyond each  side of the body.

}
\end{minipage}
 & 
\makecell[t]{\begin{minipage}{0.25\textwidth}
    \scriptsize{\vspace{-2mm}\textbf{DIVORCE}\\
    Beginning with the fingertips of both  the \textcolor{red}{D} hands touching in front of chest,  palms facing each other and index fingers  pointing up, swing the hands away from  each other by twisting the wrists, ending  with the hands in front of each side of the  body, palms facing forward.}
\end{minipage}}
\\
\bottomrule
\end{tabular}}
\caption{Example predictions of the proposed model on the MS-ZSSLR-C dataset. The first three rows show examples that are correctly predicted and the last three rows show incorrect predictions. For the correct predictions, second column includes corresponding textual description. For the incorrect predictions, the second column includes ground truth textual descriptions and frames from an example video of the ground truth class. \label{fig:prediction_supmat_ms}}
\end{figure}

\begin{figure}
\centering
\scalebox{0.9}{
\begin{tabular}{>{\raggedright\arraybackslash}m{0.5\linewidth}>{\raggedright\arraybackslash}m{0.5\linewidth}}
\toprule
\textit{ASL-Text} & \\
\includegraphics[width=0.3\linewidth, height=1.5cm]{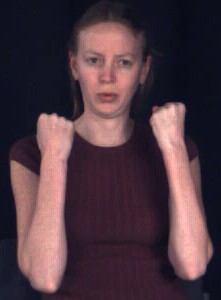}
\includegraphics[width=0.3\linewidth, height=1.5cm]{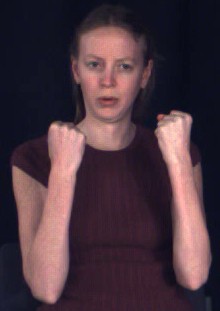}
\includegraphics[width=0.3\linewidth, height=1.5cm]{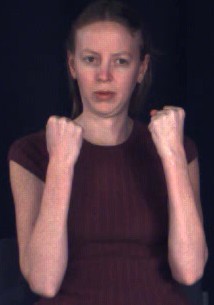} 
&
\makecell[t]{\begin{minipage}{0.25\textwidth}
\scriptsize{\vspace{-3mm}\textbf{STRONG}\\
Move both \textcolor{red}{S} hands, palms facing in, forward with a short deliberate movement from in front of each shoulder.}
\end{minipage}} \\

\includegraphics[width=0.3\linewidth, height=1.5cm]{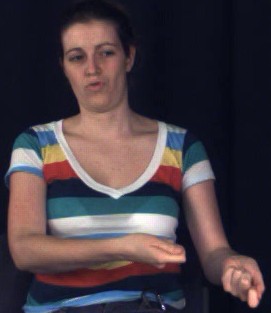}
\includegraphics[width=0.3\linewidth, height=1.5cm]{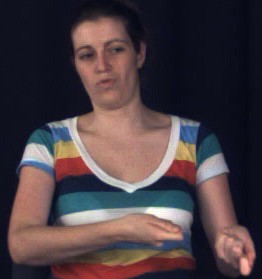}
\includegraphics[width=0.3\linewidth, height=1.5cm]{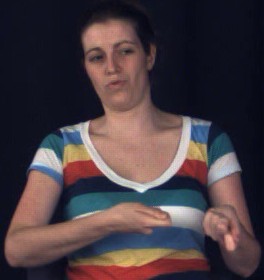} 
&
\makecell[t]{\begin{minipage}{0.25\textwidth}
\scriptsize{\vspace{-3mm}\textbf{PULL}\\
Beginning with the right curved hand in front of the body and the left curved hand somewhat forward, both palms facing up, bring the hands back toward the right side of the body while closing them into \textcolor{red}{A} hands.}
\end{minipage}} \\

\includegraphics[width=0.3\linewidth, height=1.5cm]{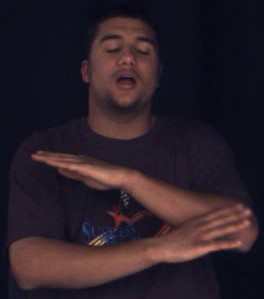}
\includegraphics[width=0.3\linewidth, height=1.5cm]{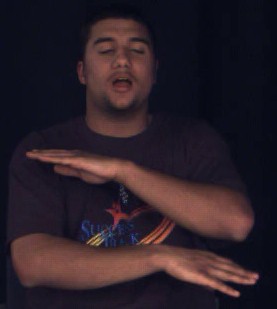}
\includegraphics[width=0.3\linewidth, height=1.5cm]{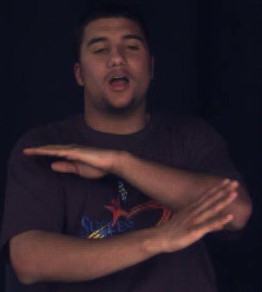} 
&
\makecell[t]{\begin{minipage}{0.25\textwidth}
\scriptsize{\vspace{-2mm}\textbf{AUTUMN}\\
Brush the index-finger side of the right \textcolor{red}{B} hand, palm facing down, downward toward the elbow of the left forearm, held bent across the chest.}
\end{minipage}} \\
\midrule
\vspace{2mm}
\includegraphics[width=0.3\linewidth, height=1.5cm]{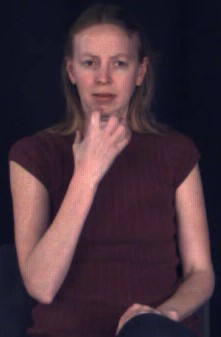}
\includegraphics[width=0.3\linewidth, height=1.5cm]{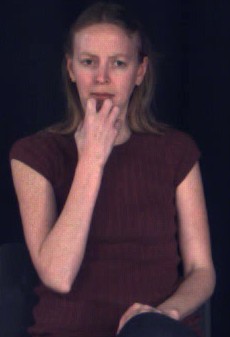}
\includegraphics[width=0.3\linewidth, height=1.5cm]{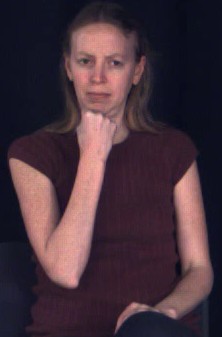} 
&
\vspace{2mm}
\makecell[t]{
\includegraphics[width=0.3\linewidth, height=1.5cm]{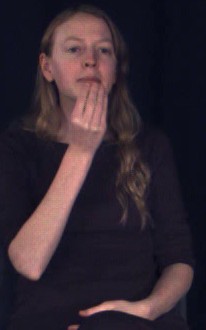}
\includegraphics[width=0.3\linewidth, height=1.5cm]{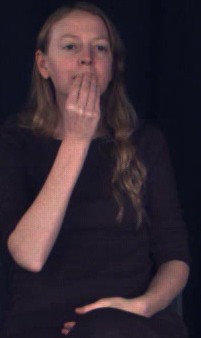}
\includegraphics[width=0.3\linewidth, height=1.5cm]{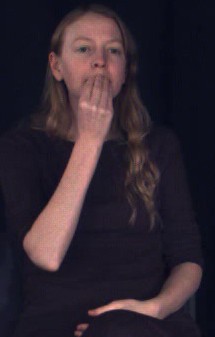}
} \\


\begin{minipage}{0.25\textwidth}
\scriptsize{\vspace{-3mm}\textbf{CHERISH}}\\
\scriptsize{Beginning with the right curved \textcolor{red}{5} hand  in front of the mouth, palm facing back,  slowly close the fingers into an \textcolor{red}{S} hand.}
\end{minipage}
 & 
\makecell[t]{\begin{minipage}{0.25\textwidth}
\scriptsize{\vspace{-2mm}\textbf{EAT}\\
Bring the fingertips of the right flattened  \textcolor{red}{O}  hand, palm facing in, to the lips with a  repeated movement.}
\end{minipage}}  \\


\includegraphics[width=0.3\linewidth, height=1.5cm]{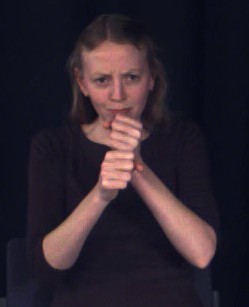}
\includegraphics[width=0.3\linewidth, height=1.5cm]{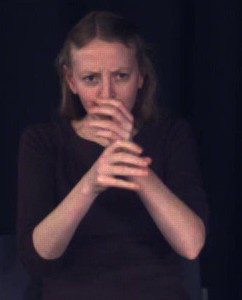}
\includegraphics[width=0.3\linewidth, height=1.5cm]{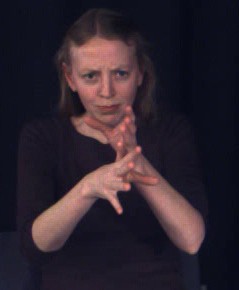} 
&
\makecell[t]{
\includegraphics[width=0.3\linewidth, height=1.5cm]{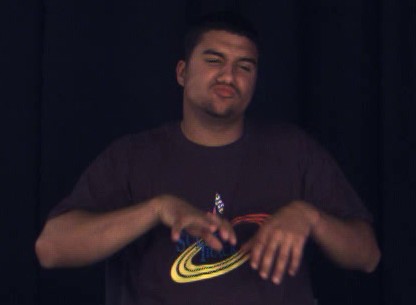}
\includegraphics[width=0.3\linewidth, height=1.5cm]{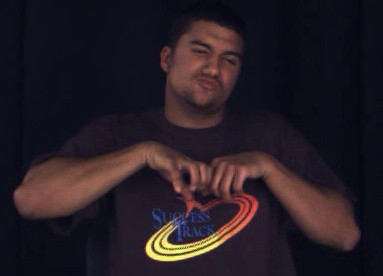}
\includegraphics[width=0.3\linewidth, height=1.5cm]{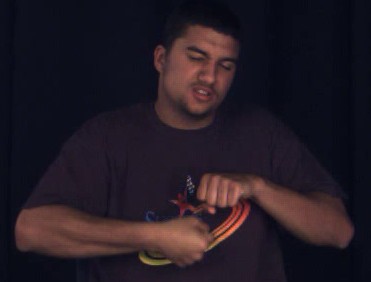}
} \\


\begin{minipage}{0.25\textwidth}
\scriptsize{\vspace{-1mm}\textbf{BAWL-OUT}}\\
\scriptsize{Beginning with the little finger of the right \textcolor{red}{S} hand on the top of the index finger side of the left \textcolor{red}{S} hand, flick the  hands forward with a deliberate double movement  while opening the  fingers into \textcolor{red}{5} hands each time.}
\end{minipage}
 & 
\makecell[t]{\begin{minipage}{0.25\textwidth}
    \scriptsize{\vspace{-3mm}\textbf{CRUSH}\\
    Beginning with the palms of both \textcolor{red}{A}  hands together in front of the chest,  twist the hands in opposite directions.}
\end{minipage}}  \\
\vspace{1mm}
\includegraphics[width=0.3\linewidth, height=1.5cm]{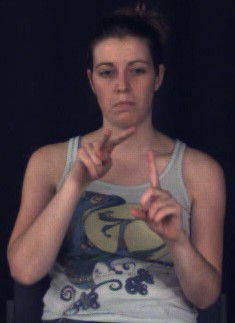}
\includegraphics[width=0.3\linewidth, height=1.5cm]{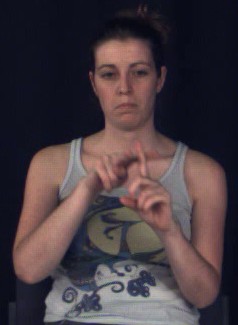}
\includegraphics[width=0.3\linewidth, height=1.5cm]{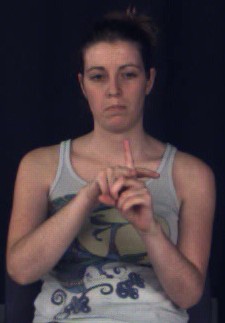} 
&
\makecell[t]{
\includegraphics[width=0.3\linewidth, height=1.5cm]{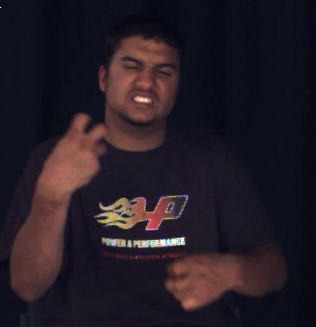}
\includegraphics[width=0.3\linewidth, height=1.5cm]{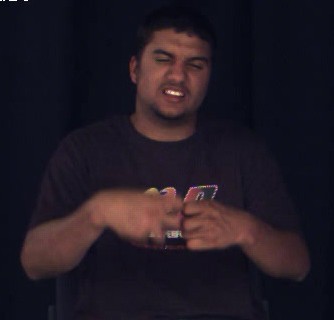}
\includegraphics[width=0.3\linewidth, height=1.5cm]{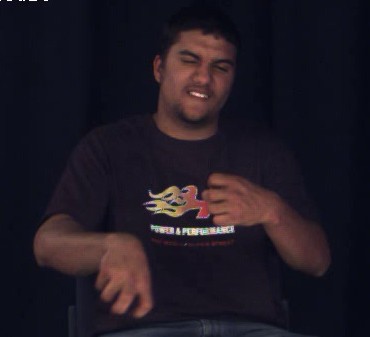}
} \\


\begin{minipage}{0.25\textwidth}
\scriptsize{\vspace{-9mm}\textbf{APPLY}}\\
\scriptsize{Move the fingers of the right \textcolor{red}{V} hand, palm  facing forward, downward on each side  of the extended left index finger, pointing  up in front of the chest.

}
\end{minipage}
 & 
\makecell[t]{\begin{minipage}{0.25\textwidth}
    \scriptsize{\vspace{-2mm}\textbf{TOUGH}\\
    Beginning with both bent \textcolor{red}{V} hands in  front of the chest, right hand higher  than the left hand, palms facing in, move  the right hand down and the left  hand upward with an alternating  movement, brushing the knuckles of each  hand as the hands move in the opposite  direction.}
\end{minipage}} \\
\bottomrule
\end{tabular}}
\caption{Example predictions of the proposed model on the ASL-Text dataset. The first three rows show examples that are correctly predicted and the last three rows show incorrect predictions. For the correct predictions, second column includes corresponding textual description. For the incorrect predictions, the second column includes ground truth textual descriptions and frames from an example video of the ground truth class. \label{fig:prediction_supmat_asl}}
\end{figure}

\section*{Additional Qualitative Results }
Figure \ref{fig:prediction_supmat_ms} and \ref{fig:prediction_supmat_asl} present examples for correct and incorrect classifications. We observe that most confusions occur across classes with similar hand movements and/or locations, highlighting the importance of handshape pattern modeling.
We also observe that class definitions also tend to be similar across the confused classes. These qualitative examples further demonstrate that the problem domain can benefit from more detailed visual and auxiliary data representations.

\end{document}